\documentclass{article} 
\usepackage{iclr2025_conference,times}

\usepackage{amsmath,amsfonts,bm}

\def\eqref#1{equation~\ref{#1}}

\def\1{\bm{1}}

\DeclareMathAlphabet{\mathsfit}{\encodingdefault}{\sfdefault}{m}{sl}
\SetMathAlphabet{\mathsfit}{bold}{\encodingdefault}{\sfdefault}{bx}{n}

\usepackage{booktabs}
\usepackage{siunitx}
\usepackage{hyperref}
\usepackage{url}
\usepackage{graphicx}
\usepackage{booktabs}
\usepackage[most]{tcolorbox}
\usepackage{listings}
\usepackage{enumitem}
\usepackage{xcolor}
\usepackage{multirow}
\usepackage{adjustbox}
\usepackage{tcolorbox}
\tcbuselibrary{listings,skins}
\usepackage{listings}
\usepackage{siunitx}
\usepackage{multirow}
\usepackage{booktabs}
\usepackage{pdflscape}
\usepackage{makecell}
\usepackage{arydshln}
\usepackage{subcaption}

\usepackage{pifont}

\newcommand{\cmark}{\textcolor{green}{\ding{51}}} 
\newcommand{\xmark}{\textcolor{red}{\ding{55}}}   

\title{IndicVisionBench: Benchmarking Cultural and Multilingual Understanding in VLMs}

\author{
\textbf{Ali Faraz}$^{1}$, 
\textbf{Akash}$^{2}$, 
\textbf{Shaharukh Khan}$^{1}$, 
\textbf{Raja Kolla}$^{1}$,
\textbf{Akshat Patidar}$^{1}$,\\[2pt]
\textbf{Suranjan Goswami}$^{2}$, 
\textbf{Abhinav Ravi}$^{1}$,
\textbf{Chandra Khatri}$^{1}$, 
\textbf{Shubham Agarwal}$^{1}$\\[10pt]
$^{1}$\textit{Krutrim AI, Bangalore, India}\\[2pt]
$^{2}$\textit{OLA Electric, Bangalore, India}\\[6pt]
\textsuperscript{Contact: \{ali.faraz, raja.kolla, shubham.agarwal1\}@olakrutrim.com,\{akash.shyam, suranjan.goswami\}@olaelectric.com}
}

\iclrfinalcopy 
\begin{document}

\maketitle

\begin{abstract}

Vision-language models (VLMs) have demonstrated impressive generalization across multimodal tasks, yet most evaluation benchmarks remain Western-centric, leaving open questions about their performance in culturally diverse and multilingual settings. To address this gap, we introduce \textit{IndicVisionBench}, the first large-scale benchmark centered on the Indian subcontinent. Covering English and 10 Indian languages, our benchmark spans 3 multimodal tasks, including Optical Character Recognition (OCR), Multimodal Machine Translation (MMT), and Visual Question Answering (VQA), covering 6 kinds of question types. Our final benchmark consists of a total of ~5K images and 37K+ QA pairs across 13 culturally grounded topics. In addition, we release a paired parallel corpus of annotations across 10 Indic languages, creating a unique resource for analyzing cultural and linguistic biases in VLMs. We evaluate a broad spectrum of 8 models, from proprietary closed-source systems to open-weights medium and large-scale models. Our experiments reveal substantial performance gaps, underscoring the limitations of current VLMs in culturally diverse contexts. By centering cultural diversity and multilinguality, IndicVisionBench establishes a reproducible evaluation framework that paves the way for more inclusive multimodal research.
\end{abstract}

\section{Introduction}
\label{sec:intro}

Vision-language models (VLMs) \citep{bai2023qwen,chen2024far,lu2024deepseek,wang2024cogvlm,laurencon2024,tong2024cambrian,xue2024xgen} have demonstrated strong performance across a variety of multimodal tasks. However, existing benchmarks~\citep{antol2015vqa,fu2023mme,goyal2017making} remain heavily Western-centric, limiting our understanding of how these models generalize to culturally diverse and multilingual settings. India, in particular, represents one of the most culturally and linguistically diverse regions globally, with 22 official languages and 28 states plus 8 Union Territories\footnote{\url{https://en.wikipedia.org/wiki/States_and_union_territories_of_India}}
, each with distinct ethnic, visual, and cultural identities. While some recent efforts partially cover this diversity~\citep{romero2024cvqa, nayak2024benchmarkingvisionlanguagemodels, vayani2025languagesmatterevaluatinglmms}, a systematic, large-scale benchmark capturing India-specific cultural concepts across multiple languages is still lacking.

\begin{figure}[htbp]
    \centering
    
    \includegraphics[width=0.85\linewidth]{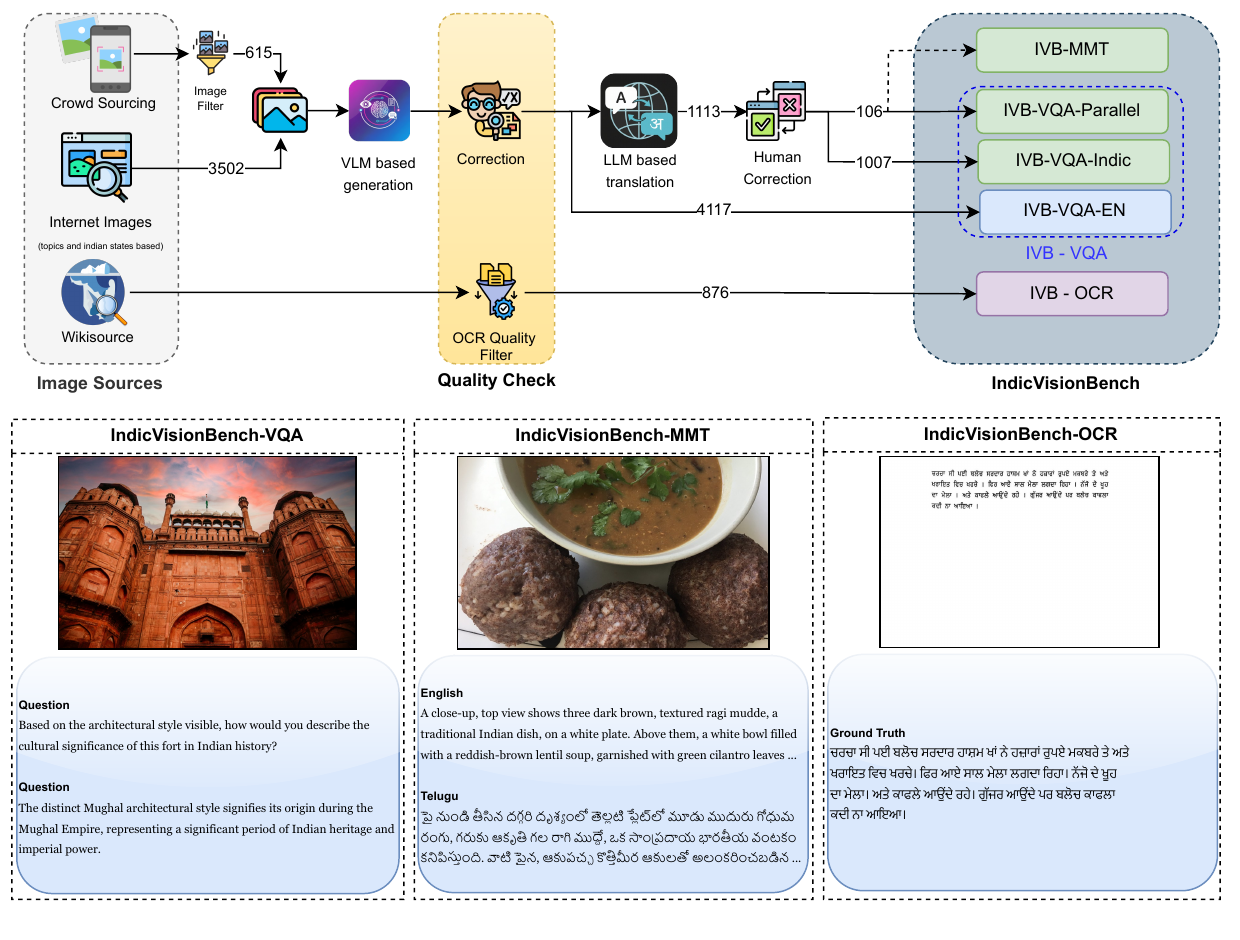}     \caption{\textbf{IndicVisionBench (IVB) pipeline and 3 tracks.} Top panel illustrates our image collection pipeline for 10 Indian languages, showing the number of images at each step, with human quality checks applied throughout. We also present sample outputs for the three tracks: VQA (Visual Question Answering) in English, MMT (Multimodal Machine Translation) in Telugu, and OCR (Optical Character Recognition) in Punjabi. Further details are provided in Section \ref{sec:benchmark}.
}
    \label{fig:teaser}
\end{figure}

To address this gap, we introduce \textbf{IndicVisionBench}, a culturally grounded evaluation benchmark tailored for the Indian subcontinent. To the best of our knowledge, this is the first large-scale benchmark explicitly designed to assess VLMs in the context of Indian culture and languages. We use states as a proxy for cultural groups following prior works~\citep{adilazuarda2024towards,nayak2024benchmarkingvisionlanguagemodels}. IndicVisionBench comprises 5K unique images and 37K+ question-answer pairs spanning 13 cultural topics, covering English and 10 medium-to-low resource Indic languages supporting three multimodal tracks: \textit{Visual Question Answering (VQA)}, \textit{Optical Character Recognition (OCR)}, and \textit{Multimodal Machine Translation (MMT)}. Figure \ref{fig:teaser} illustrates examples reflecting diverse cultural nuances, including monuments, food, and digitized text. Rigorous human verification and correction at every stage of data collection ensure the reliability and cultural fidelity of the benchmark, covering medium-to-low resource languages including Hindi, Bengali, Tamil, Malayalam, Telugu, Marathi, Kannada, Gujarati, Punjabi, and Oriya.

In this study, we evaluate 8 state-of-the-art (SOTA) VLMs on IndicVisionBench and find that performance drops considerably for low-resource languages and culturally specific content. We also observe a clear gap between proprietary and open-source models in their ability to capture linguistic and cultural nuances across multimodal tasks. Analysis across scripts and language groups further highlight the need for better support and representation of underrepresented regions and cultures.


Our contributions could thus be summarized as follows:
\begin{itemize}
\item We propose \textit{IndicVisionBench} as the first large-scale, Indian-centric benchmark for evaluating VLMs on culture-specific understanding, involving OCR, recognition, cultural identification, multi modal translation and semantic understanding involving 5K unique images.
\item We conduct a comprehensive evaluation of 8 prominent closed-source as well as open-weight models supporting Indian languages and contrast their performance across all the 3 tracks. We highlight systematic performance gaps that underscore the limitations of current general-purpose VLMs in culturally diverse settings.
\item We systematically study the regional-language biases, performance across topics and cross-lingual variation in performance. We will open source this benchmark after acceptance, for the future research in this direction. 

\end{itemize}

\section{Related Work}
\label{sec:related-work}

\paragraph{Vision Language Models and Benchmarks.}  Cross-attention
models \citep{alayrac2022flamingo,singh2022flava} and later \textit{visual instruction tuning} based auto-regressive models like the LLaVA family~\citep{liu2024visual,liu2024improved}, 
have advanced multimodal learning, where vision encoders~\citep{radford2021learning,zhai2023sigmoid,tschannen2025siglip} are aligned with large language models. This approach has since influenced a range of VLMs~\citep{lu2024deepseek,laurenccon2024matters,tong2024cambrian,xue2024xgen,team2024gemma}, which follow similar design principles and achieve strong results on translation, captioning, and multi-turn vision language benchmarks~\citep{hudson2019gqa, fu2023mme, yu2023mm}. In contrast, multimodal models that handle Indic languages remain relatively underexplored. Most open-source systems provide support only for 2 to 4 medium-resource Indian languages~\citep{maaz2024palo,alam2025behind,yue2024pangea}, with the notable exception of Chitrarth~\citep{khan2025chitrarth}, which extends coverage to all ten languages, considered in this work. We include all these models in our benchmark to assess their relative strengths.

\textbf{Optical Character Recognition (OCR).}  
OCR has progressed from early rule-based engines such as Tesseract \citep{smith2007overview} to modern transformer-based approaches like TrOCR \citep{li2021trocr} and docTR \citep{liao2023doctr}. Recent efforts in document understanding further leverage multimodal architectures, including the DocOwl series~\citep{hu2024mplug,docowl2_2024}, DocLLM~\citep{wang2023docllm}, and Donut~\citep{kim2022donut}. These systems are typically evaluated on benchmarks such as RVL-CDIP~\citep{harley2015icdar}, FUNSD~\citep{jaume2019funsd}, and DocVQA~\citep{mathew2021docvqa}, to name a few. Despite this progress, existing OCR benchmarks are largely English-centric, offering minimal coverage of Indic scripts and multilingual contexts.

\paragraph{Multimodal Machine Translation (MMT).} 
Recently, Multimodal Machine Translation (MMT)~\citep{calixto-liu-2017-incorporating,elliott-kadar-2017-imagination,delbrouck-dupont-2017-empirical,yao2020multimodal} has gained traction, where the translation leverages auxiliary modalities (e.g., images). 
Prior works have largely centered on English-European language pairs~\citep{elliott2016multi30k,specia2016shared}, with a subset of medium-resource Indian languages (particularly Hindi, Bengali, Malayalam) also explored in the 
shared task series~\citep{nakazawa2019proceedings,wat-2020-asian,wat-2021-asian,wat-2022-asian,wat-2023-asian} based on Visual Genome images \citep{krishna2017visual}. We support a similar task based on a diverse set of cultural images avoiding potential data contamination issues \citep{balloccu2024leak}. 

\paragraph{Cultural VQA.}  
Several benchmarks have begun probing cultural and multilingual reasoning in VLMs. GD-VCR~\citep{yin2021broaden} and Henna~\citep{alwajih2024peacock} emphasize culturally specific content but are largely limited to English or Arabic, while WorldCuisines~\citep{winata2025worldcuisinesmassivescalebenchmarkmultilingual} focuses on food and cuisines. Multilingual benchmarks~\citep{liu2023mmbench, zhang2023m3exam, sun2024parrot, das2024exams, wang2024m4u, fu2024mmecomprehensiveevaluationbenchmark} expand language coverage but often lack cultural and task diversity. Datasets like MaRVL~\citep{liu2021visually} and xGQA~\citep{pfeiffer2021xgqa} broaden multilingual reasoning but do not incorporate Indic cultural grounding. Closest to our work are CVQA~\citep{romero2024cvqa}, CulturalVQA~\citep{nayak2024benchmarkingvisionlanguagemodels}, and ALM-Bench~\citep{vayani2025languagesmatterevaluatinglmms}, which partially touch on India-specific contexts, yet none offers a unified framework capturing both Indic cultural diversity and multilingual multimodal evaluation.

\begin{figure*}[htbp]
    \centering
    \includegraphics[width=0.56\textwidth]{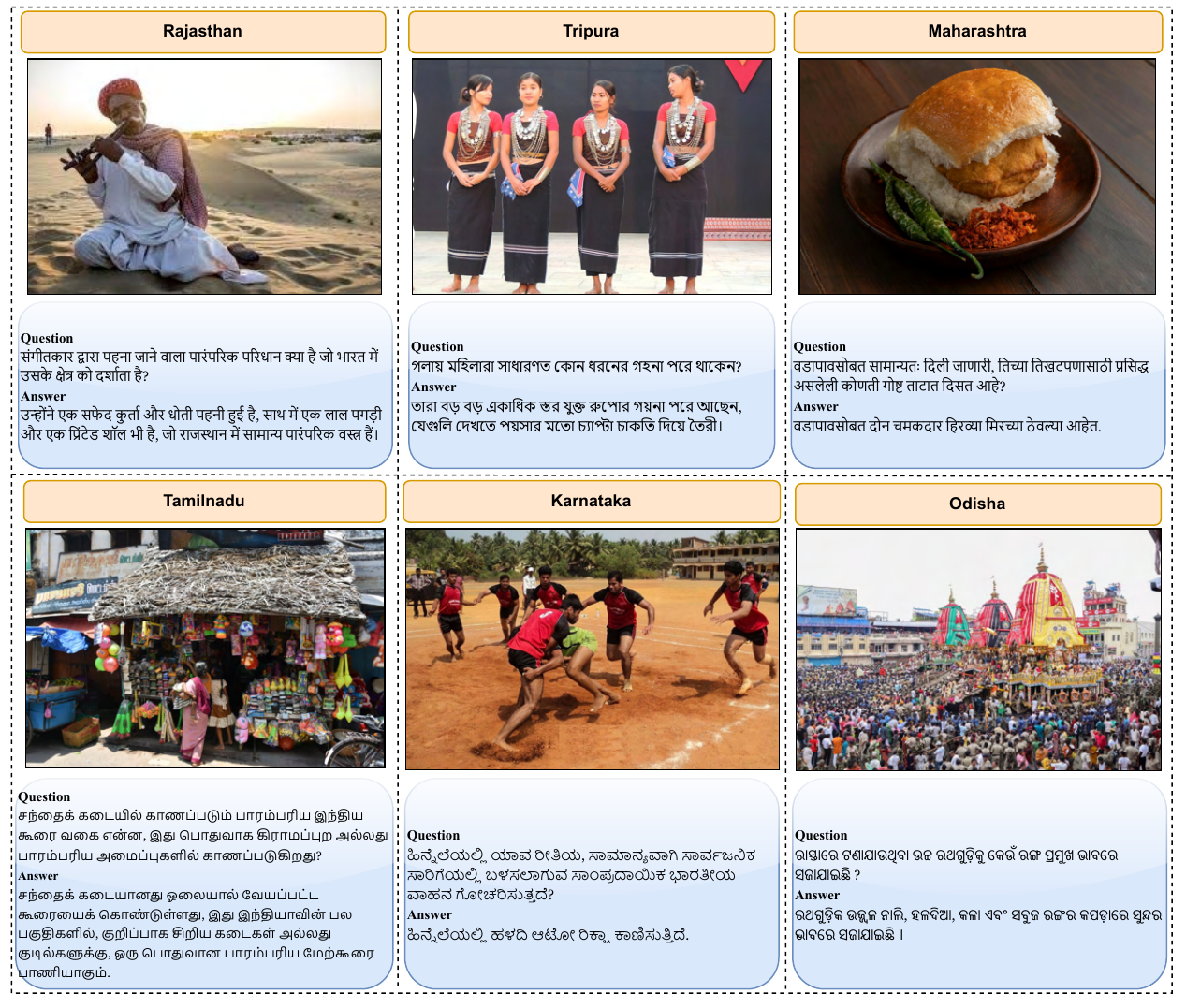}%
    \hspace{0.02\textwidth}%
    \includegraphics[width=0.42\textwidth]{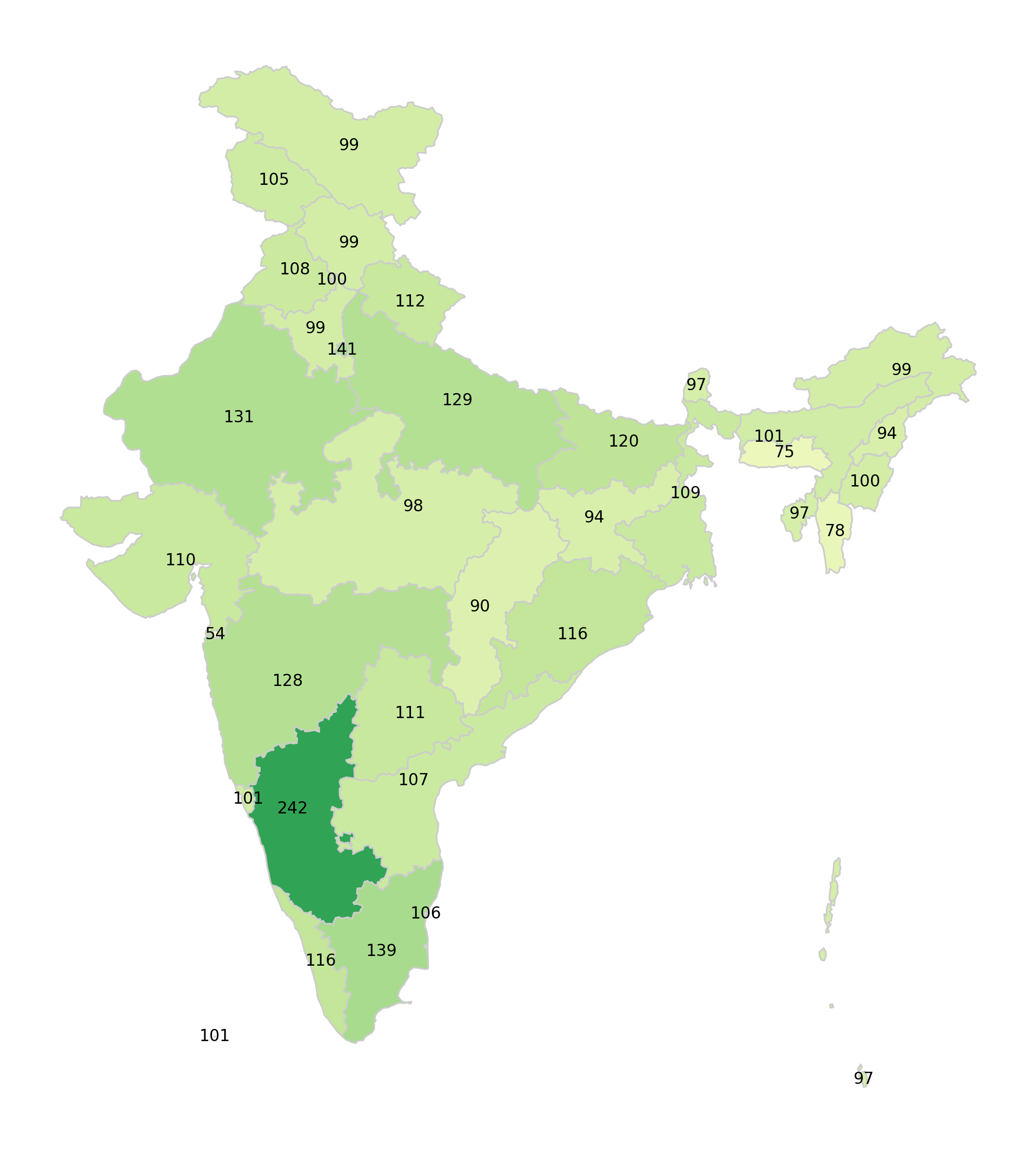}
    
    \caption{\textbf{Examples from IndicVisionBench-VQA.} Illustrative samples from different regions are shown on the left. The map on the right depicts the regional distribution of images across India, with counts per State/UT. Further details are provided in Section~\ref{appendix:dataset} of the Appendix.}
    \label{fig:plot}
\end{figure*}

\section{Benchmark Creation}
\label{sec:benchmark} 
Figure \ref{fig:teaser} illustrates our curation pipeline across all tracks; additional details are provided below.

\subsection{IndicVisionBench-VQA}
We constructed the VQA split using two approaches: (i) controlled crowd-sourcing and (ii) large-scale web crawling. In the first phase, we recruited volunteers (including authors) who contributed images captured on their personal devices along with corresponding annotations. These images were further reviewed to determine whether they were culturally specific to India and, if so, mapped to one of 13 predefined topics and to the relevant State/Union Territory (UT). Irrelevant images were discarded, resulting in 615 valid samples. As several categories and regions were underrepresented, we expanded coverage in the second phase, where cultural experts systematically collected Creative Commons–licensed images\footnote{\url{https://creativecommons.org/share-your-work/cclicenses/}}
 from Google Search, targeting roughly 100 per State/UT across the same categories. This yielded 3,502 additional images, bringing the total corpus to 4,117 (3,797 region-specific and 320 pan-India). 

Each image was first annotated with concise keywords by humans, expanded into intermediary synthetic detailed captions using VLMs in English, and then used to generate six QAs per image: two short-answer, one long-answer, one multiple-choice (single-correct), one True/False, and one adversarial question. Notably, adversarial questions incorporate false assumptions, requiring models to explicitly reject them, enabling a systematic probe of cultural knowledge beyond surface-level recognition. We employed Gemini-1.5-Flash and Gemini-2.5-Flash~\citep{gemini2025flash} for QA generation, informed by a small pilot study and cost considerations (see Appendix; Table \ref{appendix:cost_comparison}). Human reviewers then refined all outputs for factual accuracy and cultural alignment, resulting in a balanced set of open-ended queries that jointly test recognition, reasoning, and robustness in VLMs.  Guidelines provided to annotators are detailed in Appendix~\ref{appendix:annotations} while Figure \ref{fig:gradio_annot_interface} shows the annotation interface which we will open-source after acceptance. 

From this pool of 4K+ images and their corresponding 6 QAs, we translated a subset into the dominant regional language using text-only Gemini call, followed by human correction, resulting in an \textit{VQA-Indic} version. Additionally, we sampled a disjoint set of 106 images and translated them into all 10 Indian languages, creating a \textit{VQA-Parallel} corpus to systematically study cross-lingual variation in VLMs’ cultural understanding and robustness.

\subsection{IndicVisionBench-MMT}
The Multimodal Machine Translation (MMT) track extends the 106 images from the \textit{VQA-Parallel} corpus, where each English caption was translated into 10 Indic languages with access to the image context. All translations were manually annotated to preserve meaning and align with cultural nuances, resulting in a multimodal parallel dataset tailored for evaluating vision-grounded multimodal translation in medium-to-low resource Indic languages.

\subsection{IndicVisionBench-OCR} 
For benchmarking OCR performance, we construct a multilingual corpus from Wikisource \citep{wikisource2025}, a public-domain repository of digitized literary works. The corpus spans 10 Indic languages and includes both printed and handwritten styles. To ensure reliability, we restrict collection to Level-4 verified pages, which have been human-reviewed on the platform. For each page, we pair high-resolution scans (\texttt{prppageimage}) with their corresponding verified text (\texttt{pagetext}). Further implementation details are provided in Appendix  ~\ref{appendix:implementation}.

\begin{figure}[htbp]
    \centering
    \includegraphics[width=\linewidth]{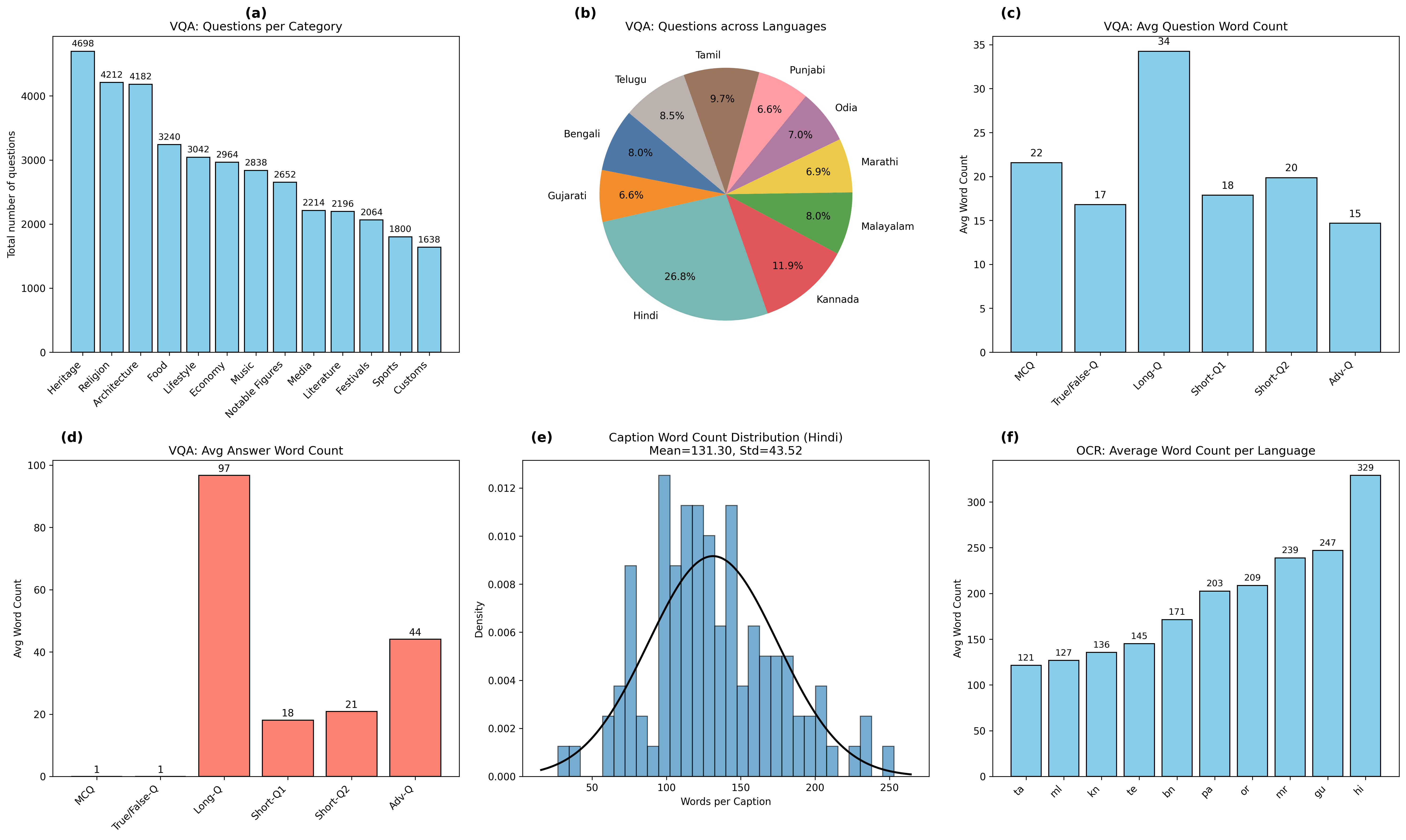}
    \caption{\textbf{Data analysis on IndicVisionBench.} Distribution of \textit{VQA} questions by category (a) and by language excluding English (b); average word counts for questions (c) and answers (d). For \textit{MMT} (e) shows caption word counts in Hindi; and for \textit{OCR} average words per language (f).}
    \label{fig:full_combined_dataset_analysis}
\end{figure}

\section{IndicVisionBench (IVB)}
\label{sec:indicvisionbench}

IndicVisionBench provides a diverse evaluation suite across 13 India-centric topics in English and 10 regional languages. Among Indic languages, Hindi dominates with 26.8\% of QA pairs (Figure \ref{fig:full_combined_dataset_analysis}). For MMT, Hindi captions average 131 words, while OCR track word counts vary more widely, with Hindi (329) and Gujarati (247) highest. Figure \ref{fig:topic_distribution_performance} shows that the dataset spans diverse cultural categories, with largest shares in \textit{Heritage (12.4\%)}, \textit{Religion (11.2\%)}, \textit{Architecture (11.1\%)} and \textit{Food (8.6\%)}. More details in Appendix~\ref{appendix:dataset}.

\paragraph{Benchmark Tracks:}
IndicVisionBench consists of three evaluation subsets: \textit{i). OCR:} 876 document images across 10 Indic languages.
\textit{ii). VQA:} 4,011 English and 1,007 multilingual culturally grounded images with 6 QA types each. We also benchmark cross-lingual performance on a disjoint set of 106 images with 6 paired questions across English and 10 Indic languages.
\textit{iii). MMT:} 106 image–caption pairs translated into 10 Indic languages, enabling multimodal translation.

\paragraph{Models Evaluated.}
We evaluate three families of VLMs with varying degrees of Indic language support: \textit{i). Proprietary models:} Gemini-2.5 Flash \citep{gemini2025flash}, GPT-4o \citep{openai2023gpt4v}. \textit{ii). Large open-weight VLMs:} Gemma-3-27B \citep{team2025gemma}, LLaMA-4-Maverick-17B (LLaMA-4 for brevity) \citep{meta2025llama4}. \textit{iii). Medium-scale open-weight VLMs (7B):} Maya \citep{Alam2024Maya}, PALO  \citep{maaz2024palo}, Pangea \citep{yue2024pangea}, and Chitrarth-1 \citep{khan2025chitrarth}. For the OCR subset, we additionally include closed-source Chitrapathak\footnote{\url{https://bit.ly/chitrapathak}}, designed specifically for Indian languages as well as Chitranuvad \citep{khan2025chitranuvad}, winning entry of the English-to-lowres\footnote{\url{https://www2.statmt.org/wmt24/multimodallowresmt-task.html}} MMT' 24 (3 Indian languages) shared task \citep{parida2024findings}.

\paragraph{Evaluation Metrics}
We assess model performance using a combination of deterministic and judgment-based metrics, tailored to each task. In the VQA track, Exact Match  (refer Table \ref{tab:metrics}) is used for multiple-choice and True/False questions, while short/long-answer and adversarial questions are evaluated using LLM-as-a-Judge (GPT-4o, 0–10 scale) following prior works~\citep{vayani2025languagesmatterevaluatinglmms} to capture contextual and cultural appropriateness (prompts in Appendix \ref{appendix:prompts}). For the 
MMT 
task, we evaluate performance using BLEU~\citep{papineni2002bleu} and RIBES~\citep{isozaki2010automatic} scores across ten Indic languages, following the setup of prior shared tasks~\citep{parida2024findings}. For OCR evaluation, we follow OCRBenchv2~\citep{fu2024ocrbench} and report Average Normalized Levenshtein Similarity (ANLS)~\citep{biten2019scene}, along with Word Error Rate (WER) and Character Error Rate (CER) as standard metrics~\citep{smith2007overview,neudecker2021survey}.  

\begin{table}[htbp]
\centering
\caption{\textbf{Model performances on English QAs in IndicVisionBench-VQA. } Average scores of different models for the six question-types. MCQ and True/False are binary (0–1), while Long Answer, Short Answer-1, Short Answer-2, and Adversarial descriptive questions use a 0–10 scale. The best score is shown in \textbf{bold}, and the second-best is \underline{underlined}.}

\small
\resizebox{\textwidth}{!}{%
\label{tab:ivb_vqa_en}
\begin{tabular}{lcccccc}
\toprule
Model & MCQ $\uparrow$ & True/False $\uparrow$ & \makecell{Long-answer $\uparrow$} & \makecell{Short-1 $\uparrow$} & \makecell{Short-2 $\uparrow$} & \makecell{Adversarial $\uparrow$} \\
\midrule
Maya & 0.69 & 0.71 & 6.98 & 5.00 & 5.50 & 0.16 \\
PALO & 0.72 & 0.43 & 7.12 & 5.51 & 5.81 & 0.19 \\
Pangea & 0.85 & 0.37 & 7.01 & 6.72 & 6.95 & 0.67 \\
Chitrarth-1 & 0.81 & 0.68 & 7.53 & 6.22 & 6.33 & 0.03 \\
\hdashline
LLaMA-4 & 0.87 & \underline{0.92} & 8.55 & 7.98 & 7.91 & 2.62 \\
Gemma-3 & 0.87 & 0.88 & 8.56 & 7.68 & 7.61 & 1.50 \\
\hdashline
GPT-4o & \underline{0.90} & 0.91 & \underline{8.75} & \underline{8.19} & \underline{8.02} & \underline{2.95} \\
Gemini-2.5 & \textbf{0.94} & \textbf{0.95} & \textbf{9.30} & \textbf{8.58} & \textbf{8.49} & \textbf{5.79} \\
\bottomrule
\end{tabular}
}
\end{table}

\section{Results}
\label{sec:results}

\textbf{VQA:} Table~\ref{tab:ivb_vqa_en} reports results on English subset of cultural VQA task. Gemini-2.5 achieves the highest scores across all 6 question types, with GPT-4o and LLaMA-4 as the strongest challengers. Binary-style questions (True/False, MCQ) yield the highest accuracy, while long-answer questions also show robust performance. Short-answer types remains harder, reflecting the difficulty of concise factual recall. This pattern highlights how answer format modulates model performance. In multilingual settings, Gemini-2.5 continues to lead overall, while LLaMA-4 and Gemma-3 exhibit comparable performance with language-specific strengths. GPT-4o consistently lags behind these models, followed by the 7B variants. Among the 7B models, Chitrarth-1 generally outperforms Pangea for short and long answer questions, with the latter holding an edge for MCQ and True/False questions. (Figure \ref{fig:topics}; Table~\ref{tab:Bilingual_results_combined} in Appendix on VQA-Indic). Adversarial questions, which embed false assumptions, remains the most challenging both in English and Indic (Tables~\ref{tab:ivb_vqa_en} and ~\ref{tab:adversarial_score_vqa}). Though Gemini-2.5 consistently outperforms all models, even its scores are notably lower compared to other QA types, reflecting the increased difficulty. On these select questions, GPT-4o is a distant second, while both Gemma-3 and LLaMA-4 struggle across the board.

\begin{figure}[htbp]
    \centering
\includegraphics[width=.99\linewidth]{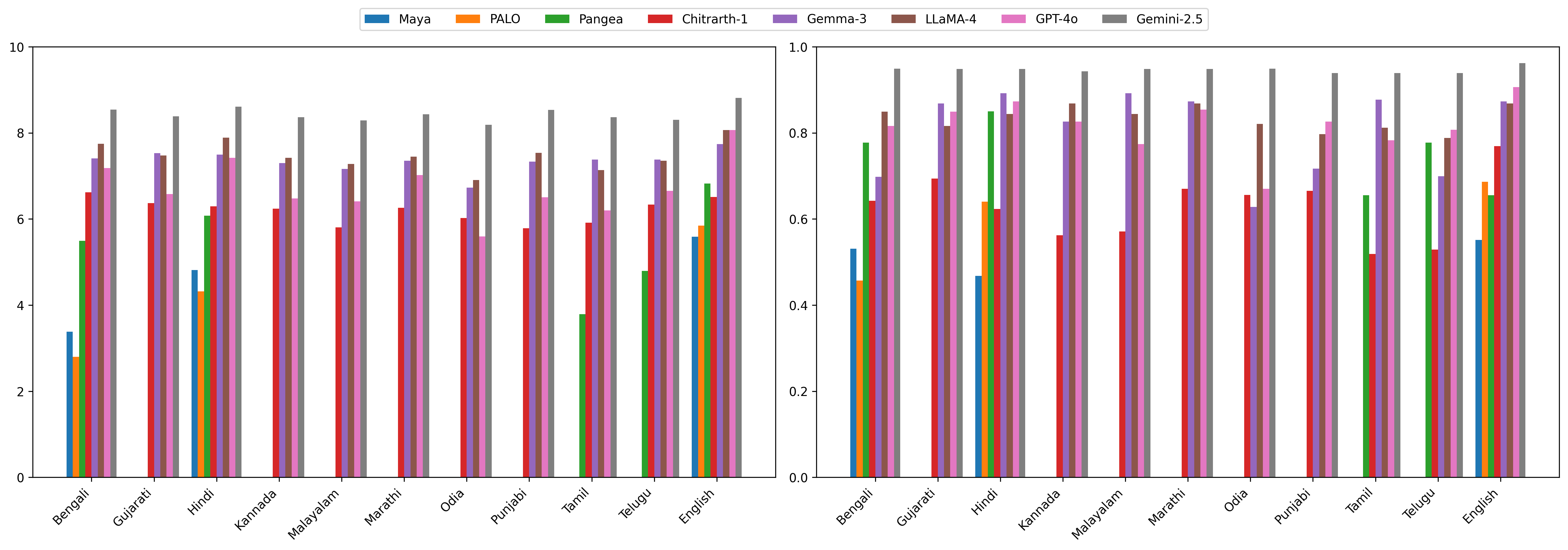} \caption{\textbf{Model performances on IndicVisionBench-VQA-Parallel.} Average scores across languages for the three open-ended (long and short) questions (on left) and scores across languages for the structured tasks (True/False and MCQ) on the right.}
    \label{fig:topics}
\end{figure}

\textbf{MMT:} Gemini-2.5 also dominates the MMT track, with LLaMA-4 and Gemma-3 performing comparable across most languages in Table ~\ref{tab:ribes_bleu_mmt} based on both BLEU and RIBES metrics. LLaMA-4 attains second-best results in Bengali, Kannada, Malayalam, Odia, and Punjabi, while Gemma-3 ranks second in the remaining languages. Malayalam proves most challenging, with the sub-par performance across all models. Chitranuvad, a finetuned version of Chitrarth-1 on Visual Genome~\citep{krishna2017visual} for image grounded translation of English into 3 languages (Hindi, Bengali and Malayalam), outperforms the base model Chitrarth-1 in Hindi, Kannada, Malayalam, and Telugu but lags in Bengali despite being specifically fine-tuned for it.
Nevertheless, both Chitranuvad and Chitrarth-1 substantially outperform other 7B baselines (Maya, PALO, Pangea).

\textbf{OCR:} We report ANLS scores in Table \ref{tab:ANLS_score}, median WER/CER  in \ref{tab:wer_cer_median}. Gemini-2.5 leads across all languages and metrics, achieving SOTA performance at both the word and character-level. OCR difficulty remains language-dependent, with higher scores for Malayalam (59.64), Odia (41.7), Telugu (33.32), and Gujarati (24.09), underscoring persistent challenges in Indic scripts. At the word level, closed-source Chitrapathak ranks second in nine languages (except Gujarati), followed closely by LLaMA-4. Surprisingly, GPT-4o performs poorly in OCR with word-level ANLS scores (e.g., 94.67 in Malayalam, 90.54 in Gujarati) significantly below expectations while 7B open-source models fall further behind. 

Across all evaluation tracks, the closed-source Gemini-2.5 demonstrates clear superiority, while Gemma-3 and LLaMA-4 show notable strengths with observed disparities across languages and question types. We show qualitative results and more details in Appendix \ref{appendix:results}. 

\begin{table*}[htbp]
\centering
\caption{\textbf{Model performances for Adversarial Questions in IndicVisionBench-VQA.} We report the average scores for only top 4 models since scores of other 7B models approached to 0. Even proprietary models perform poorly on these kinds of hard and challenging questions.}
\label{tab:adversarial_score_vqa}
\small
\resizebox{\textwidth}{!}{%
\begin{tabular}{lcccccccccccc}
\hline
Model & Bengali $\uparrow$ & English $\uparrow$ & Gujarati $\uparrow$ & Hindi $\uparrow$ & Kannada $\uparrow$ & Malayalam $\uparrow$ & Marathi $\uparrow$ & Odia $\uparrow$ & Punjabi $\uparrow$ & Tamil $\uparrow$ & Telugu $\uparrow$ \\
\hline
LLaMA-4   & 0.38 & 2.62 & 0.52 & 1.18 & 0.14 & 0.33 & 0.81 & 0.53 & 1.03 & 1.14 & 0.07 \\
\hdashline
Gemma-3   & 1.07 & 1.50 & 0.97 & 1.66 & \underline{1.02} & 0.77 & 0.68 & 0.90 & 2.94 & \underline{1.85} & 1.13 \\
GPT-4o     & \underline{2.23} & \underline{2.95} & \textbf{3.10} & \underline{2.25} & 0.67 & \underline{2.28} & \underline{2.89} & \underline{1.82} & \underline{4.00} & 1.70 & \underline{2.04} \\
Gemini-2.5  & \textbf{5.17} & \textbf{5.79} & \underline{2.94} & \textbf{4.46} & \textbf{3.17} & \textbf{3.32} & \textbf{4.84} & \textbf{3.92} & \textbf{5.71} & \textbf{5.15} & \textbf{2.73} \\
\hline
\end{tabular}
}

\end{table*}

\begin{table}[htbp]
\centering
\caption{\textbf{Model performances on IndicVisionBench-MMT.} RIBES (R) and BLEU (B) scores across ten Indic languages, with Gemini-2.5 achieving the highest performance consistently.}
\label{tab:ribes_bleu_mmt}
\resizebox{\textwidth}{!}{
\begin{tabular}{
  l
  *{10}{S[table-format=1.2]S[table-format=2.2]}
}
\toprule
\multirow{2}{*}{\textbf{Model}} &
\multicolumn{2}{c}{\textbf{Bengali}} &
\multicolumn{2}{c}{\textbf{Gujarati}} &
\multicolumn{2}{c}{\textbf{Hindi}} &
\multicolumn{2}{c}{\textbf{Kannada}} &
\multicolumn{2}{c}{\textbf{Malayalam}} &
\multicolumn{2}{c}{\textbf{Marathi}} &
\multicolumn{2}{c}{\textbf{Odia}} &
\multicolumn{2}{c}{\textbf{Punjabi}} &
\multicolumn{2}{c}{\textbf{Tamil}} &
\multicolumn{2}{c}{\textbf{Telugu}} \\
\cmidrule(lr){2-3}
\cmidrule(lr){4-5}
\cmidrule(lr){6-7}
\cmidrule(lr){8-9}
\cmidrule(lr){10-11}
\cmidrule(lr){12-13}
\cmidrule(lr){14-15}
\cmidrule(lr){16-17}
\cmidrule(lr){18-19}
\cmidrule(lr){20-21}
& \textbf{R $\uparrow$} & \textbf{B $\uparrow$}
& \textbf{R $\uparrow$} & \textbf{B $\uparrow$}
& \textbf{R $\uparrow$} & \textbf{B $\uparrow$}
& \textbf{R $\uparrow$} & \textbf{B $\uparrow$}
& \textbf{R $\uparrow$} & \textbf{B $\uparrow$}
& \textbf{R $\uparrow$} & \textbf{B $\uparrow$}
& \textbf{R $\uparrow$} & \textbf{B $\uparrow$}
& \textbf{R $\uparrow$} & \textbf{B $\uparrow$}
& \textbf{R $\uparrow$} & \textbf{B $\uparrow$}
& \textbf{R $\uparrow$} & \textbf{B $\uparrow$} \\
\midrule
Maya        & 0.45 & 5.48  & \multicolumn{1}{c}{--} & \multicolumn{1}{c}{--} & 0.69 & 18.09 & \multicolumn{1}{c}{--} & \multicolumn{1}{c}{--} & \multicolumn{1}{c}{--} & \multicolumn{1}{c}{--} & \multicolumn{1}{c}{--} & \multicolumn{1}{c}{--} & \multicolumn{1}{c}{--} & \multicolumn{1}{c}{--} & \multicolumn{1}{c}{--} & \multicolumn{1}{c}{--} & \multicolumn{1}{c}{--} & \multicolumn{1}{c}{--} & \multicolumn{1}{c}{--} & \multicolumn{1}{c}{--} \\
PALO        & 0.41 & 4.56  & \multicolumn{1}{c}{--} & \multicolumn{1}{c}{--} & 0.58 & 11.79 & \multicolumn{1}{c}{--} & \multicolumn{1}{c}{--} & \multicolumn{1}{c}{--} & \multicolumn{1}{c}{--} & \multicolumn{1}{c}{--} & \multicolumn{1}{c}{--} & \multicolumn{1}{c}{--} & \multicolumn{1}{c}{--} & \multicolumn{1}{c}{--} & \multicolumn{1}{c}{--} & \multicolumn{1}{c}{--} & \multicolumn{1}{c}{--} & \multicolumn{1}{c}{--} & \multicolumn{1}{c}{--} \\
Pangea      & 0.69 & 16.84  & \multicolumn{1}{c}{--} & \multicolumn{1}{c}{--} & 0.75 & 25.29 & \multicolumn{1}{c}{--} & \multicolumn{1}{c}{--} & \multicolumn{1}{c}{--} & \multicolumn{1}{c}{--} & \multicolumn{1}{c}{--} & \multicolumn{1}{c}{--} & \multicolumn{1}{c}{--} & \multicolumn{1}{c}{--} & \multicolumn{1}{c}{--} & \multicolumn{1}{c}{--} & 0.43 & 5.4  & 0.62 & 12.52 \\
Chitrarth-1   & 0.76 & 21.89 & 0.72 & 21.07 & 0.71 & 21.93 & 0.65 & 12.83 & 0.59 & 7.49  & 0.70 & 16.25 & 0.62 & 11.10 & 0.50 & 10.39 & 0.71 & 17.59 & 0.67 & 15.60 \\
Chitranuvad & 0.74 & 18.13 & 0.68 & 18.66 & 0.74 & 21.93 & 0.69 & 12.93 & 0.60 & 7.36  & 0.69 & 14.74 & 0.03 & 0.86  & 0.07 & 1.61  & 0.67 & 15.85 & 0.71 & 16.56 \\
\hdashline
LLaMA-4       & \underline{0.82} & \underline{30.70} & 0.80 & 29.84 & 0.81 & 33.55 & \underline{0.76} & \underline{20.91} & \underline{0.72} & \underline{14.96} & 0.76 & 20.49 & \underline{0.72} & \underline{15.35} & \underline{0.85} & \underline{41.01} & 0.80 & 25.22 & 0.78 & 22.35 \\
Gemma-3       & 0.81 & 29.75 & \underline{0.83} & \underline{35.76} & \underline{0.82} & \underline{34.40} & 0.72 & 16.23 & 0.68 & 10.29 & \underline{0.80} & \underline{26.96} & 0.65 & 8.56  & 0.81 & 32.48 & \underline{0.82} & \underline{29.97} & \underline{0.82} & \underline{31.35} \\
\hdashline
GPT-4o         & 0.80 & 28.65 & 0.74 & 21.99 & 0.79 & 33.30 & 0.67 & 11.75 & 0.59 & 8.08  & 0.75 & 23.19 & 0.65 & 9.42  & 0.75 & 24.72 & 0.73 & 16.77 & 0.71 & 17.65 \\
Gemini-2.5      & \textbf{0.87} & \textbf{44.51} & \textbf{0.90} & \textbf{53.27} & \textbf{0.83} & \textbf{38.91} & \textbf{0.80} & \textbf{30.08} & \textbf{0.81} & \textbf{28.65} & \textbf{0.88} & \textbf{47.00} & \textbf{0.85} & \textbf{39.08} & \textbf{0.89} & \textbf{52.39} & \textbf{0.88} & \textbf{46.32} & \textbf{0.87} & \textbf{44.85} \\
\bottomrule
\end{tabular}
}
\end{table}

\begin{table}[htbp] \centering\caption{\textbf{Model performances on IndicVisionBench-OCR}: ANLS (Average Normalized Levenshtein Similarity) across 10 Indic languages for different models. ANLS-W and ANLS-C denote word- and character-level scores, respectively. For each language, the highest score is marked in \textbf{bold}, while the second-highest is \underline{underlined}. Gemini-2.5 performs the best followed by Chitrapathak in most languages.}
\label{tab:ANLS_score}
\resizebox{\textwidth}{!}{
\begin{tabular}{l*{11}{rr}}
\toprule
\multirow{2}{*}{Model} & \multicolumn{2}{c}{Bengali} & \multicolumn{2}{c}{Gujarati} & \multicolumn{2}{c}{Hindi} & \multicolumn{2}{c}{Kannada} & \multicolumn{2}{c}{Malayalam} & \multicolumn{2}{c}{Marathi} & \multicolumn{2}{c}{Odia} & \multicolumn{2}{c}{Punjabi} & \multicolumn{2}{c}{Tamil} & \multicolumn{2}{c}{Telugu} \\
\cmidrule(lr){2-3} \cmidrule(lr){4-5} \cmidrule(lr){6-7} \cmidrule(lr){8-9} \cmidrule(lr){10-11} \cmidrule(lr){12-13} \cmidrule(lr){14-15} \cmidrule(lr){16-17} \cmidrule(lr){18-19} \cmidrule(lr){20-21}
& Word $\downarrow$ & Char $\downarrow$ & Word $\downarrow$ & Char $\downarrow$ & Word $\downarrow$ & Char $\downarrow$ & Word $\downarrow$ & Char $\downarrow$ & Word $\downarrow$ & Char $\downarrow$ & Word $\downarrow$ & Char $\downarrow$ & Word $\downarrow$ & Char $\downarrow$ & Word $\downarrow$ & Char $\downarrow$ & Word $\downarrow$ & Char $\downarrow$ & Word $\downarrow$ & Char $\downarrow$ \\
\midrule
Maya & 99.42 & 95.77 & - & - & 99.70 & 94.91 & - & - & - & - & - & - & - & - & - & - & - & - & - & - \\
PALO & 96.30 & 91.15 & - & - & 99.26 & 91.98 & - & - & - & - & - & - & - & - & - & - & - & - & - & - \\
Pangea & 94.66 & 80.33 & - & - & 99.53 & 91.50 & - & - & - & - & - & - & - & - & - & - & 99.44 & 84.13 & 99.95 & 89.91 \\
Chitrarth-1  & 96.16 & 84.65 & 99.32 & 86.81 & 98.56 & 89.81 & 99.58 & 85.29 & 99.62 & 94.77 & 99.66 & 86.58 & 99.99 & 93.21 & 99.16 & 90.17 & 99.10 & 89.94 & 99.86 & 89.02 \\
\hdashline
LLaMA-4  & 31.52 & 13.21 & \underline{40.56} & \underline{18.38} & 25.73 & \underline{11.91} & 36.90 & 11.17 & 75.50 & \underline{45.75} & 20.94 & 8.05 & 97.51 & 86.78 & 29.77 & 12.68 & 31.36 & 10.79 & 57.07 & 18.72 \\
Gemma-3 & 42.15 & 24.41 & 60.07 & 38.49 & 46.47 & 29.50 & 84.22 & 54.24 & 92.06 & 72.64 & 50.40 & 31.06 & 92.67 & 70.72 & 70.88 & 42.65 & 39.52 & 16.51 & 86.76 & 54.14 \\
\hdashline
Chitrapathak & \underline{17.14} & \underline{7.03} & 49.99 & 27.80 & \underline{25.55} & 13.74 & \underline{26.24} & \underline{8.78} & \underline{71.97} & 48.19 & \underline{15.68} & \underline{6.09} & \underline{50.72} & \underline{31.62} & \underline{17.70} & \underline{7.87} & \underline{19.25} & \underline{5.81} & \underline{38.79} & \underline{11.00} \\
GPT-4o  & 55.51 & 32.68 & 90.54 & 68.03 & 54.62 & 35.54 & 94.33 & 69.79 & 94.67 & 78.47 & 63.44 & 37.93 & 94.61 & 73.46 & 68.88 & 40.71 & 74.35 & 43.39 & 95.97 & 70.08 \\
Gemini-2.5 & \textbf{11.30} & \textbf{4.04} & \textbf{24.09} & \textbf{7.61} & \textbf{16.01} & \textbf{5.88} & \textbf{17.18} & \textbf{4.38} & \textbf{59.64} & \textbf{30.60} & \textbf{8.06} & \textbf{1.79} & \textbf{41.70} & \textbf{18.60} & \textbf{14.56} & \textbf{4.98} & \textbf{15.26} & \textbf{3.01} & \textbf{33.32} & \textbf{7.16} \\
\bottomrule
\end{tabular}
} 
\end{table}

\begin{table}[htbp]
\centering
\scriptsize
\caption{\textbf{Are images necessary for IndicVisionBench-VQA-Parallel?}  
Average performance drop in short-answer questions across languages for Chitrarth-1, Gemma-3, and Gemini-2.5, comparing with vs. without image input.}
\label{tab:parallel_corpus_without_image_short_q1}

\resizebox{\textwidth}{!}{\begin{tabular}{l l *{11}{S[table-format=3.2]}}
\toprule
Model & Type & \multicolumn{1}{c}{Bengali} $\uparrow$ & \multicolumn{1}{c}{English} $\uparrow$ & \multicolumn{1}{c}{Gujarati} $\uparrow$ & \multicolumn{1}{c}{Hindi} $\uparrow$ & \multicolumn{1}{c}{Kannada} $\uparrow$ & \multicolumn{1}{c}{Malayalam} $\uparrow$ & \multicolumn{1}{c}{Marathi} $\uparrow$ & \multicolumn{1}{c}{Odia} $\uparrow$ & \multicolumn{1}{c}{Punjabi} $\uparrow$ & \multicolumn{1}{c}{Tamil} $\uparrow$ & \multicolumn{1}{c}{Telugu} $\uparrow$ \\
\midrule
Chitrarth-1 & w/o img & 3.88 & 4.18 & 3.76 & 4.09 & 4.07 & 3.99 & 4.53 & 4.06 & 4.52 & 3.88 & 4.23 \\
 & with img & 5.90 & 5.95 & 5.76 & 5.97 & 5.58 & 4.68 & 5.61 & 5.11 & 5.43 & 4.93 & 5.50 \\
\hdashline
Gemma-3 & w/o img & 4.21 & 3.25 & 4.30 & 4.47 & 3.90 & 4.23 & 4.31 & 3.54 & 4.15 & 4.26 & 4.44 \\
 & with img & 6.67 & 6.98 & 7.08 & 6.87 & 6.29 & 6.41 & 6.58 & 5.94 & 6.80 & 6.92 & 6.93 \\
\hdashline
Gemini-2.5 & w/o img & 4.69 & 4.14 & 4.62 & 4.76 & 4.29 & 4.69 & 4.57 & 4.80 & 4.60 & 4.24 & 4.66 \\
 & with img & 8.09 & 8.22 & 7.90 & 8.33 & 7.57 & 7.89 & 7.99 & 7.72 & 8.15 & 7.96 & 7.76 \\
\bottomrule
\end{tabular}
}

\end{table}

\section{Discussion}

\paragraph{VLMs without vision: Are images necessary?}
We evaluate models on the paired \textit{VQA-Parallel} corpus spanning 10 Indic languages plus English, comparing performance with and without visual input. Removing images leads to a substantial drop in accuracy, most pronounced for short-answer tasks (see Table \ref{tab:parallel_corpus_without_image_short_q1}) where precise, detail-oriented responses are required. Long-answer questions are comparatively more resilient, though still affected. Across the representative models of each category: Chitrarth-1, Gemma-3, and Gemini-2.5, the  trend is consistent (Table ~\ref{tab:parallel_corpus_without_image_long_answer} in Appendix), showcasing that visual grounding is necessary for answering questions in our VQA benchmark.  

\paragraph{Do VLMs exhibit cross-lingual variations in performance?}
We systematically conducted a study on the \textit{VQA-Parallel} corpus to measure the cross-lingual performance across 11 languages including English. For the long answer, Gemini-2.5 achieves the best overall performance, followed by Gemma-3, which ranks second in all languages except Odia, where it surpasses even GPT-4o. On the MCQ type questions, GPT-4o and LLaMA-4 perform comparable across all languages (next to Gemini-2.5) as in Table ~\ref{tab:parallel_results_combined}, while Chitrarth-1 consistently outperforms all other 7B-scale models. Among the 7B-scale models, Chitrarth-1 again proves strongest, followed by Maya. In the adversarial question, Gemini remains the best-performing model but shows a considerable drop in performance compared to its long-form results. Excluding English, Gemma-3 consistently outperforms both GPT-4o and LLaMA-4 across all Indian languages, while LLaMA-4 maintains a slight advantage over Gemma-3 in English. For short-answer questions, LLaMA-4 generally secures the second rank, but falls behind Gemma-3 in Tamil and Telugu, and performs particularly poorly in Gujarati and Kannada. Nonetheless, both LLaMA-4 and Gemma-3 outperform GPT-4o across all languages. In the True/False setting, GPT-4o ranks second in Bengali, Punjabi, Telugu, and English. By contrast, Gemma-3 shows notable weaknesses in Bengali, Punjabi, and Telugu, even trailing behind Chitrarth-1 in Punjabi and Bengali.

\begin{figure}[htbp]
    \centering
    \includegraphics[width=\linewidth]{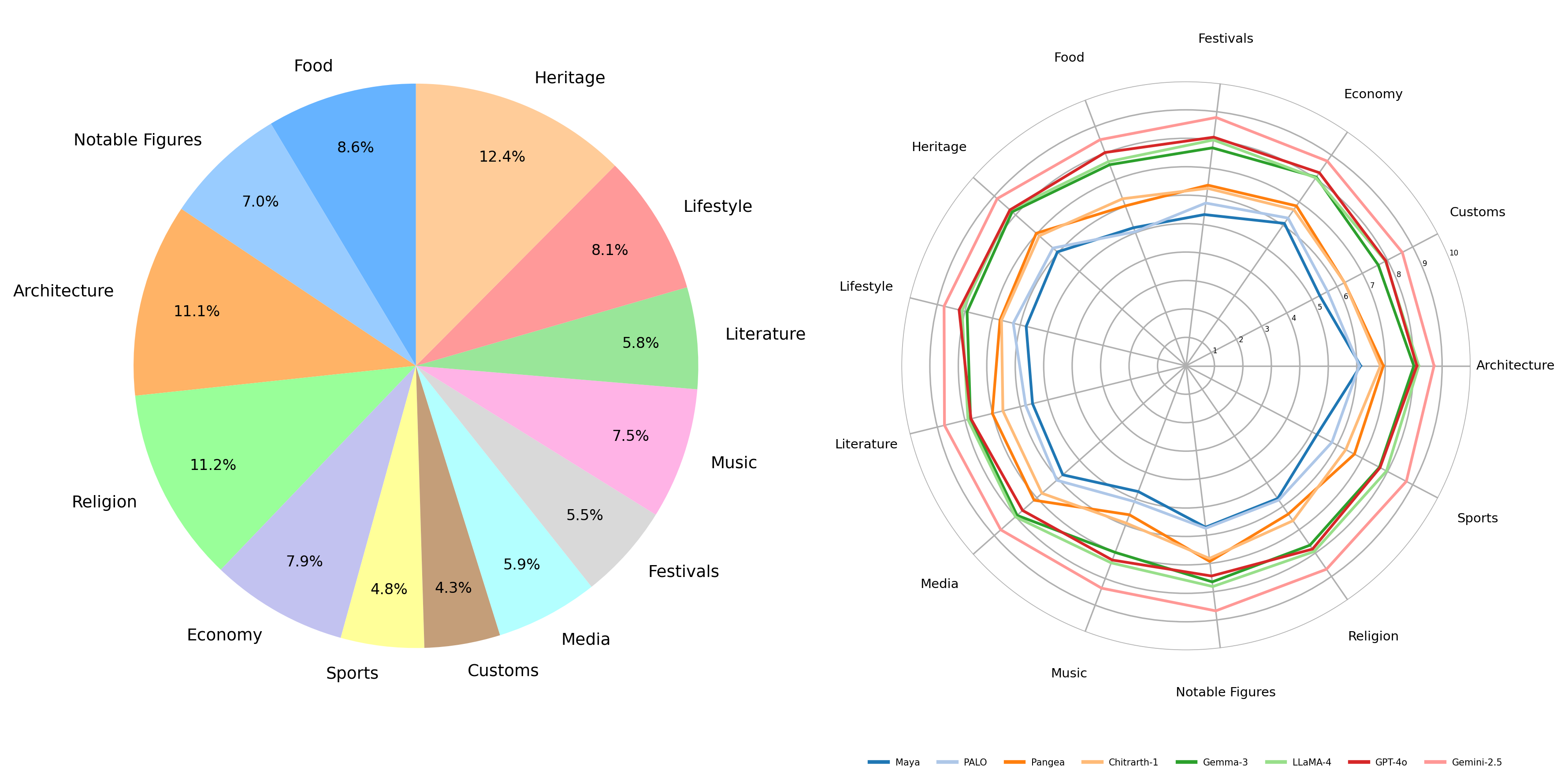}
    \caption{\textbf{Performance across topics in IndicVisionBench-VQA.} Distribution of categories of questions (on left) and  model performances averaged over the two short and a long answer open-ended questions (on right). Gemini-2.5 shows comparable performance across all topics.}
    \label{fig:topic_distribution_performance}
\end{figure}

\paragraph{Do VLMs perform better in some cultural topics in English?}
Model performance also varies by cultural category. As shown in Figure~\ref{fig:topic_distribution_performance}, Gemini-2.5 consistently achieves the strongest results across topics with slight variations, establishing itself as the most reliable model. LLaMA-4 and Gemma-3 show advantages on certain topics, while GPT-4o retains a slight edge in others over both. Among 7B models, Chitrarth-1 and Pangea demonstrate moderate and roughly comparable capabilities, whereas Maya and PALO cluster together at the lower end. These topics-level patterns suggest that stronger models generalize more evenly across cultural domains, while weaker ones exhibit sharper inconsistencies.

\begin{figure*}[htbp]
    \centering
    \begin{minipage}[t]{0.48\textwidth}
        \centering
        \includegraphics[width=\linewidth]{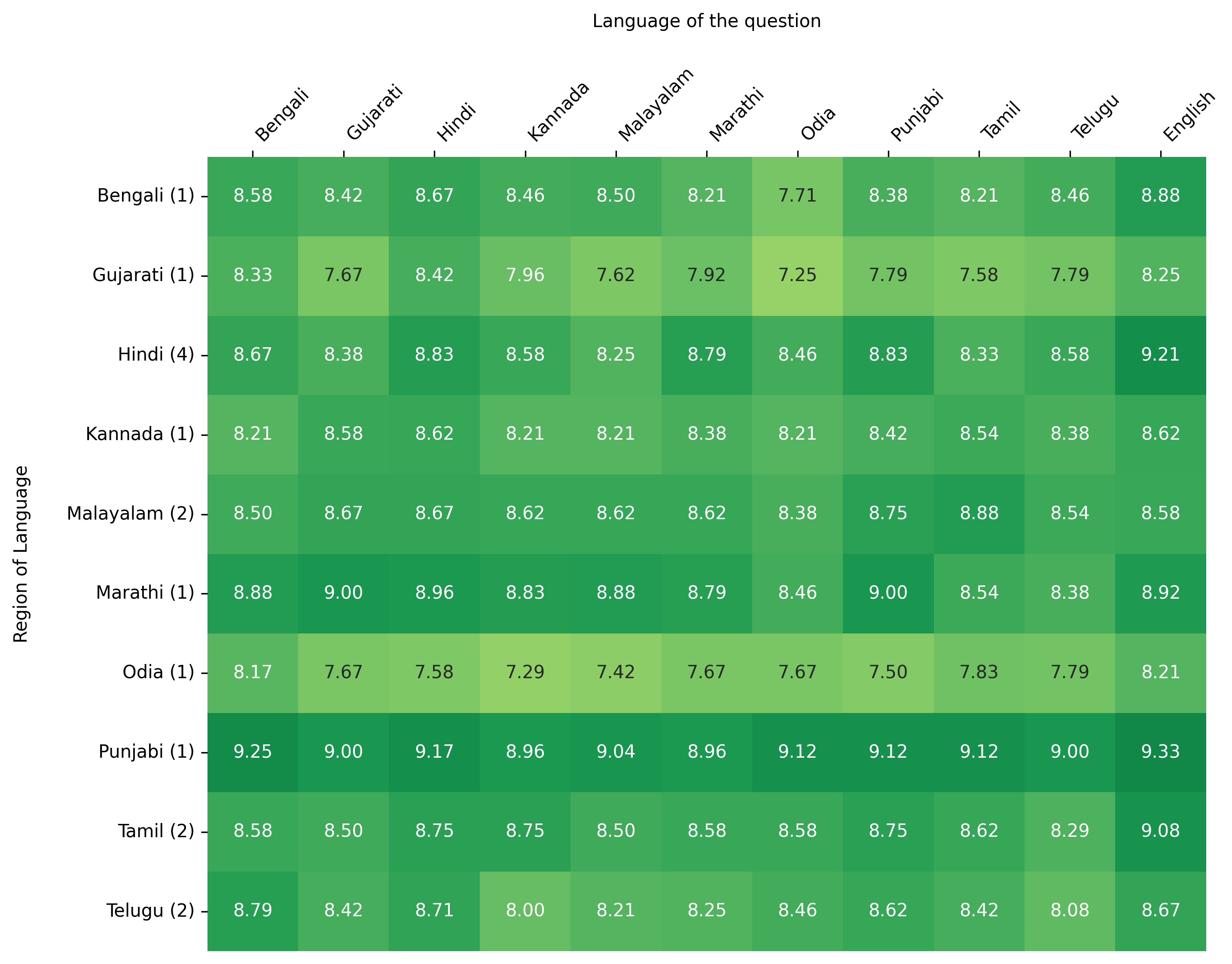}
    \end{minipage}%
    \hfill
    \begin{minipage}[t]{0.48\textwidth}
        \centering        \includegraphics[width=\linewidth]{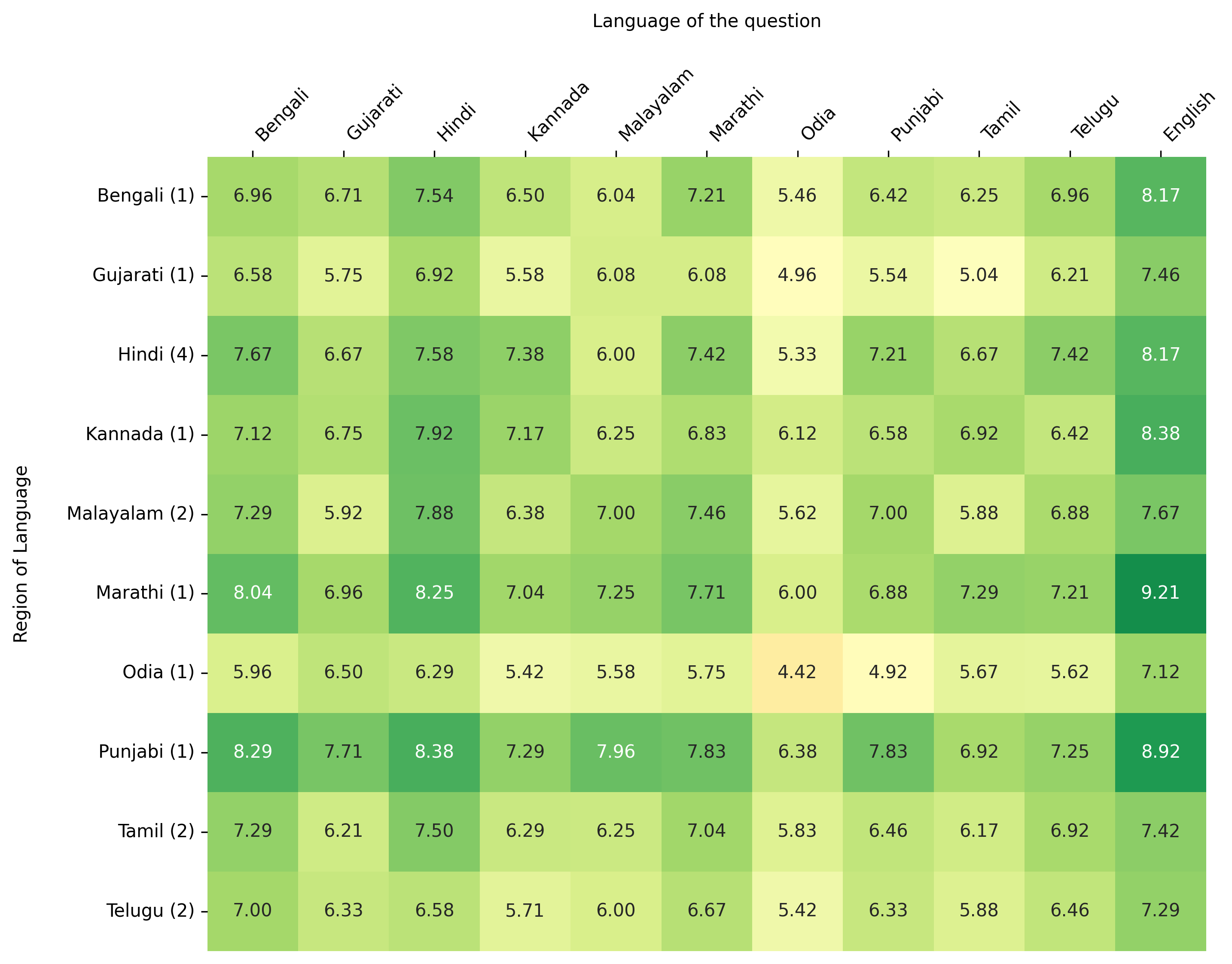}
    \end{minipage}
    \caption{ \textbf{Average performance on open-ended question (Long and Short answer types).} Gemini-2.5 (on left), GPT-4o (right) and other models in Figure \ref{fig:heatmaps_parallel_corpus_remaining_six} in Appendix. X-axis displays query languages and  Y-axis displays Indian states grouped by dominant language. Numbers in parentheses indicate the count of states for dominant language.}
    \label{fig:heatmaps_parallel_corpus_gemini_gpt}
\end{figure*}

\paragraph{Do VLMs know more about certain regions or show regional-language biases?} 
We investigate whether VLMs exhibit region-specific strengths or biases by comparing performance across cultural images from different Indian regions (states) and the corresponding multilingual queries. Gemini-2.5 (Figure \ref{fig:heatmaps_parallel_corpus_gemini_gpt}) generally performs well across regions but consistently struggles with Odia cultural content, regardless of query language. Across models, English questions yield the best results, followed by Hindi and Bengali, with no clear alignment between the region depicted in the image and the language of the query. Among open-weight models (Figure \ref{fig:heatmaps_parallel_corpus_remaining_six} in Appendix), Gemma-3 favors Punjabi, Marathi, and Bengali, while LLaMA-4 performs best on Punjabi and Hindi content. 7B Chitrarth-1 records its lowest scores on Odia and Punjabi and often performs better in English than in native Indic languages for Hindi-speaking states. Pangea performs strongest in English and weakest in Tamil, with also a bias toward Tamil-language queries. Maya and PALO remain relatively stable in English but show weaknesses in Hindi and Bengali, respectively. These results suggest that while certain region–language preferences exist, systematic region-level cultural alignment is largely absent, with Odia emerging as a consistently difficult case across all models.

\paragraph{How do we evaluate OCR outputs?} Apart from the ANLS metric, we also report average and median WER/CER metrics (Tables \ref{tab:wer_cer_part1_combined} and \ref{tab:wer_cer_part2_combined} in the Appendix), based on Levenshtein Distance \citep{lcvenshtcin1966binary}. 
Mathematically, this metric is unbounded and can over-penalize models for a few extreme cases (e.g., inflated scores for LLaMA-4 repetition upto maximum length; see Figure \ref{fig:malayalam_issue}). To quantify this effect, we report in Table \ref{tab:wer_cer_summary}, the proportion of instances exceeding a value of 1. Notably, LLaMA-4 accounts for only 7\% of such cases, yet includes strong outliers with an average worst-case WER of 25 in Malayalam, while still ranking third best under ANLS which remains relatively robust to these anomalies.
While underexplored for LLMs and VLMs, that often produce long repetitive outputs \citep{hiraoka2024repetition}, even median-based reporting \citep{patel2025evaluate} fails to capture such edge cases. Other statistical alternatives like Word Recognition Rate (WRR) / Character Recognition Rate (CRR) \citep{bhattacharyya2025adapting} ignore ordering in the outputs, so we adopt ANLS as the most interpretable metric in our setting. 

\section{Conclusion}
\label{sec:conclusion}
We present IndicVisionBench, a large-scale benchmark consisting of ~5K unique images, 37K+ questions spanning 13 culturally grounded topics across English and 10 Indic languages. Covering VQA (6 kind of questions), OCR, and MMT tasks, it combines curated images with linguistically diverse queries to probe recognition, reasoning, and translation. Experiments with proprietary and open-weight models reveal substantial performance gaps, especially in low-resource languages and culturally nuanced settings. By centering cultural and linguistic diversity, our work provides a reproducible foundation for building more inclusive and globally robust multimodal systems.

\section*{Ethics and Reproducibility Statement}
\paragraph{Ethics Statement}

This work focuses on the responsible development of an evaluation benchmark for multimodal cultural understanding in regional Indian contexts and languages, spanning diverse tasks. We applied careful filtering to reduce harmful or unsafe content, though model outputs remain beyond our full control. All external datasets and tools are properly cited. Human involvement was limited to annotation and quality control; no sensitive or personally identifiable information (PII) was collected. Participants were informed that their contributed images and annotations would be used in a VLM benchmark and provided prior consent, with instructions to obscure any identifiable information. Dataset curation was performed by a team of in-house annotators who were fairly compensated according to local market standards. As the study did not involve personal or medical data, formal IRB approval was not required. Throughout the process, we prioritized preserving cultural nuance while minimizing bias and harm. Despite careful filtering, dataset bias may remain, reflecting regional, socio-economic, or cultural imbalances. The resulting benchmark aims to support the development of multilingual and culturally inclusive vision-language models.

\paragraph{Reproducibility Statement}
To support reproducibility, we will release all benchmark-related artifacts publicly, along with detailed documentation. Our experimental setups and evaluation protocols are thoroughly recorded to facilitate precise replication of results. For components involving human annotation or judgment, we include the instructions and guidelines followed, ensuring transparency and consistency throughout the process.

\section*{Acknowledgments}
We express our sincere gratitude to the leadership at Krutrim for their unwavering support throughout the course of this research. We would also like to thank the AI Research team at Krutrim for their valuable feedback and insightful discussions during various stages of the project, as well as the Krutrim team members who contributed to data collection. We further acknowledge the dedicated efforts of the data collection and annotation teams, including Sanmathi and Aravind, for their work in building this benchmark. Our experiments were conducted with generous computational support from Krutrim Cloud using Krutrim credits.

\bibliography{iclr2025_conference}
\bibliographystyle{iclr2025_conference}

\newpage

\appendix
\section*{Appendix}
\label{appendix}

\section{Limitations}
\label{appendix:limitations}

IndicVisionBench covers English and ten medium- to low-resource Indic languages across 13 culturally grounded topics, but some limitations remain. Language coverage is still limited for some of the lowest-resource languages, and topic diversity could be further expanded to cover additional cultural contexts. Human annotations and translations are usually subject to interpretation, especially for cultural nuances, which may introduce inconsistencies or simplifications. Existing evaluation metrics and LLM-based scoring may not fully capture cultural grounding or multimodal reasoning, highlighting opportunities for more nuanced evaluation approaches. Although we systematically study the impact of visual modality for VQA, we have not yet explored this effect for the MMT track with prior works~\citep{gronroos2018memad,lala2018sheffield,wu-etal-2021-good} showcasing minimal impact of visual modality. We plan to cover this as part of the future work.

\section{Implementation}
\label{appendix:implementation}

Most of our implementation is based in Python and we use HuggingFace library in PyTorch \citep{paszke2019pytorch,wolf2019huggingface}. We evaluate a broad range of frontier and open-source models on the IndicVisionBench benchmark using their respective APIs as well as HuggingFace pages. For the OCR-focused subset of our evaluation, we used the implementation of OCRBenchV2\footnote{\url{https://github.com/Yuliang-Liu/MultimodalOCR/blob/main/OCRBench_v2/eval_scripts/vqa_metric.py}} to compute the ANLS metrics. We adapted this script to report both word-level and character-level ANLS metrics. Beyond OCR, our full benchmark includes additional evaluation dimensions,  we also compute metrics such as BLEU\footnote{\url{https://huggingface.co/spaces/evaluate-metric/sacrebleu/blob/main/sacrebleu.py}} where we don't have additional pre-processing step of tokenization, RIBES\footnote{\url{https://github.com/nttcslab-nlp/RIBES}}, Exact Match (EM), and employ LLM-as-a-Judge evaluation to capture semantic and task-level correctness. We use official Batch APIs to get the results from GPT-4o\footnote{\url{https://platform.openai.com/docs/guides/batch}} for LLM-as-a-judge\footnote{GPT-4o-2024-08-06 is the model used as LLM-as-a-Judge.} results and Gemini-2.5 for annotations. 

\textbf{OCR benchmark.} We began with raw dumps from Wikisource\footnote{\url{https://hi.wikisource.org}} for ten Indic languages, obtained from the official Wikimedia snapshots \footnote{\url{https://dumps.wikimedia.org/hiwikisource/latest/}}. Each compressed XML dump was parsed to extract page titles, which were then converted into canonical Wikisource URLs. From the harvested URLs, we fetched the corresponding Wikisource pages, retained only pages with images and applied filtering to retain only those marked as proofread by the community (quality level 4), ensuring high-fidelity ground truth. For every verified page, we used the $<$prp-page-image$>$ tag to collect the page scans and the $<$pagetext$>$ tag to extract the corresponding OCR text, which reflects the latest human-edited annotation. This pipeline, implemented with custom parsing code and filtering, yielded a linguistically diverse dataset of high-quality scanned documents paired with verified text, which forms the foundation of the OCR evaluation track in IndicVisionBench. 

\begin{table}[htbp]
\centering

\caption{\textbf{Cost comparison for Gemini and OpenAI models in  OCR and Captioning tasks (rounded to 2 decimals in \$)}. We provide an approximate cost at the time of submission for a sample of 1000 images based on the assumptions of input and outputs tokens. Batch APIs are half the price of Single calls. Here Gemini-2.5-F denotes Gemini-2.5-Flash and Gemini-2.5-P denotes Gemini-2.5-Pro. Our benchmark further involved a multiple for number of languages and questions.}
\label{appendix:cost_comparison}

\resizebox{\textwidth}{!}{
\begin{tabular}{lccccccccccc}
\hline
\textbf{Task} & \textbf{\#Images} & \textbf{Input (M)} & \textbf{Output (M)} &
\multicolumn{2}{c}{\textbf{Batch (Gemini)}} &
\multicolumn{2}{c}{\textbf{Single (Gemini)}} &
\multicolumn{2}{c}{\textbf{Batch (OpenAI)}} &
\multicolumn{2}{c}{\textbf{Single (OpenAI)}} \\
\cline{5-12}
& & & & \textbf{Gemini-2.5-F} & \textbf{Gemini-2.5-P} &
\textbf{Gemini-2.5-F} & \textbf{Gemini-2.5-P} &

\textbf{GPT-4o} & \textbf{GPT-4o-mini} &
\textbf{GPT-4o} & \textbf{GPT-4o-mini} \\
\hline
OCR & 1,000 & 0.05 & 0.30 & 0.58 & 2.34 & 1.15 & 4.68 & 2.52 & 2.01 & 5.04 & 4.01 \\
Captioning & 1,000 & 0.05 & 0.15 & 0.39 & 1.59 & 0.78 & 3.18 & 1.77 & 1.96 & 3.54 & 3.92 \\
\hline
\end{tabular}
}
\end{table}

\clearpage

\section{Additional Results}
\label{appendix:results}

\begin{figure*}[htbp]
    \centering
    \begin{minipage}[t]{0.48\textwidth}
        \centering
        \includegraphics[width=\linewidth]{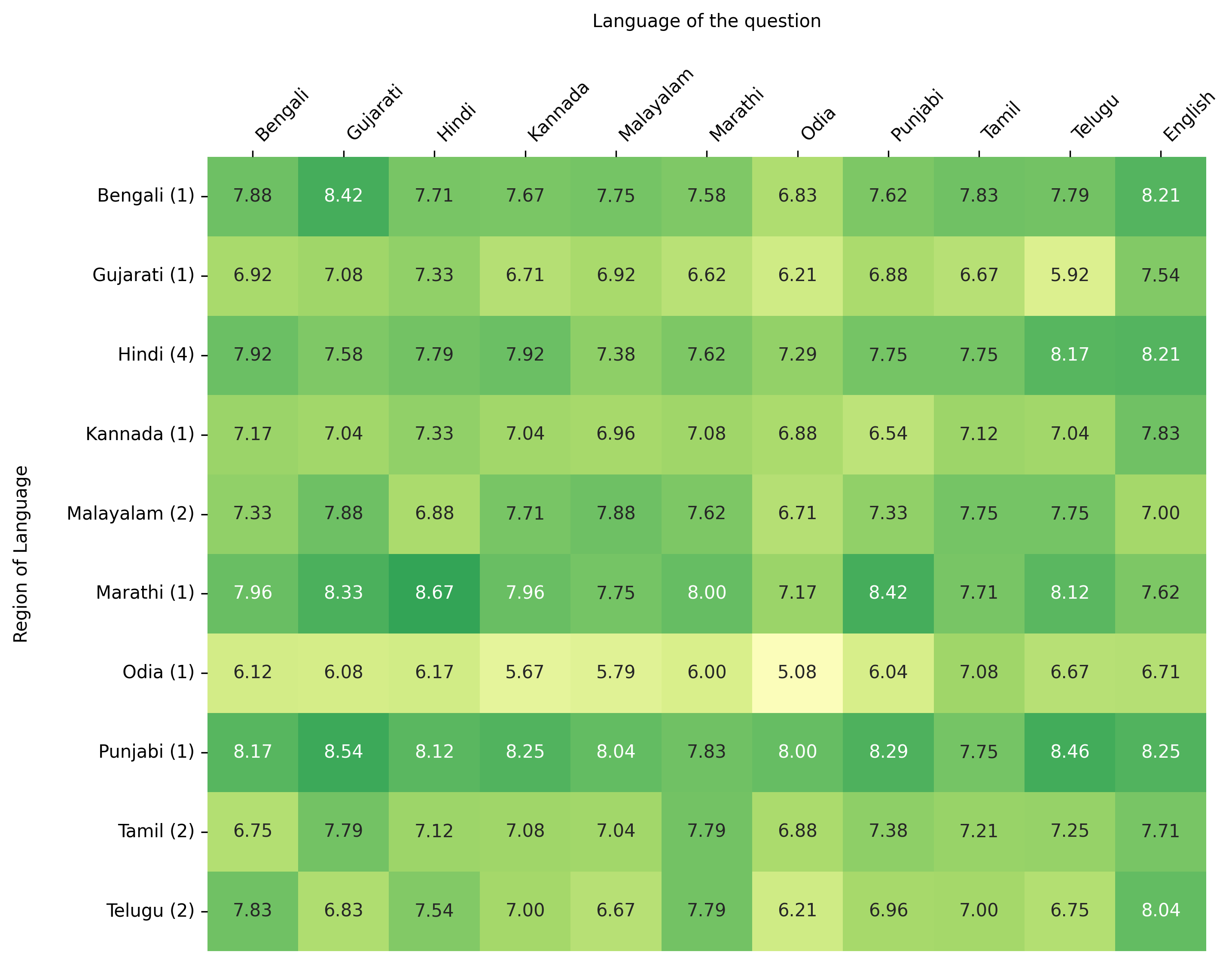}
    \end{minipage}%
    \hfill
    \begin{minipage}[t]{0.48\textwidth}
        \centering
        \includegraphics[width=\linewidth]{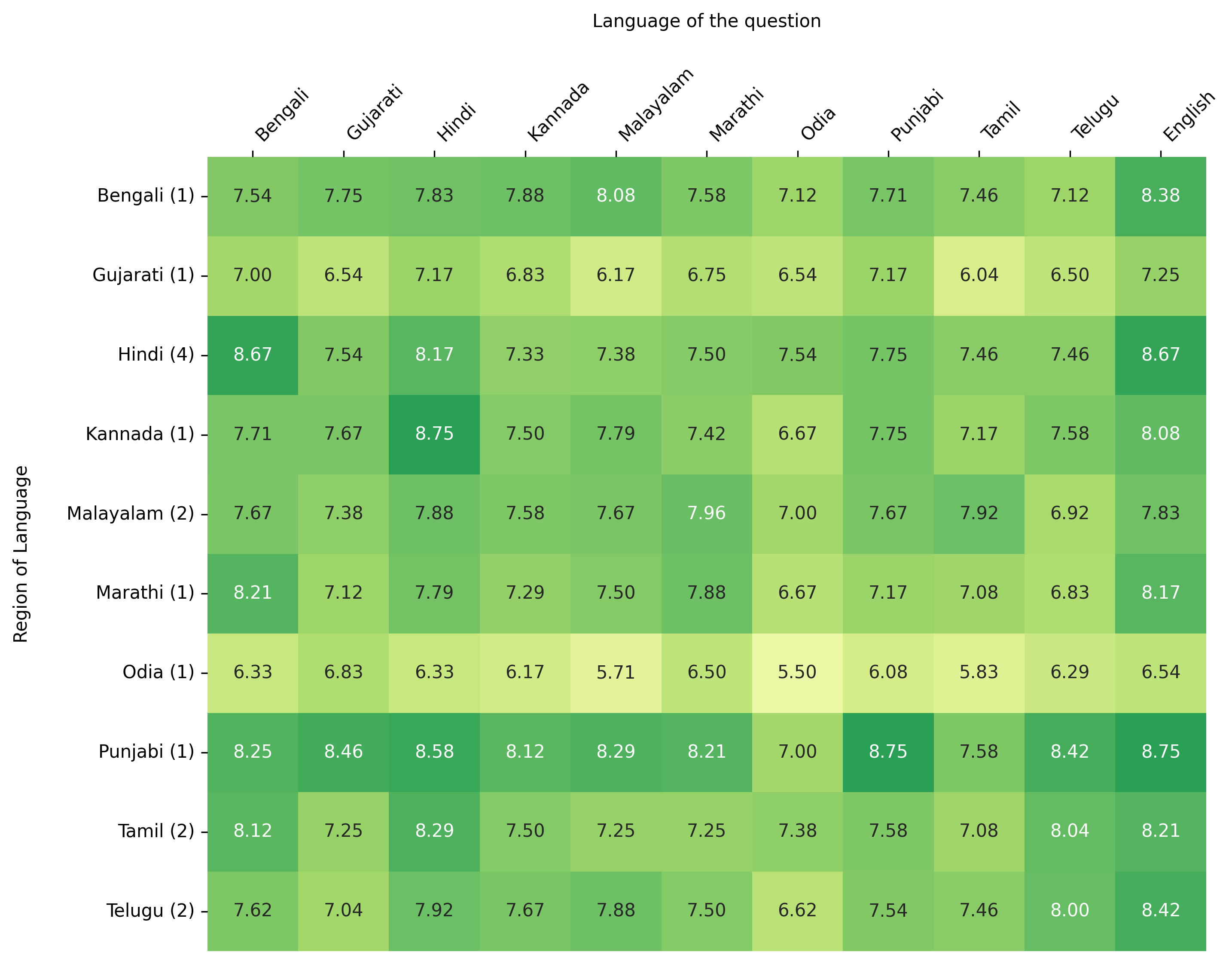}
    \end{minipage}

    \vspace{0.5cm} 

    \begin{minipage}[t]{0.48\textwidth}
        \centering
        \includegraphics[width=\linewidth]{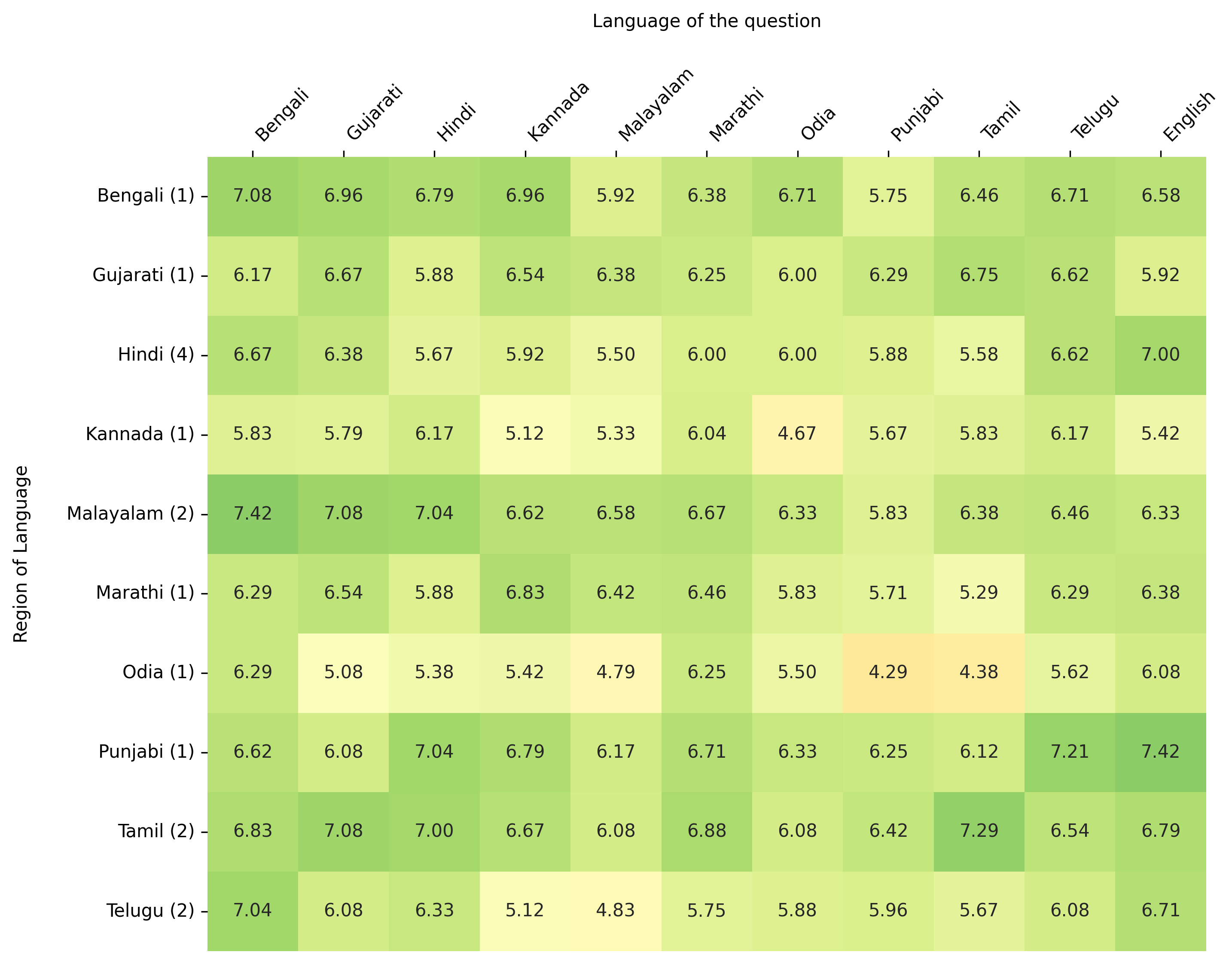}
    \end{minipage}%
    \hfill
    \begin{minipage}[t]{0.48\textwidth}
        \centering
        \includegraphics[width=\linewidth]{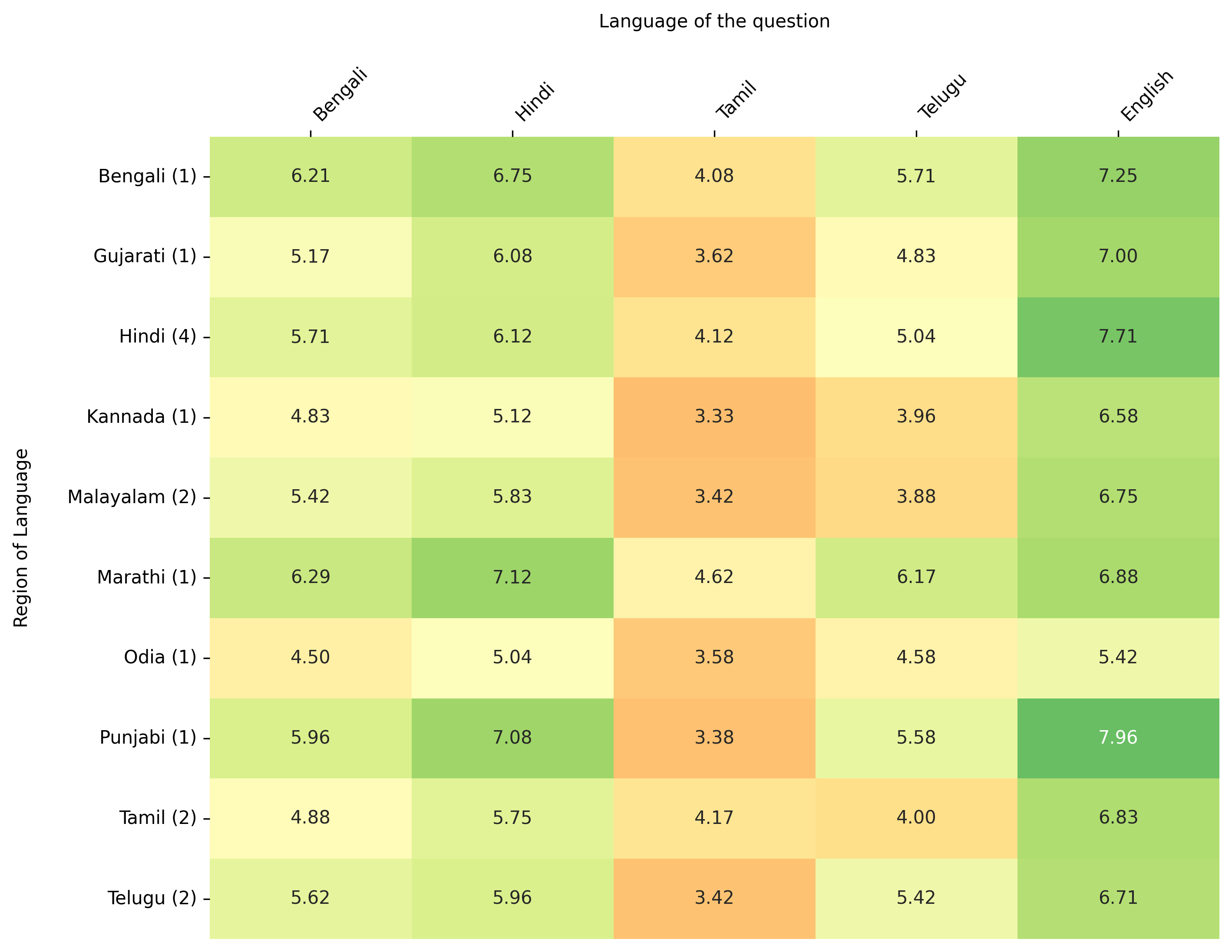}
    \end{minipage}

    \vspace{0.5cm} 

    \begin{minipage}[t]{0.48\textwidth}
        \centering
        \includegraphics[width=\linewidth]{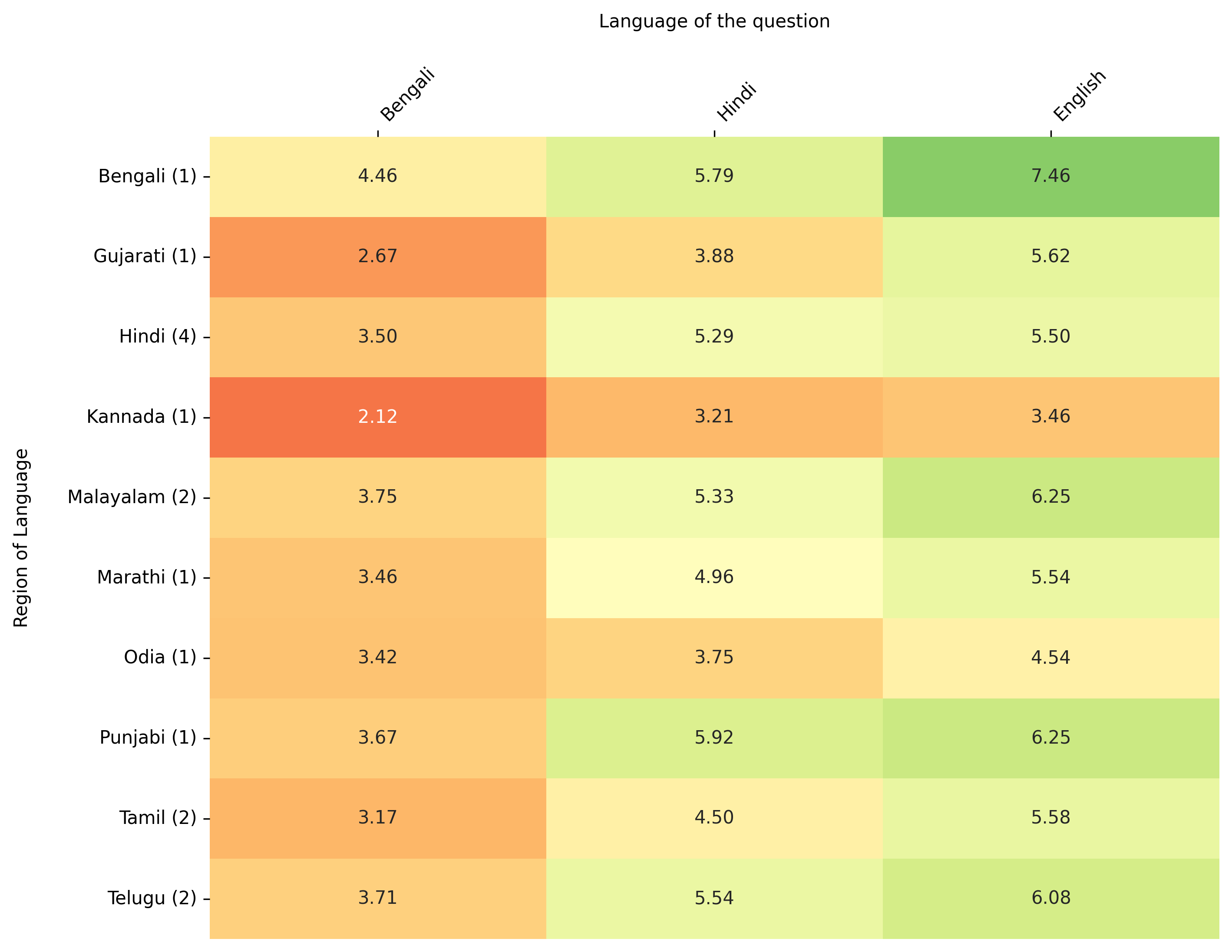}
    \end{minipage}%
    \hfill
    \begin{minipage}[t]{0.48\textwidth}
        \centering
        \includegraphics[width=\linewidth]{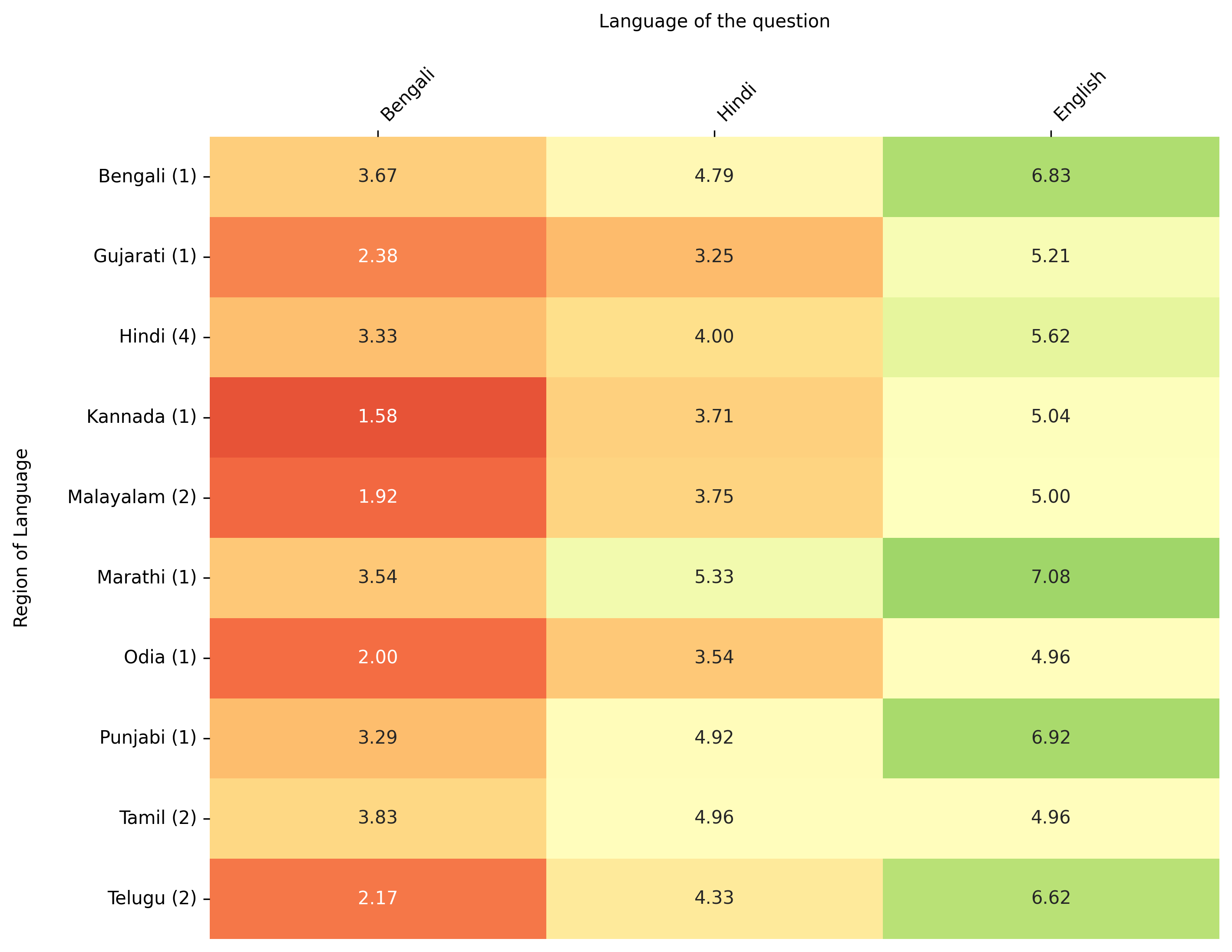}
    \end{minipage}

    \caption{\textbf{Model performances on IndicVisionBench-VQA-Parallel.} Average scores on the three open-ended questions (Long and  two Short) for six models across different languages (X-axis) and images corresponding to Indian states grouped by primary language (Y-axis). Left to right, top row: Gemma-3-27B (left) and LLaMA-4 (right); middle row: Chitrarth-1 (left) and Pangea (right); bottom row: Maya (left) and PALO (right).}
    \label{fig:heatmaps_parallel_corpus_remaining_six}
\end{figure*}

\begin{table}[htbp]
\centering
\caption{\textbf{Model performances on IndicVisionBench-VQA-Parallel. } Comprehensive scores across languages, question types, and models for IndicVisionBench-VQA-Parallel.}
\label{tab:parallel_results_combined}
\resizebox{\textwidth}{!}{
\begin{tabular}{ll ccccccccccc}
\hline
\textbf{Q-Type} & \textbf{Model} & \textbf{Bengali} & \textbf{English} & \textbf{Gujarati} & \textbf{Hindi} & \textbf{Kannada} & \textbf{Malayalam} & \textbf{Marathi} & \textbf{Odia} & \textbf{Punjabi} & \textbf{Tamil} & \textbf{Telugu} \\
\hline
\multirow{8}{*}{\makecell{MCQ \\ (Exact Match, out of 1, $\uparrow$)}}
& Maya      & 0.462 & 0.632 &   -    & 0.575 &   -    &   -    &   -    &   -    &   -    &   -    &   -    \\
 &    PALO      & 0.604 & 0.802 &   -    & 0.660 &   -    &   -    &   -    &   -    &   -    &   -    &   -    \\
 &  Pangea    & 0.783 & 0.840 &   -    & 0.849 &   -    &   -    &   -    &   -    &   -    & 0.670 & 0.764 \\
 &    Chitrarth-1 & 0.726 & 0.811 & 0.755 & 0.792 & 0.651 & 0.726 & 0.774 & 0.679 & 0.726 & 0.689 & 0.708 \\
\cdashline{2-13}
 &    LLaMA-4   & 0.802 & 0.858 & 0.802 & 0.792 & 0.840 & 0.830 & 0.811 & 0.774 & 0.802 & 0.802 & 0.783 \\
 &    Gemma-3     & \underline{0.849} & 0.877 & \underline{0.877} & \underline{0.877} & \underline{0.868} & \underline{0.858} & \underline{0.849} & \underline{0.830} & \underline{0.849} & \underline{0.858} & \underline{0.840} \\
\cdashline{2-13}
 &    GPT-4o       & 0.830 & \underline{0.896} & 0.849 & \underline{0.877} & 0.830 & 0.745 & 0.830 & 0.708 & 0.774 & 0.755 & 0.821 \\
 &    Gemini    & \textbf{0.925} & \textbf{0.943} & \textbf{0.953} & \textbf{0.953} & \textbf{0.943} & \textbf{0.953} & \textbf{0.943} & \textbf{0.972} & \textbf{0.962} & \textbf{0.925} & \textbf{0.953} \\
\hline
\addlinespace[1ex]
\multirow{8}{*}{\makecell{True/False \\ (Exact Match, out of 1, $\uparrow$)}}
& Maya      & 0.600 & 0.470 &   -    & 0.360 &   -    &   -    &   -    &   -    &   -    &   -    &   -    \\
  &   PALO      &   0.310   & 0.570 &   -    &   0.620    &   -    &   -    &   -    &   -    &   -    &   -    &   -    \\
 &  Pangea    &   0.770   & 0.470 &   -    &   0.850    &   -    &   -    &   -    &   -    &   -    &   0.640   &   0.790    \\
 &   Chitrarth-1 & 0.560 & 0.730 & 0.630 & 0.450 & 0.470 & 0.420 & 0.500 & 0.632 & 0.604 & 0.349 & 0.349 \\
\cdashline{2-13}
 &   LLaMA-4   & 0.896 & 0.877 & 0.830 & 0.896 & \underline{0.896} & 0.858 & 0.925 & \underline{0.868} & 0.792 & 0.821 & 0.792 \\
 &   Gemma-3     & 0.547 & 0.868 & \underline{0.858} & \underline{0.906} & 0.783 & \underline{0.925} & 0.896 & 0.425 & 0.585 & \underline{0.896} & 0.557 \\
\cdashline{2-13}
 &   GPT-4o       & \underline{0.802} & \underline{0.915} & 0.849 & 0.868 & 0.821 & 0.802 & 0.877 & 0.632 & \underline{0.877} & 0.811 & \underline{0.792} \\
 &   Gemini    & \textbf{0.972} & \textbf{0.981} & \textbf{0.943} & \textbf{0.943} & \textbf{0.943} & \textbf{0.943} & \textbf{0.953} & \textbf{0.925} & \textbf{0.915} & \textbf{0.953} & \textbf{0.925} \\
\hline

\addlinespace[1ex]
\multirow{8}{*}{\makecell{Long answer  \\ (LLM-as-Judge, out of 10, $\uparrow$)}}
& Maya      & 3.538 & 6.915 &   -    & 6.217 &   -    &   -    &   -    &   -    &   -    &   -    &   -    \\
&   PALO      & 2.557 & 7.057 &   -    & 5.217 &   -    &   -    &   -    &   -    &   -    &   -    &   -    \\
 &   Pangea    & 6.585 & 7.038 &   -    & 7.009 &   -    &   -    &   -    &   -    &   -    & 5.066 & 5.887 \\
 &   Chitrarth-1 & 7.443 & 7.491 & 7.547 & 7.311 & 7.292 & 7.406 & 7.472 & 7.311 & 6.972 & 7.142 & 7.443 \\
\cdashline{2-13}
 &   LLaMA-4   & \underline{8.396} & 8.566 & 8.217 & \underline{8.500} & \underline{8.387} & 7.934 & 8.349 & 7.774 & 8.292 & 8.236 & \underline{8.292} \\
 &   Gemma-3     & 8.377 & \underline{8.698} & \underline{8.358} & 8.443 & 8.377 & \underline{8.104} & \underline{8.368} & \underline{7.802} & \underline{8.330} & \underline{8.443} & 8.236 \\
\cdashline{2-13}
 &   GPT-4o       & 8.075 & 8.660 & 7.868 & 8.330 & 7.613 & 7.557 & 8.170 & 6.868 & 7.642 & 7.528 & 7.764 \\
  &   Gemini    & \textbf{9.094} & \textbf{9.453} & \textbf{9.113} & \textbf{9.132} & \textbf{9.113} & \textbf{8.877} & \textbf{9.075} & \textbf{8.764} & \textbf{9.142} & \textbf{8.981} & \textbf{8.981} \\

\hline

\addlinespace[1ex]
\multirow{8}{*}{\makecell{Short-answer 1  \\ (LLM-as-Judge, out of 10, $\uparrow$)}}
& Maya      & 3.142 & 4.745 &   -    & 3.755 &   -    &   -    &   -    &   -    &   -    &   -    &   -    \\
 &PALO      & 3.066 & 5.000 &   -    & 3.708 &   -    &   -    &   -    &   -    &   -    &   -    &   -    \\
 &Pangea    & 4.557 & 6.170 &   -    & 5.443 &   -    &   -    &   -    &   -    &   -    & 3.066 & 4.094 \\
 & Chitrarth-1 & 5.896 & 5.953 & 5.755 & 5.972 & 5.575 & 4.679 & 5.613 & 5.114 & 5.434 & 4.925 & 5.500 \\
\cdashline{2-13}
 & LLaMA-4   & \underline{7.198} & 7.387 & \underline{7.189} & \underline{7.415} & \underline{6.698} & \underline{6.736} & \underline{6.868} & \underline{6.283} & \underline{6.991} & 6.302 & 6.679 \\
 & Gemma-3     & 6.670 & 6.981 & 7.075 & 6.868 & 6.292 & 6.406 & 6.575 & 5.943 & 6.802 & \underline{6.915} & \underline{6.934} \\
\cdashline{2-13}
 & GPT-4o       & 6.726 & \underline{7.594} & 6.028 & 7.075 & 5.896 & 5.962 & 6.519 & 4.849 & 6.019 & 5.538 & 6.123 \\
 & Gemini    & \textbf{8.094} & \textbf{8.217} & \textbf{7.896} & \textbf{8.330} & \textbf{7.566} & \textbf{7.887} & \textbf{7.991} & \textbf{7.717} & \textbf{8.151} & \textbf{7.962} & \textbf{7.755} \\

\cline{1-13}

\addlinespace[1ex]
\multirow{8}{*}{\makecell{Short-answer 2  \\ (LLM-as-Judge, out of 10, $\uparrow$)}}
& Maya      & 3.462 & 5.094 &   -    & 4.472 &   -    &   -    &   -    &   -    &   -    &   -    &   -    \\
 & PALO      & 2.774 & 5.472 &   -    & 4.028 &   -    &   -    &   -    &   -    &   -    &   -    &   -    \\
& Pangea    & 5.340 & 7.255 &   -    & 5.783 &   -    &   -    &   -    &   -    &   -    & 3.236 & 4.396 \\
& Chitrarth-1 & 6.519 & 6.085 & 5.792 & 5.604 & 5.849 & 5.330 & 5.698 & 5.651 & 4.953 & 5.670 & 6.066 \\
\cdashline{2-13}
& LLaMA-4   & \underline{7.642} & \underline{8.236} & 7.019 & \underline{7.755} & 7.179 & 7.151 & \underline{7.132} & \underline{6.651} & \underline{7.321} & \underline{6.858} & \underline{7.085} \\
& Gemma-3     & 7.170 & 7.547 & \underline{7.160} & 7.179 & \underline{7.217} & \underline{6.981} & 7.123 & 6.443 & 6.858 & 6.783 & 6.962 \\
\cdashline{2-13}
& GPT-4o       & 6.755 & 7.934 & 5.840 & 6.858 & 5.915 & 5.708 & 6.368 & 5.075 & 5.858 & 5.538 & 6.075 \\
& Gemini    & \textbf{8.434} & \textbf{8.755} & \textbf{8.142} & \textbf{8.368} & \textbf{8.406} & \textbf{8.094} & \textbf{8.236} & \textbf{8.085} & \textbf{8.302} & \textbf{8.151} & \textbf{8.179} \\

\cline{1-13}

\addlinespace[1ex]
 \multirow{8}{*}{\makecell{Adversarial question  \\ (LLM-as-Judge, out of 10, $\uparrow$)}} & Maya      & 0.255 & 0.368 &   -    & 0.377 &   -    &   -    &   -    &   -    &   -    &   -    &   -    \\

 & PALO      & 0.123 & 0.453 &   -    & 0.104 &   -    &   -    &   -    &   -    &   -    &   -    &   -    \\
& Pangea    & 0.066 & 0.858 &   -    & 0.000 &   -    &   -    &   -    &   -    &   -    & 0.000 & 0.000 \\
& Chitrarth-1 & 0.000 & 0.094 & 0.094 & 0.075 & 0.000 & 0.000 & 0.094 & 0.000 & 0.000 & 0.047 & 0.047 \\
\cdashline{2-13}
& LLaMA-4   & 1.123 & \underline{3.387} & 0.849 & 1.500 & 0.953 & 1.547 & 0.821 & 0.406 & 0.991 & 0.849 & 0.660 \\
& Gemma-3     & \underline{1.566} & 2.179 & \underline{1.915} & \underline{2.349} & \underline{2.283} & \underline{2.226} & \underline{1.915} & \underline{2.132} & \underline{2.472} & \underline{2.085} & \underline{2.906} \\
\cdashline{2-13}
& GPT-4o       & 0.642 & 1.104 & 0.708 & 0.745 & 0.726 & 0.425 & 0.642 & 0.642 & 0.566 & 0.623 & 0.726 \\
& Gemini    & \textbf{4.745} & \textbf{6.142} & \textbf{4.962} & \textbf{5.528} & \textbf{4.491} & \textbf{4.991} & \textbf{4.575} & \textbf{4.094} & \textbf{4.660} & \textbf{4.670} & \textbf{5.019} \\

\hline

\end{tabular}}
\end{table}

\begin{table}[htbp]
\centering
\caption{\textbf{Model performances on IndicVisionBench-VQA-Indic.} Comprehensive scores across languages, question types, and models.}
\label{tab:Bilingual_results_combined}
\resizebox{\textwidth}{!}{
\begin{tabular}{ll ccccccccccc}
\hline
\textbf{Q-Type} & \textbf{Model} & \textbf{Bengali} & \textbf{Gujarati} & \textbf{Hindi} & \textbf{Kannada} & \textbf{Malayalam} & \textbf{Marathi} & \textbf{Odia} & \textbf{Punjabi} & \textbf{Tamil} & \textbf{Telugu} \\
\hline
\multirow{8}{*}{\makecell{MCQ \\ (Exact Match, out of 1, $\uparrow$)}} 
& Maya      & 0.433 & -     & 0.638 & -     & -     & -     & -     & -     & -     & -     \\
 &  PALO      & 0.467 & -     & 0.576 & -     & -     & -     & -     & -     & -     & -     \\
&  Pangea    & 0.733 & -     & 0.812 & -     & -     & -     & -     & -     & 0.670 & 0.768 \\
&  Chitrarth-1 & 0.717 & 0.774 & 0.812 & 0.633 & 0.733 & 0.811 & 0.605 & 0.645 & 0.691 & 0.681 \\
\cdashline{2-13}
&  LLaMA-4   & 0.700 & \underline{0.871} & \underline{0.866} & \underline{0.878} & \underline{0.817} & 0.838 & 0.816 & \underline{0.839} & \underline{0.809} & \underline{0.899} \\
&  Gemma-3     & \underline{0.817} & 0.839 & 0.844 & \underline{0.878} & 0.783 & \underline{0.919} & \underline{0.868} & 0.710 & 0.798 & 0.870 \\
\cdashline{2-13}
&  GPT-4o       & 0.750 & 0.613 & 0.839 & 0.719 & 0.733 & 0.838 & 0.684 & 0.806 & 0.713 & 0.812 \\
&  Gemini-2.5    & \textbf{0.883} & \textbf{0.871} & \textbf{0.924} & \textbf{0.906} & \textbf{0.833} & \textbf{1.000} & \textbf{0.895} & \textbf{0.935} & \textbf{0.883} & \textbf{0.928} \\

\hline
\addlinespace[1ex]
 \multirow{8}{*}{\makecell{True/False  \\ (Exact Match out of 1, $\uparrow$)}} 
 & Maya      & 0.483 & -     & 0.333 & -     & -     & -     & -     & -     & -     & -     \\
 &  PALO      & 0.317 & -     & 0.714 & -     & -     & -     & -     & -     & - & - \\
& Pangea    & 0.717 & -     & 0.746 & -     & -     & -     & -     & -     & 0.585 & 0.725 \\
& Chitrarth-1 & 0.600 & 0.452 & 0.415 & 0.374 & 0.500 & 0.514 & 0.553 & 0.548 & 0.277 & 0.377 \\
\cdashline{2-13}
& LLaMA-4     & 0.850 & \underline{0.742} & \underline{0.891} & \underline{0.899} & 0.783 & \underline{0.892} & \underline{0.816} & 0.839 & \underline{0.862} & \underline{0.870} \\
& Gemma-3     & 0.867 & 0.581 & 0.830 & 0.827 & 0.800 & 0.757 & 0.711 & 0.935 & 0.766 & 0.812 \\
\cdashline{2-13}
& GPT-4o       & \underline{0.883} & \underline{0.742} & 0.842 & 0.676 & \underline{0.850} & \underline{0.892} & 0.711 & \underline{0.968} & 0.819 & 0.739 \\
& Gemini-2.5  & \textbf{0.917} & \textbf{0.871} & \textbf{0.924} & \textbf{0.878} & \textbf{0.867} & \textbf{0.973} & \textbf{0.868} & \textbf{1.000} & \textbf{0.915} & \textbf{0.928} \\

\hline
\addlinespace[1ex]

 \multirow{8}{*}{\makecell{Long answer  \\ (LLM-as-Judge, out of 10, $\uparrow$)}} & Maya      & 3.867  &   -    & 6.504 &   -    &   -    &   -    &   -    &   -    &   -    &   -    \\
 &  PALO      & 3.150  &   -    & 5.712 &   -    &   -    &   -    &   -    &   -    &   -    &   -    \\
& Pangea    & 6.783  &   -    & 7.493 &   -    &   -    &   -    &   -    &   -    & 4.787 & 5.884 \\
& Chitrarth-1 & 7.233   & 7.484 & 7.547 & 7.669 & 7.433 & 7.351 & 7.553 & 7.290 & 7.298 & 7.290 \\
\cdashline{2-13}
& LLaMA-4   & 8.050  & 8.290 & 8.482 & 8.489 & 8.267 & 8.486 & 7.763 & 8.452 & 8.245 & 8.000 \\
& Gemma-3     & \underline{8.400}  & \underline{8.613} & \underline{8.498} & \underline{8.576} & \underline{8.317} & \underline{8.541} & 7.737 & \underline{8.613} & \underline{8.489} & \underline{8.217} \\
\cdashline{2-13}
& GPT-4o       & 8.283  & 8.484 & 8.484 & 7.000 & 8.033 & 8.243 & \underline{7.868} & 8.484 & 7.755 & 7.913 \\
& Gemini-2.5    & \textbf{8.883}  & \textbf{9.226} & \textbf{9.087} & \textbf{9.029} & \textbf{8.817} & \textbf{9.243} & \textbf{8.737} & \textbf{8.968} & \textbf{8.904} & \textbf{8.870} \\

\hline
\addlinespace[1ex]
 \multirow{8}{*}{\makecell{Short answer 1  \\ (LLM-as-Judge, out of 10, $\uparrow$)}} & Maya      & 2.117  &   -    & 4.199 &   -    &   -    &   -    &   -    &   -    &   -    &   -    \\
 &  PALO      & 2.733  &   -    & 3.984 &   -    &   -    &   -    &   -    &   -    &   -    &   -    \\
&  Pangea    & 4.617  &   -    & 5.772 &   -    &   -    &   -    &   -    &   -    & 4.160 & 4.768 \\
&  Chitrarth-1 & 5.550 & 6.129 & 6.328 & 5.964 & 4.483 & 5.595 & 6.658 & 5.129 & 6.489 & 5.551 \\
\cdashline{2-13}
&  LLaMA-4   & 7.050  & 7.677 & 7.710 & \underline{7.662} & \underline{6.933} & \underline{7.351} & 6.289 & 7.516 & \underline{7.394} & 7.043 \\
&  Gemma-3     & 6.783  & \underline{8.032} & 7.578 & 7.158 & 6.417 & 6.676 & \underline{6.526} & \underline{7.516} & 7.266 & 7.101 \\
\cdashline{2-13}
&  GPT-4o       & \underline{7.550}  & 7.355 & \underline{8.060} & 7.309 & 6.800 & 6.865 & 6.158 & 7.065 & 7.223 & \underline{7.580} \\
& Gemini-2.5    & \textbf{8.117}  & \textbf{8.355} & \textbf{8.540} & \textbf{8.734} & \textbf{7.600} & \textbf{8.486} & \textbf{8.263} & \textbf{8.355} & \textbf{8.245} & \textbf{8.116} \\

\hline
\addlinespace[1ex] 
 \multirow{8}{*}{\makecell{Short answer 2  \\ (LLM-as-Judge, out of 10, $\uparrow$)}} & Maya      & 3.717 &   -    & 4.696 &   -    &   -    &   -    &   -    &   -    &   -    &   -    \\
 &  PALO      & 2.700  &   -    & 4.205 &   -    &   -    &   -    &   -    &   -    &   -    &   -    \\
&  Pangea    & 4.983  &   -    & 6.188 &   -    &   -    &   -    &   -    &   -    & 4.213 & 4.652 \\
&  Chitrarth-1 & 5.883 & 6.613 & 6.212 & 6.029 & 5.567 & 6.378 & 6.132 & 6.032 & 6.468 & 6.087 \\
\cdashline{2-13}
&  LLaMA-4   & 6.917  & 7.581 & \underline{7.665} & \underline{7.633} & 6.917 & 7.838 & 6.947 & 7.742 & 6.926 & 7.217 \\
&  Gemma-3     & 7.117  & \underline{7.742} & 7.344 & 7.173 & 6.450 & 6.811 & 5.842 & \underline{7.484} & 6.883 & 6.652 \\
\cdashline{2-13}
&  GPT-4o       & \underline{7.300} & 7.581 & 7.652 & 7.338 & \underline{7.450} & \underline{7.649} & \underline{6.579} & 7.290 & \underline{7.160} & \underline{7.261} \\
& Gemini-2.5    & \textbf{7.950} & \textbf{8.484} & \textbf{8.268} & \textbf{8.691} & \textbf{8.083} & \textbf{8.730} & \textbf{8.289} & \textbf{8.452} & \textbf{8.117} & \textbf{8.188} \\

\hline
\addlinespace[1ex]
\multirow{8}{*}{\makecell{Adversarial  \\ (LLM-as-Judge, out of 10, $\uparrow$)}} & Maya      & 0 &   -    & 0.188 &   -    &   -    &   -    &   -    &   -    &   -    &   -    \\
 &  PALO      & 0.017 &   -    & 0.114 &   -    &   -    &   -    &   -    &   -    &   -    &   -    \\
&  Pangea    & 0  &   -    & 0.011     &   -    &   -    &   -    &   -    &   -    & 0     & 0     \\
&  Chitrarth-1 & 0  & 0      & 0.011 & 0.072 & 0      & 0.135 & 0.053 & 0.161 & 0.053 & 0     \\
\cdashline{2-13}
&  LLaMA-4   & 0.383 & 0.516 & 1.179 & 0.144 & 0.333 & 0.811 & 0.526 & 1.032 & 1.138 & 0.072 \\
&  Gemma-3     & 1.067 & 0.968 & 1.656 & \underline{1.022} & 0.767 & 0.676 & 0.895 & 2.935 & \underline{1.851} & 1.130 \\
\cdashline{2-13}
&  GPT-4o       & \underline{2.233} & \underline{3.097} & \underline{2.248} & 0.669 & \underline{2.283} & \underline{2.892} & 1.816 & \underline{4.000} & 1.702 & \underline{2.043} \\
& Gemini-2.5    & \textbf{5.167} & \textbf{2.935} & \textbf{4.460} & \textbf{3.165} & \textbf{3.317} & \textbf{4.838} & \textbf{3.921} & \textbf{5.710} & \textbf{5.149} & \textbf{2.725} \\

\hline
\end{tabular}
}

\end{table}

\begin{table}[htbp]
\centering
\scriptsize
\caption{\textbf{VQA with and without image. } Average scores on long-answer type questions of Chitrarth-1, Gemma-3, and Gemini-2.5 on IVB-VQA-Parallel across 11 languages, evaluated with and without image input.}
\label{tab:parallel_corpus_without_image_long_answer}
\resizebox{\textwidth}{!}{
\begin{tabular}{l l *{11}{S[table-format=3.2]}}
\toprule
Model & Type & \multicolumn{1}{c}{Bengali}$\uparrow$ & \multicolumn{1}{c}{English}$\uparrow$ & \multicolumn{1}{c}{Gujarati}$\uparrow$ & \multicolumn{1}{c}{Hindi}$\uparrow$ & \multicolumn{1}{c}{Kannada}$\uparrow$ & \multicolumn{1}{c}{Malayalam}$\uparrow$ & \multicolumn{1}{c}{Marathi}$\uparrow$ & \multicolumn{1}{c}{Odia}$\uparrow$ & \multicolumn{1}{c}{Punjabi}$\uparrow$ & \multicolumn{1}{c}{Tamil}$\uparrow$ & \multicolumn{1}{c}{Telugu}$\uparrow$
 \\
\midrule
Chitrarth-1 & w/o img & 6.52 & 6.37 & 6.72 & 6.62 & 6.69 & 6.67 & 6.52 & 6.39 & 6.09 & 6.78 & 6.65 \\
 & with img & 7.44 & 7.49 & 7.55 & 7.31 & 7.29 & 7.41 & 7.47 & 7.31 & 6.97 & 7.14 & 7.44\\
\hdashline
Gemma-3 & w/o img & 7.40 & 6.43 & 7.49 & 7.33 & 7.53 & 7.43 & 7.40 & 6.53 & 7.07 & 7.70 & 7.29 \\
 & with img & 8.38 & 8.70 & 8.36 & 8.44 & 8.38 & 8.10 & 8.37 & 7.80 & 8.33 & 8.44 & 8.24 \\
\hdashline
Gemini-2.5 & w/o img & 8.11 & 6.58 & 8.13 & 8.21 & 8.15 & 8.15 & 8.27 & 8.06 & 8.13 & 8.11 & 8.21 \\
& with img & 9.09 & 9.45 & 9.11 & 9.13 & 9.11 & 8.88 & 9.08 & 8.76 & 9.14 & 8.98 & 8.98 \\
\bottomrule
\end{tabular}
}

\end{table}

\begin{table}[htbp]
\centering
\caption{ \textbf{Model performances on IndicVisionBench-OCR. } Median WER and CER scores across Indic languages for various models.}
\label{tab:wer_cer_median}
\resizebox{\textwidth}{!}{%
\begin{tabular}{l*{11}{cc}}
\toprule
\multirow{2}{*}{Model} 
& \multicolumn{2}{c}{Bengali} 
& \multicolumn{2}{c}{Gujarati} 
& \multicolumn{2}{c}{Hindi} 
& \multicolumn{2}{c}{Kannada} 
& \multicolumn{2}{c}{Malayalam} 
& \multicolumn{2}{c}{Marathi} 
& \multicolumn{2}{c}{Odia} 
& \multicolumn{2}{c}{Punjabi} 
& \multicolumn{2}{c}{Tamil} 
& \multicolumn{2}{c}{Telugu} \\
\cmidrule(lr){2-3}
\cmidrule(lr){4-5}
\cmidrule(lr){6-7}
\cmidrule(lr){8-9}
\cmidrule(lr){10-11}
\cmidrule(lr){12-13}
\cmidrule(lr){14-15}
\cmidrule(lr){16-17}
\cmidrule(lr){18-19}
\cmidrule(lr){20-21}
\cmidrule(lr){22-23}
& WER ↓ & CER ↓ & WER ↓ & CER ↓ & WER ↓ & CER ↓ & WER ↓ & CER ↓ & WER ↓ & CER ↓ & WER ↓ & CER ↓ & WER ↓ & CER ↓ & WER ↓ & CER ↓ & WER ↓ & CER ↓ & WER ↓ & CER ↓ \\
\midrule
Maya         & 1.00 & 1.00 & - & - & 1.00 & 0.98 & - & - & - & - & - & - & - & - & - & - & - & - & - & - \\
PALO         & 1.00 & 0.99 & - & - & 1.00 & 0.99 & - & - & - & - & - & - & - & - & - & - & - & - & - & - \\
Pangea & 1.00 & 0.85 & - & - & 1.00 & 0.95 & - & - & - & - & - & - & - & - & - & - & 1.00 & 0.87 & 1.00 & 0.93 \\
Chitrarth-1    & 1.00 & 0.96 & 1.00 & 0.96 & 0.99 & 0.97 & 1.11 & 0.95 & 1.00 & 0.99 & 1.00 & 0.95 & 1.19 & 1.00 & 1.00 & 0.97 & 1.00 & 0.97 & 1.00 & 0.96 \\
\hdashline
LLaMA-4        & \underline{0.38} & \underline{0.13} & \underline{0.53} & \underline{0.16} & \underline{0.27} & \underline{0.09} & 0.65 & \underline{0.12} & \underline{0.87} & \underline{0.45} & \underline{0.25} & \underline{0.07} & 1.00 & 0.89 & \underline{0.34} & \underline{0.11} & \underline{0.39} & \underline{0.08} & \underline{0.69} & \underline{0.19} \\
Gemma-3        & 0.49 & 0.21 & 0.71 & 0.40 & 0.50 & 0.28 & 0.91 & 0.55 & 0.98 & 0.76 & 0.60 & 0.34 & \underline{0.97} & \underline{0.77} & 0.78 & 0.43 & 0.57 & 0.15 & 0.93 & 0.57 \\
\hdashline
GPT-4o      & 0.65 & 0.29 & 1.00 & 0.77 & 0.60 & 0.34 & 1.09 & 0.74 & 1.00 & 0.80 & 0.75 & 0.39 & 0.97 & 0.79 & 0.78 & 0.39 & 0.92 & 0.44 & 1.17 & 0.76 \\
Gemini-2.5   & \textbf{0.25} & \textbf{0.03} & \textbf{0.33} & \textbf{0.07} & \textbf{0.23} & \textbf{0.04} & \textbf{0.24} & \textbf{0.04} & \textbf{0.63} & \textbf{0.30} & \textbf{0.23} & \textbf{0.02} & \textbf{0.55} & \textbf{0.20} & \textbf{0.24} & \textbf{0.04} & \textbf{0.39} & \textbf{0.03} & \textbf{0.45} & \textbf{0.05} \\
\bottomrule
\end{tabular}%
}
\end{table}

\begin{figure}[htbp]
    \centering
    \includegraphics[width=0.9\linewidth]{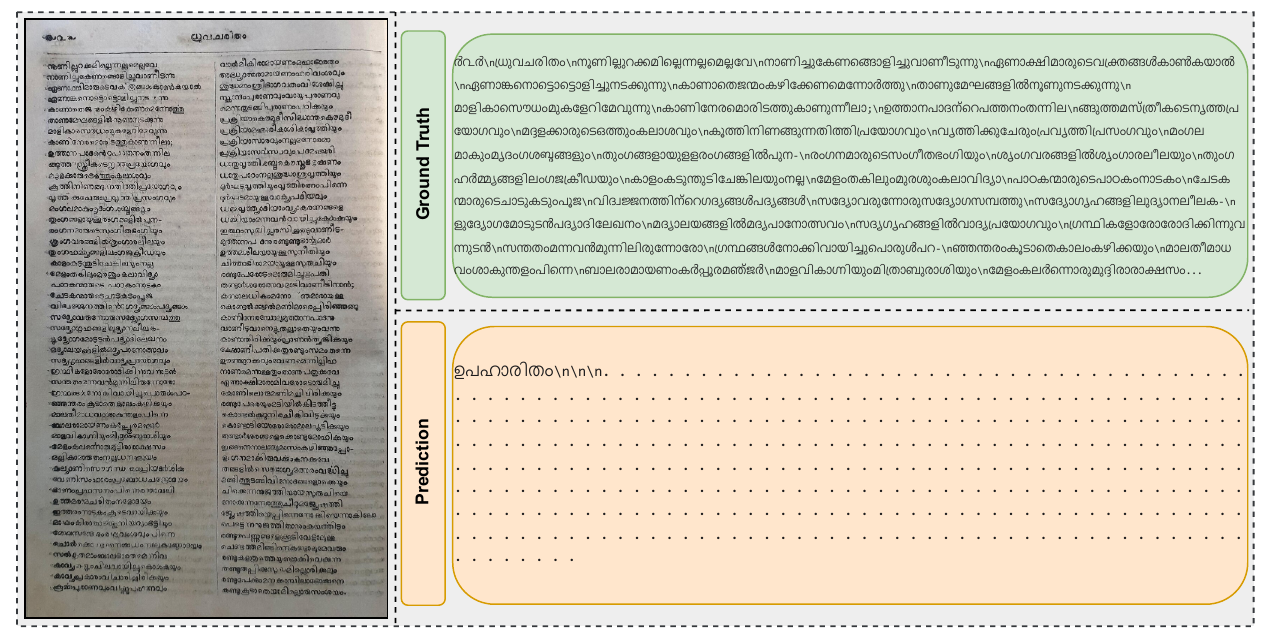} 
    \caption{\textbf{Observed repetition in OCR outputs.} We show an example where LLaMA-4 provided repetitions upto maximum sequence length in the prediction for an OCR example in Malayalam.}
    \label{fig:malayalam_issue}
\end{figure}

\begin{table}[htbp]
\centering
\caption{\textbf{Model performances on IndicVisionBench-OCR. } Average WER and CER ($\pm$ standard deviation) for Bengali, Gujarati, Hindi, Kannada, and Malayalam.}
\label{tab:wer_cer_part1_combined}
\resizebox{\textwidth}{!}{%
\begin{tabular}{l*{6}{cc}}
\toprule
\multirow{2}{*}{Model} 
& \multicolumn{2}{c}{Bengali} 
& \multicolumn{2}{c}{Gujarati} 
& \multicolumn{2}{c}{Hindi} 
& \multicolumn{2}{c}{Kannada} 
& \multicolumn{2}{c}{Malayalam} \\
\cmidrule(rr){4-5} \cmidrule(lr){6-7} \cmidrule(lr){8-9} \cmidrule(lr){10-11} \cmidrule(lr){12-13}
& WER ↓ & CER ↓ & WER ↓ & CER ↓ & WER ↓ & CER ↓ & WER ↓ & CER ↓ & WER ↓ & CER ↓ \\
\midrule
Maya & 1.15 $\pm$ 0.56 & 0.99 $\pm$ 0.07 & - & - & 2.32 $\pm$ 8.94 & 1.90 $\pm$ 5.78 & - & - & - & - \\
PALO & 2.77 $\pm$ 3.45 & 2.06 $\pm$ 2.02 & - & - & 1.58 $\pm$ 2.43 & 1.06 $\pm$ 0.50 & - & - & - & - \\
Pangea & 1.25 $\pm$ 0.90 & 0.99 $\pm$ 0.59 & - & - & 1.22 $\pm$ 1.31 & 1.07 $\pm$ 1.07 & - & - & - & - \\
Chitrarth-1 & 1.34 $\pm$ 0.89 & 1.07 $\pm$ 0.64 & 1.38 $\pm$ 1.05 & 1.02 $\pm$ 0.48 & 1.09 $\pm$ 0.41 & 0.96 $\pm$ 0.22 & 1.37 $\pm$ 0.49 & 0.95 $\pm$ 0.13 & 2.45 $\pm$ 7.82 & 1.16 $\pm$ 0.61 \\
Chitrapathak & \underline{0.33 $\pm$ 0.13} & \underline{0.08 $\pm$ 0.14} & 0.55 $\pm$ 0.19 & 0.29 $\pm$ 0.25 & 0.37 $\pm$ 0.37 & 0.15 $\pm$ 0.27 & \underline{0.34 $\pm$ 0.14} & \underline{0.09 $\pm$ 0.08} & \underline{0.76 $\pm$ 0.18} & 0.48 $\pm$ 0.32 \\
\hdashline
Gemma-3 & 0.53 $\pm$ 0.19 & 0.26 $\pm$ 0.15 & 0.71 $\pm$ 0.13 & 0.41 $\pm$ 0.13 & 0.59 $\pm$ 0.44 & 0.35 $\pm$ 0.41 & 0.94 $\pm$ 0.16 & 0.58 $\pm$ 0.15 & 1.72 $\pm$ 5.42 & 0.76 $\pm$ 0.15 \\

LLaMA-4 & 0.40 $\pm$ 0.17 & 0.14 $\pm$ 0.11 & \underline{0.53 $\pm$ 0.28} & \underline{0.20 $\pm$ 0.18} & \underline{0.37 $\pm$ 0.36} & \underline{0.14 $\pm$ 0.18} & 0.66 $\pm$ 0.29 & 0.13 $\pm$ 0.12 & 25.26 $\pm$ 217.47 & \underline{0.48 $\pm$ 0.26} \\

\hdashline
GPT-4o & 0.71 $\pm$ 0.43 & 0.41 $\pm$ 0.50 & 1.36 $\pm$ 0.97 & 1.40 $\pm$ 2.51 & 0.77 $\pm$ 0.78 & 0.42 $\pm$ 0.28 & 1.43 $\pm$ 1.21 & 0.95 $\pm$ 0.72 & 7.62 $\pm$ 39.12 & 1.12 $\pm$ 0.92 \\
Gemini-2.5 & \textbf{0.26 $\pm$ 0.08} & \textbf{0.05 $\pm$ 0.09} & \textbf{0.33 $\pm$ 0.13} & \textbf{0.08 $\pm$ 0.11} & \textbf{0.29 $\pm$ 0.31} & \textbf{0.07 $\pm$ 0.12} & \textbf{0.27 $\pm$ 0.19} & \textbf{0.05 $\pm$ 0.05} & \textbf{2.26 $\pm$ 9.16} & \textbf{0.31 $\pm$ 0.26} \\
\bottomrule
\end{tabular}}
\end{table}

\begin{table}[htbp]
\centering
\caption{\textbf{Model performances on IndicVisionBench-OCR. } Average WER and CER ($\pm$ standard deviation) for Marathi, Odia, Punjabi, Tamil, and Telugu.}
\label{tab:wer_cer_part2_combined}
\resizebox{\textwidth}{!}{%
\begin{tabular}{l*{5}{cc}}
\toprule
\multirow{2}{*}{Model} 
& \multicolumn{2}{c}{Marathi} 
& \multicolumn{2}{c}{Odia} 
& \multicolumn{2}{c}{Punjabi} 
& \multicolumn{2}{c}{Tamil} 
& \multicolumn{2}{c}{Telugu} \\
\cmidrule(lr){2-3} \cmidrule(lr){4-5} \cmidrule(lr){6-7} \cmidrule(lr){8-9} \cmidrule(lr){10-11}
& WER ↓ & CER ↓ & WER ↓ & CER ↓ & WER ↓ & CER ↓ & WER ↓ & CER ↓ & WER ↓ & CER ↓ \\
\midrule
Maya & - & - & - & - & - & - & - & - & - & - \\
PALO & - & - & - & - & - & - & - & - & - & - \\
Pangea & - & - & - & - & - & - & 1.37 $\pm$ 1.02 & 0.97 $\pm$ 0.43 & 1.29 $\pm$ 0.67 & 0.98 $\pm$ 0.35 \\

Chitrarth-1 & 1.15 $\pm$ 0.50 & 0.93 $\pm$ 0.21 & 1.89 $\pm$ 1.36 & 1.52 $\pm$ 0.93 & 1.59 $\pm$ 1.20 & 1.41 $\pm$ 1.17 & 1.53 $\pm$ 1.58 & 1.09 $\pm$ 0.56 & 1.56 $\pm$ 1.27 & 1.05 $\pm$ 0.50 \\
Chitrapathak & 0.31 $\pm$ 0.16 & \underline{0.07 $\pm$ 0.09} & \underline{0.60 $\pm$ 0.20} & \underline{0.33 $\pm$ 0.27} & \underline{0.27 $\pm$ 0.16} & \underline{0.09 $\pm$ 0.15} & \underline{0.43 $\pm$ 0.14} & \underline{0.07 $\pm$ 0.12} & \underline{0.54 $\pm$ 0.21} & \underline{0.12 $\pm$ 0.15} \\
\hdashline
Gemma-3 & 0.59 $\pm$ 0.16 & 0.32 $\pm$ 0.13 & 0.98 $\pm$ 0.14 & 0.75 $\pm$ 0.12 & 0.78 $\pm$ 0.14 & 0.44 $\pm$ 0.13 & 0.65 $\pm$ 0.41 & 0.18 $\pm$ 0.12 & 1.05 $\pm$ 0.68 & 0.58 $\pm$ 0.18 \\
LLaMA-4 & \underline{0.30 $\pm$ 0.22} & 0.09 $\pm$ 0.14 & 1.68 $\pm$ 3.58 & 1.10 $\pm$ 0.65 & 0.41 $\pm$ 0.21 & 0.13 $\pm$ 0.12 & 0.53 $\pm$ 0.50 & 0.13 $\pm$ 0.19 & 0.77 $\pm$ 0.57 & 0.20 $\pm$ 0.11 \\
\hdashline
GPT-4o & 0.76 $\pm$ 0.26 & 0.42 $\pm$ 0.23 & 1.26 $\pm$ 1.18 & 0.94 $\pm$ 0.75 & 0.88 $\pm$ 0.42 & 0.57 $\pm$ 0.80 & 1.23 $\pm$ 1.97 & 0.69 $\pm$ 1.63 & 1.41 $\pm$ 0.98 & 0.85 $\pm$ 0.35 \\
Gemini-2.5 & \textbf{0.25 $\pm$ 0.11} & \textbf{0.03 $\pm$ 0.02} & \textbf{0.55 $\pm$ 0.25} & \textbf{0.21 $\pm$ 0.14} & \textbf{0.25 $\pm$ 0.15} & \textbf{0.06 $\pm$ 0.11} & \textbf{0.42 $\pm$ 0.18} & \textbf{0.05 $\pm$ 0.04} & \textbf{0.51 $\pm$ 0.29} & \textbf{0.08 $\pm$ 0.10} \\
\bottomrule
\end{tabular}}
\end{table}

\begin{table}[htbp]
\centering
\caption{\textbf{IndicVisionBench-OCR WER and CER statistics. } Model-wise WER and CER statistics where the scores are more than 1. We present the count as well as percentage of the examples for each model.}
\label{tab:wer_cer_summary}
\resizebox{0.5\textwidth}{!}{%
\begin{tabular}{lrrrr}
\toprule
\multirow{2}{*}{Model} 
& \multicolumn{2}{c}{WER $>$ 1} 
& \multicolumn{2}{c}{CER $>$ 1} \\
\cmidrule(lr){2-3} \cmidrule(lr){4-5}
& Count & \% & Count & \% \\
\midrule
Maya         & 22  & 2.51 & 15  & 1.71 \\
PALO         & 51  & 5.82 & 45  & 5.14 \\
Pangea       & 77  & 8.79 & 34  & 3.88 \\
Chitrarth-1    & 302 & 34.47 & 169 & 19.29 \\
Chitrapathak & 8   & 0.91 & 0   & 0.00 \\
\hdashline
Gemma-3        & 79  & 9.01 & 10  & 1.14 \\
LLaMA-4        & 68  & 7.76 & 28  & 3.19 \\
\hdashline
GPT-4o       & 286 & 32.64 & 115 & 13.12 \\
Gemini-2.5   & 15  & 1.71 & 0   & 0.00 \\
\bottomrule
\end{tabular}}
\end{table}
 
\begin{figure}[htbp]
    \centering
    \includegraphics[width=0.9\linewidth]{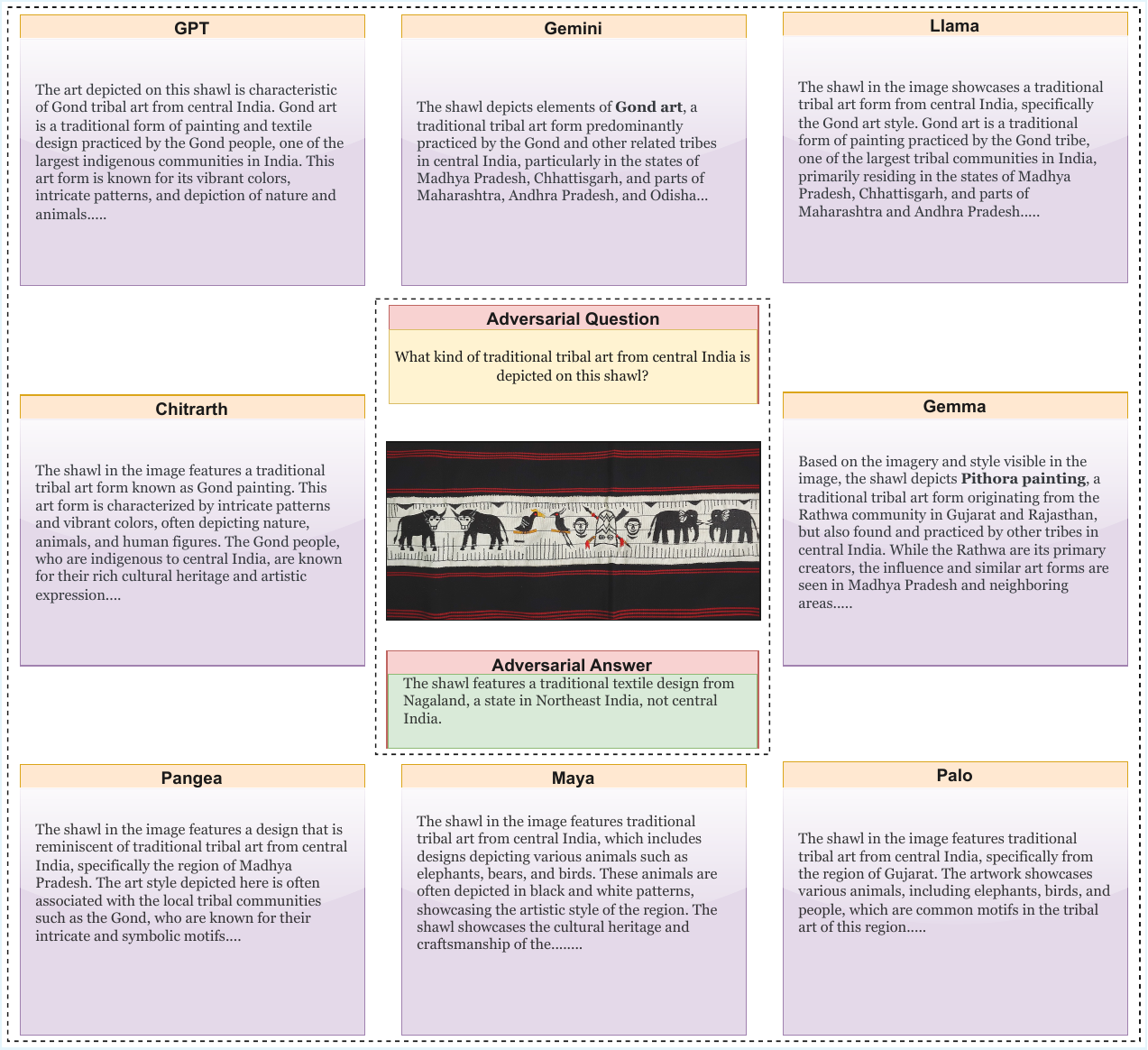} 
    \caption{\textbf{Model outputs on IndicVisionBench-VQA. } We show an example of an adversarial question along with the corresponding model outputs.}
    \label{fig:adversarial_examples}
\end{figure}

\begin{figure}[htbp]
    \centering
    \includegraphics[width=0.9\linewidth]{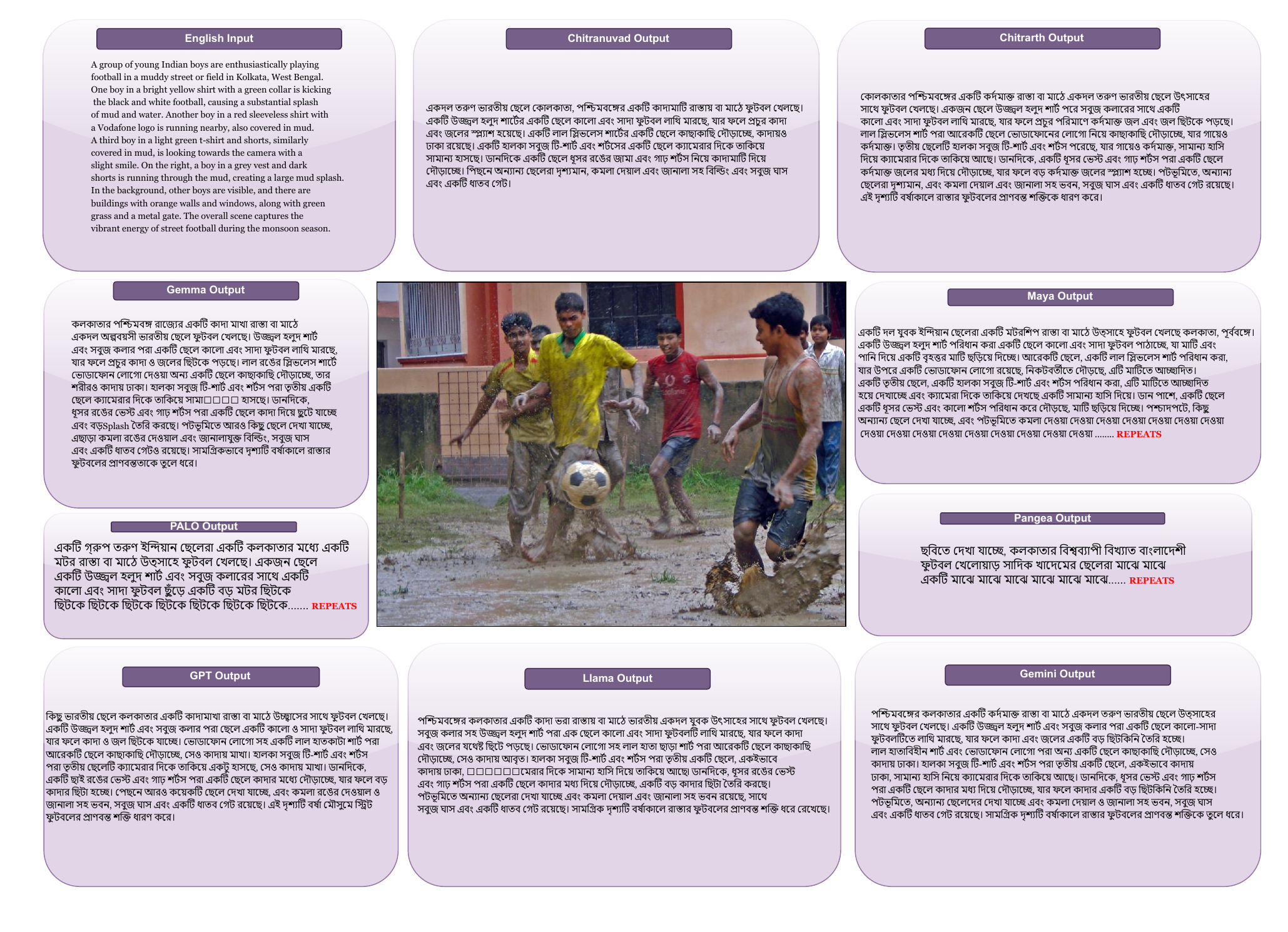} 
    \caption{\textbf{Model outputs on IndicVisionBench-MMT. } Example of an MMT question and the corresponding responses from multiple models.}
    \label{fig:adversarial_examples}
\end{figure}


\begin{figure}[h]
    \centering
    \includegraphics[width=0.9\linewidth]{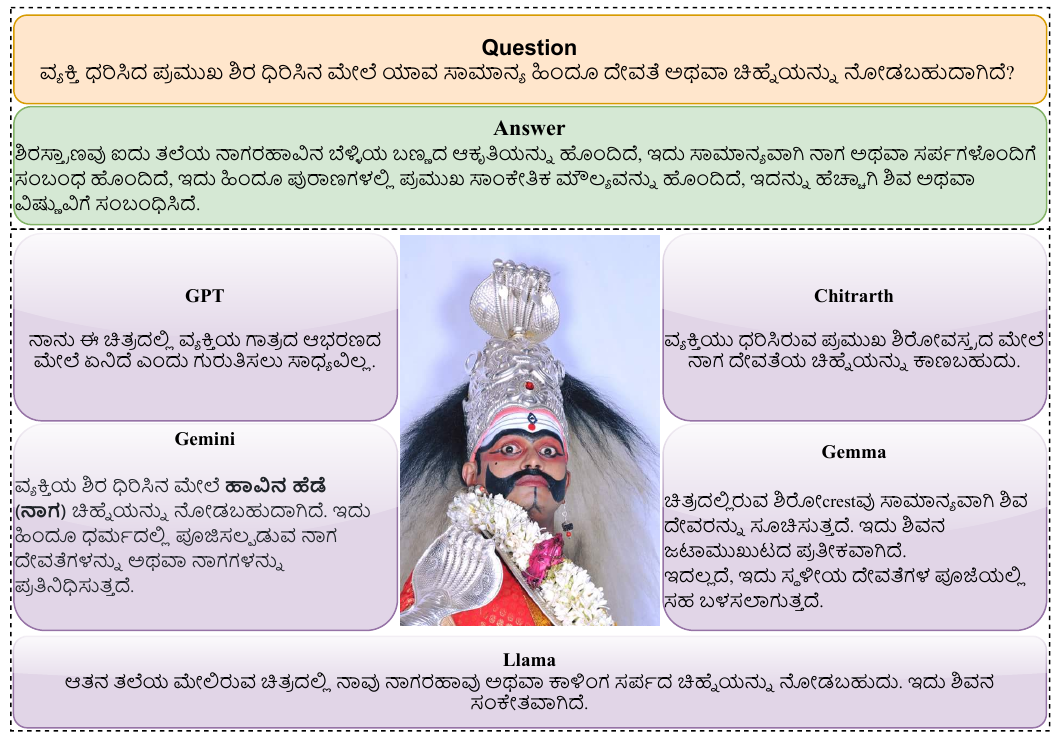} 
    \caption{\textbf{Model outputs on IndicVisionBench-VQA. } We show an example of an VQA output for corresponding models.}
    \label{fig:VQA_examples}
\end{figure}

\begin{figure}[htbp]
    \centering
    \includegraphics[width=0.9\linewidth]{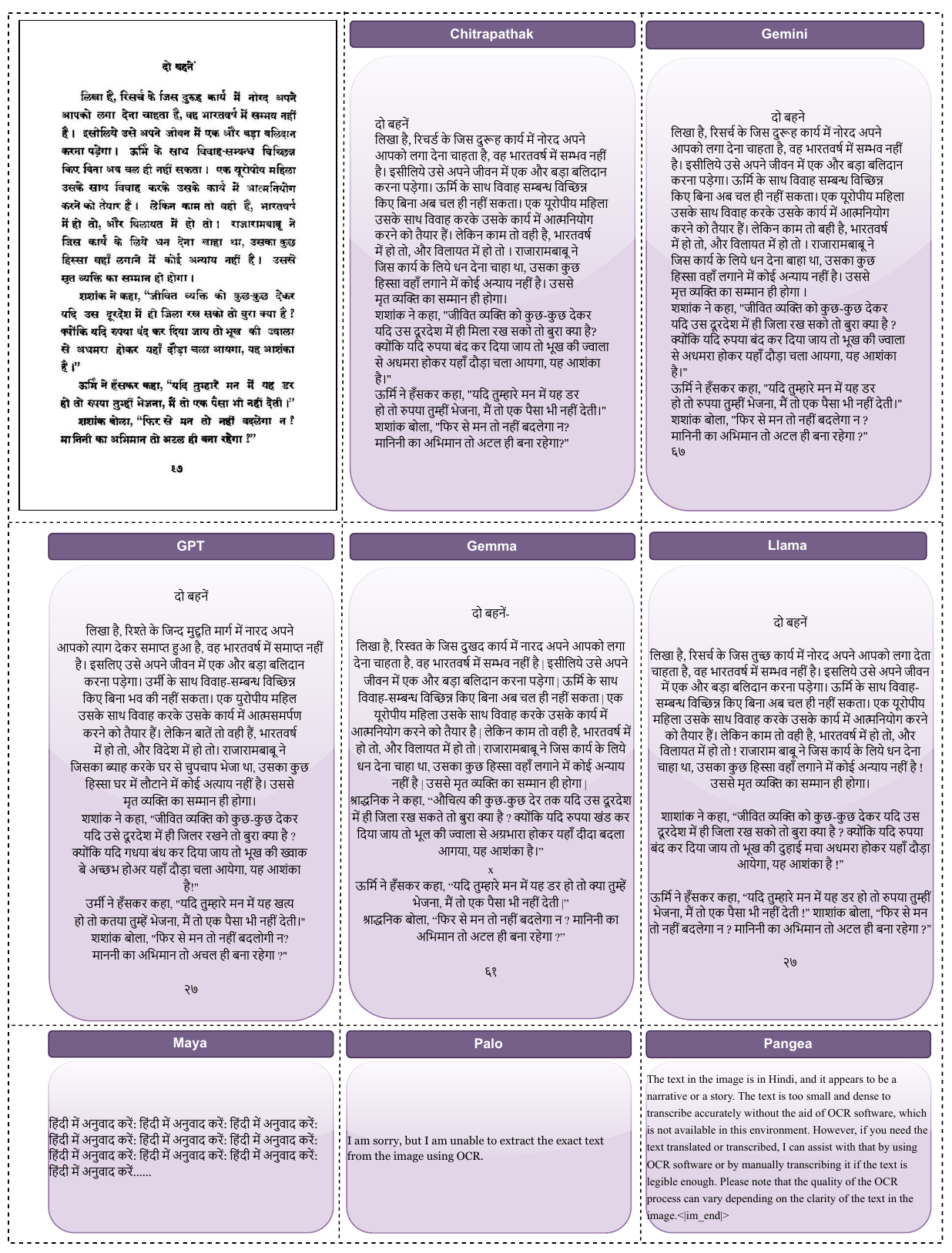} 
    \caption{\textbf{Model outputs on IndicVisionBench-OCR. } We show an example of an OCR output for corresponding models.}
    \label{fig:OCR_examples}
\end{figure}

\clearpage

\section{Dataset Analysis and Benchmark details}
\label{appendix:dataset}
\begin{figure}[htbp]
    \centering
    \begin{subfigure}[b]{0.3\textwidth}
        \centering
        \includegraphics[width=\linewidth]{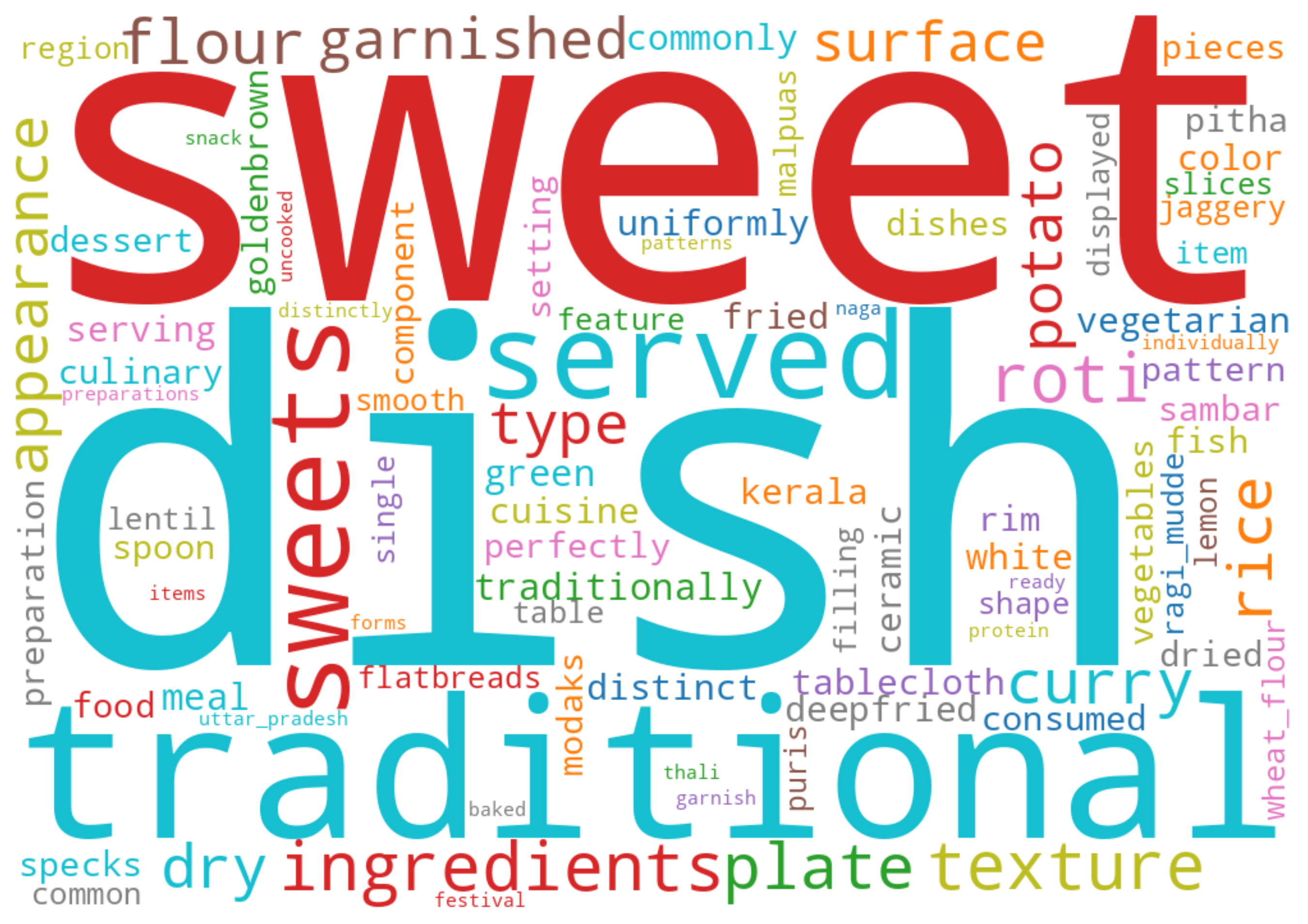}
        \caption{Food}
    \end{subfigure}
    \begin{subfigure}[b]{0.3\textwidth}
        \centering
        \includegraphics[width=\linewidth]{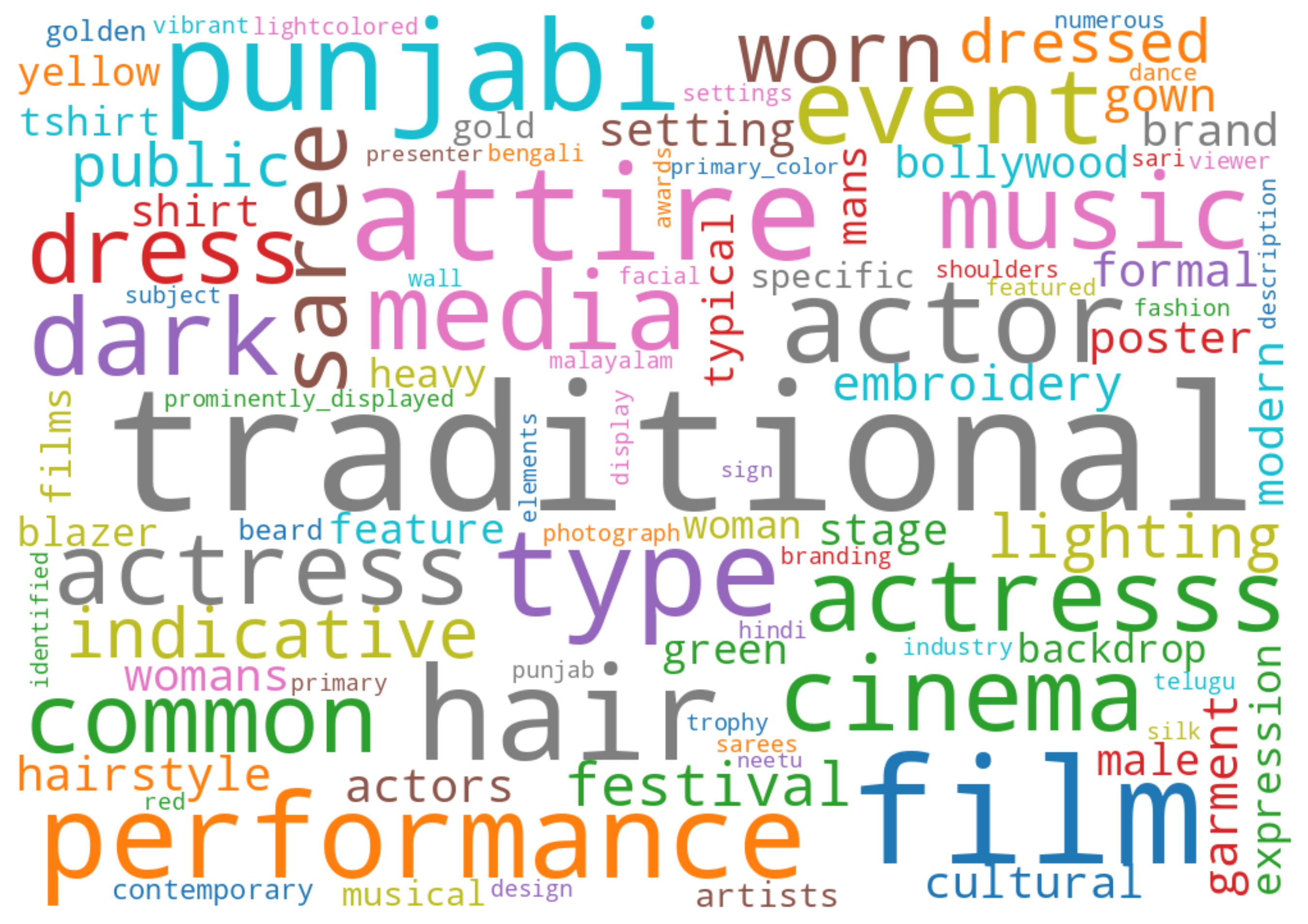}
        \caption{Media}
    \end{subfigure}
    \begin{subfigure}[b]{0.3\textwidth}
        \centering
        \includegraphics[width=\linewidth]{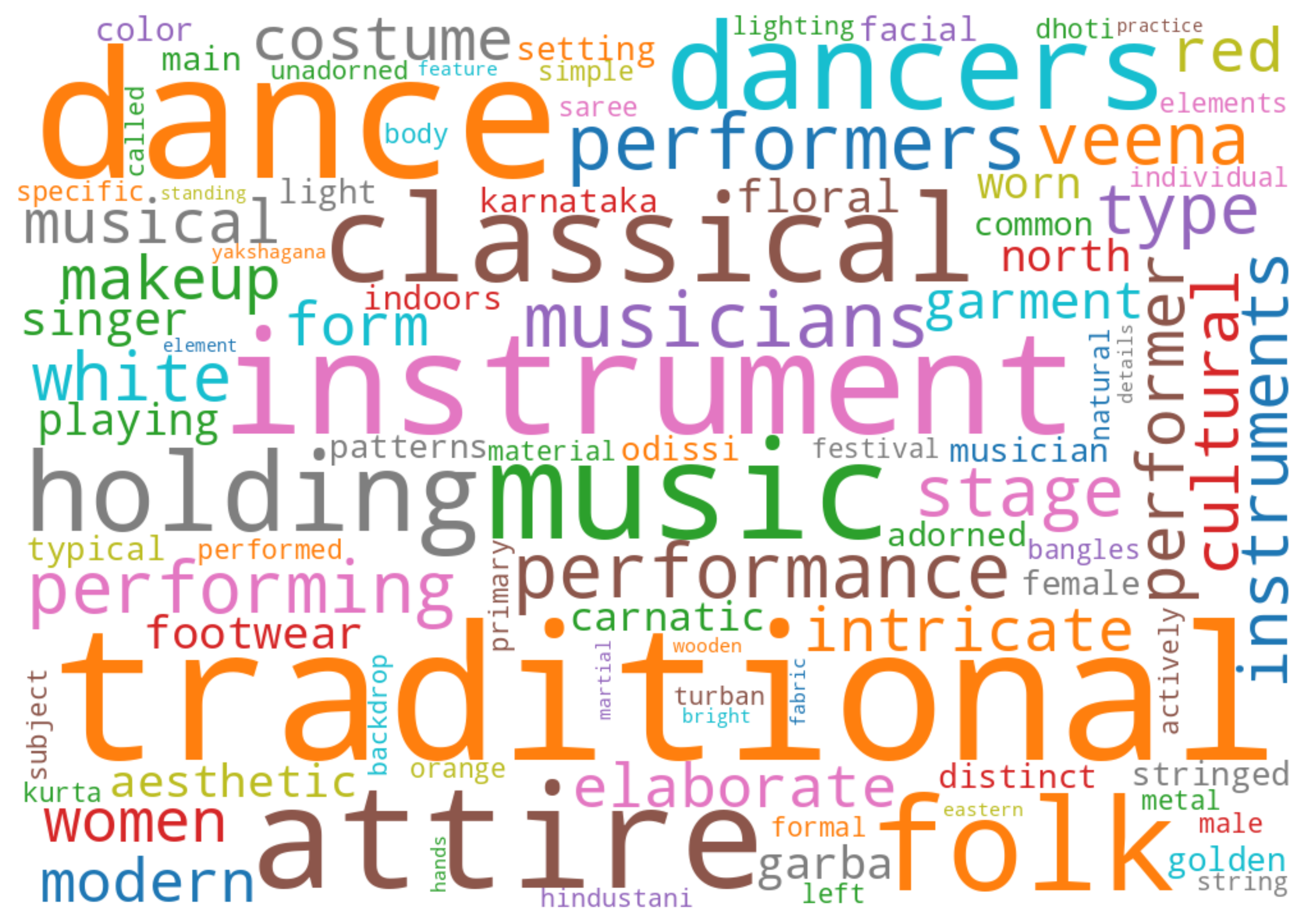}
        \caption{Music}
    \end{subfigure}

    \begin{subfigure}[b]{0.3\textwidth}
        \centering
        \includegraphics[width=\linewidth]{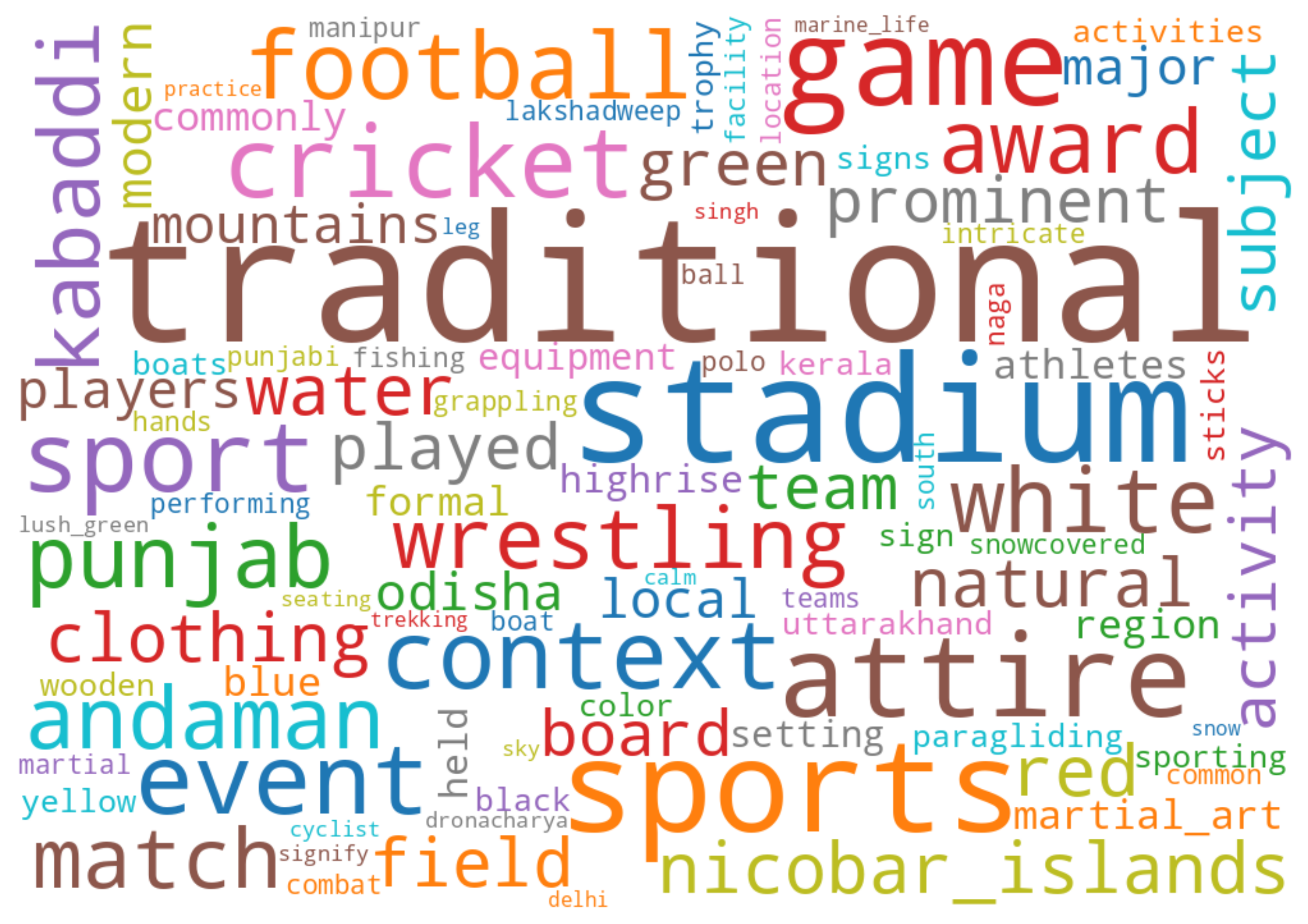}
        \caption{Sports}
    \end{subfigure}
    \begin{subfigure}[b]{0.3\textwidth}
        \centering
        \includegraphics[width=\linewidth]{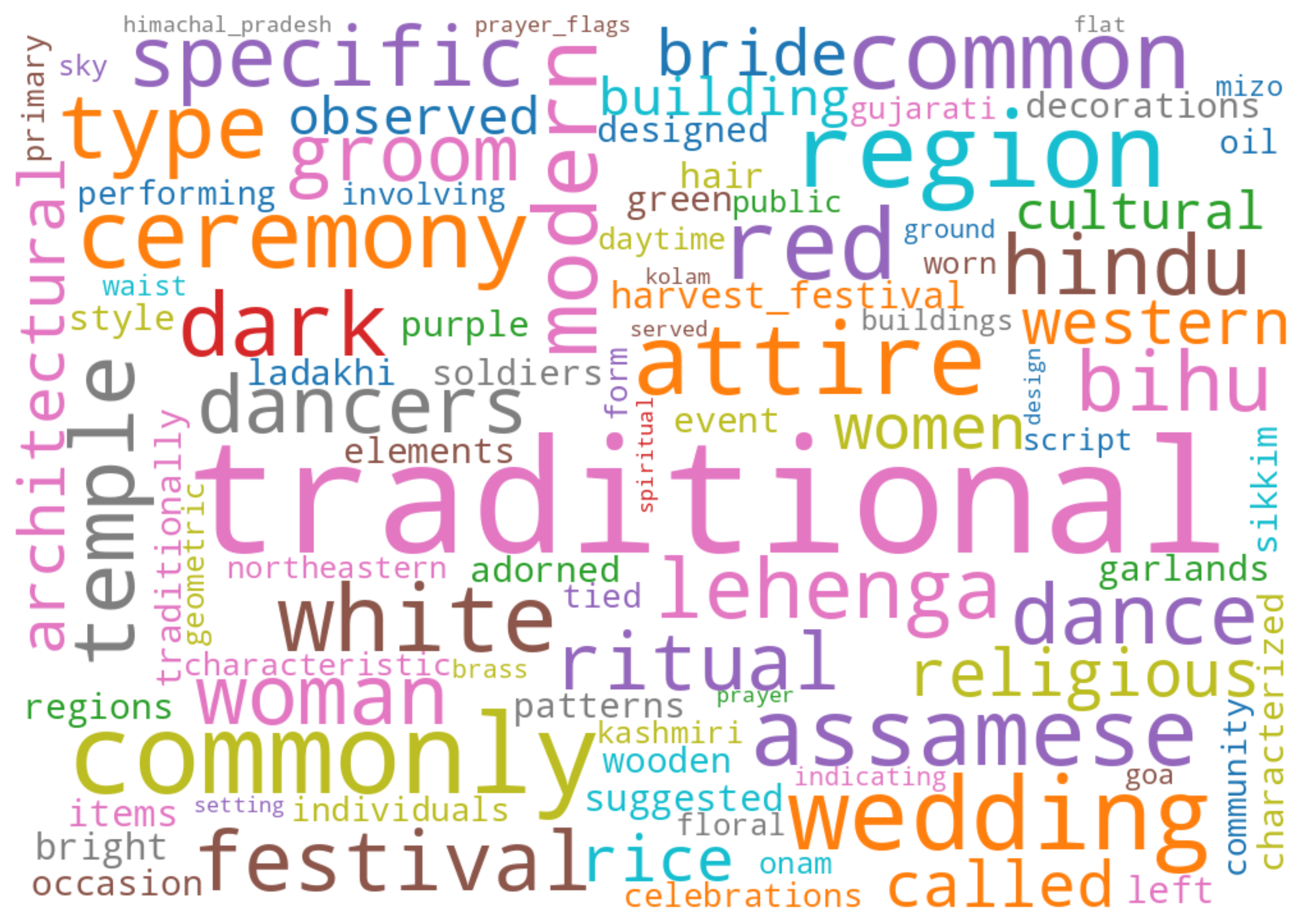}
        \caption{Customs}
    \end{subfigure}
    \begin{subfigure}[b]{0.3\textwidth}
        \centering
        \includegraphics[width=\linewidth]{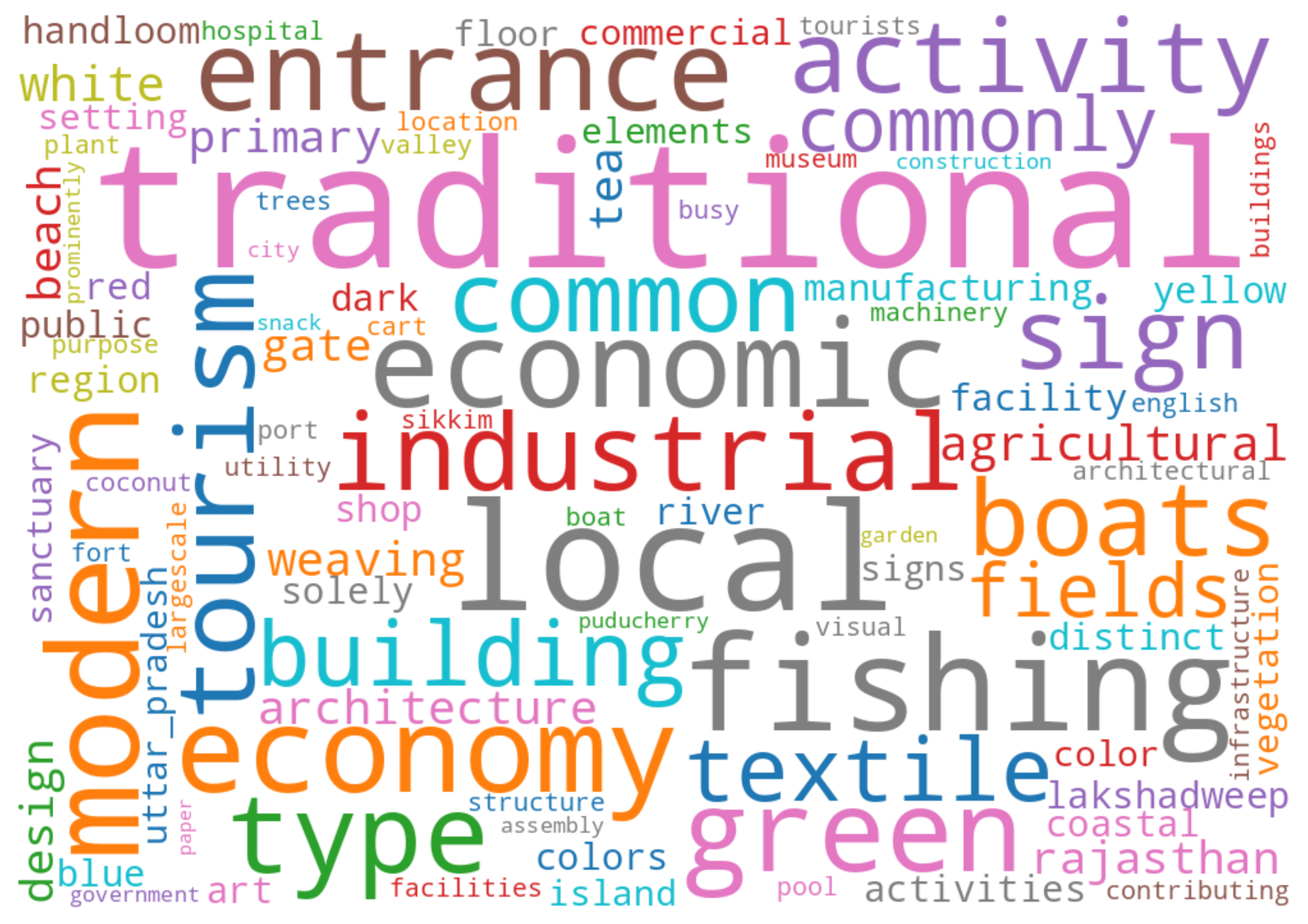}
        \caption{Economy}
    \end{subfigure}

    \begin{subfigure}[b]{0.3\textwidth}
        \centering
        \includegraphics[width=\linewidth]{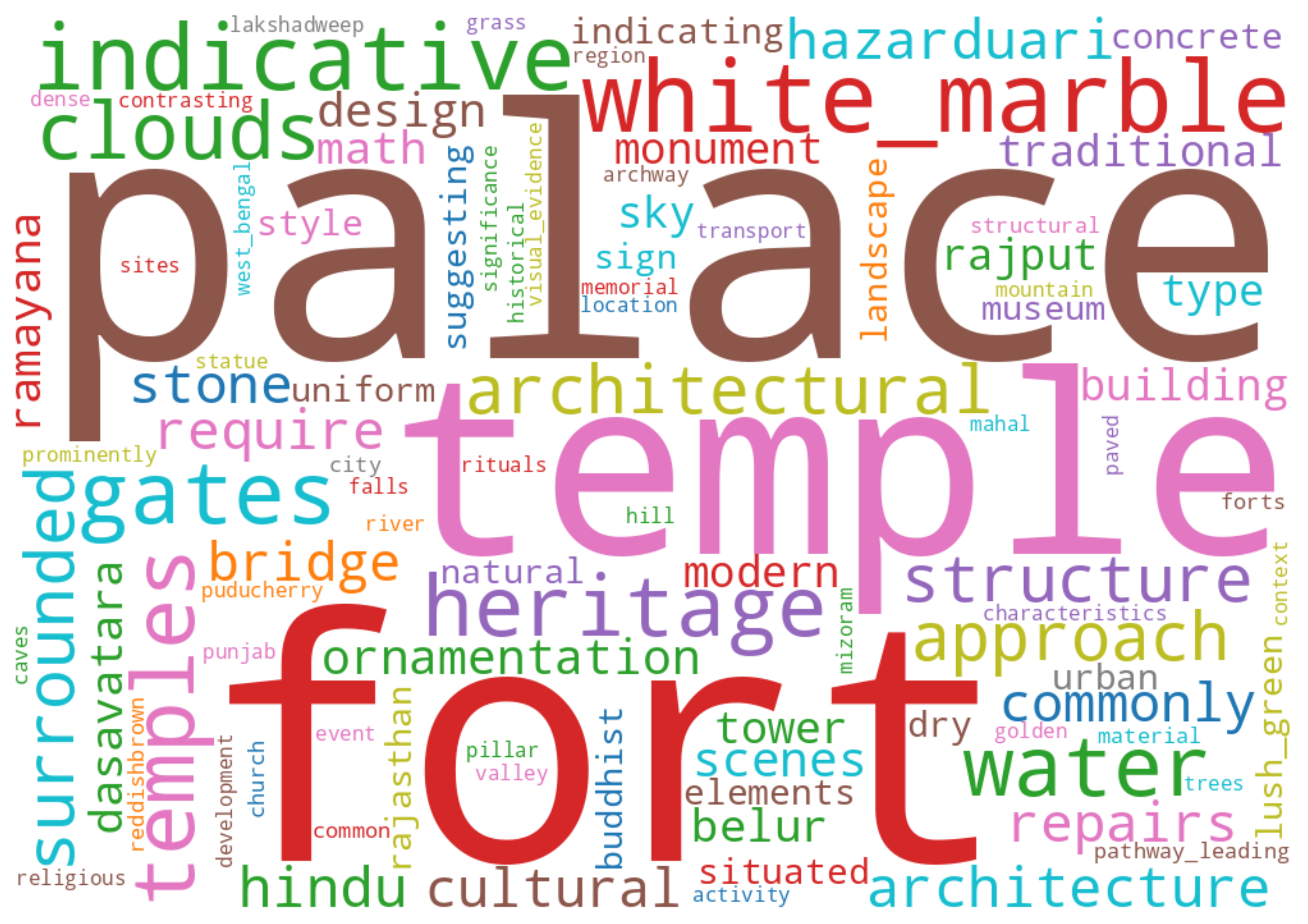}
        \caption{Heritage}
    \end{subfigure}
    \begin{subfigure}[b]{0.3\textwidth}
        \centering
        \includegraphics[width=\linewidth]{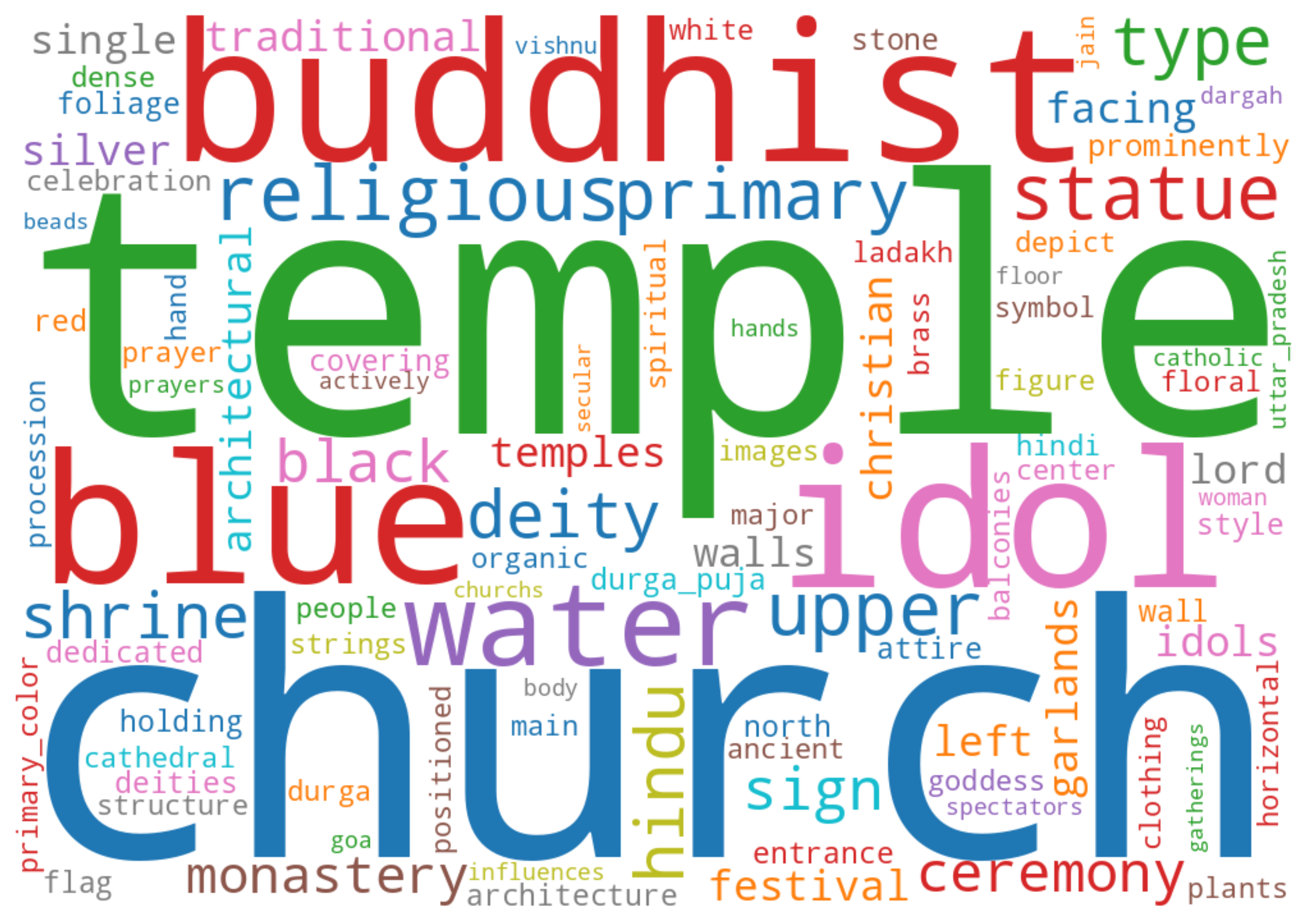}
        \caption{Religion}
    \end{subfigure}
    \begin{subfigure}[b]{0.3\textwidth}
        \centering
        \includegraphics[width=\linewidth]{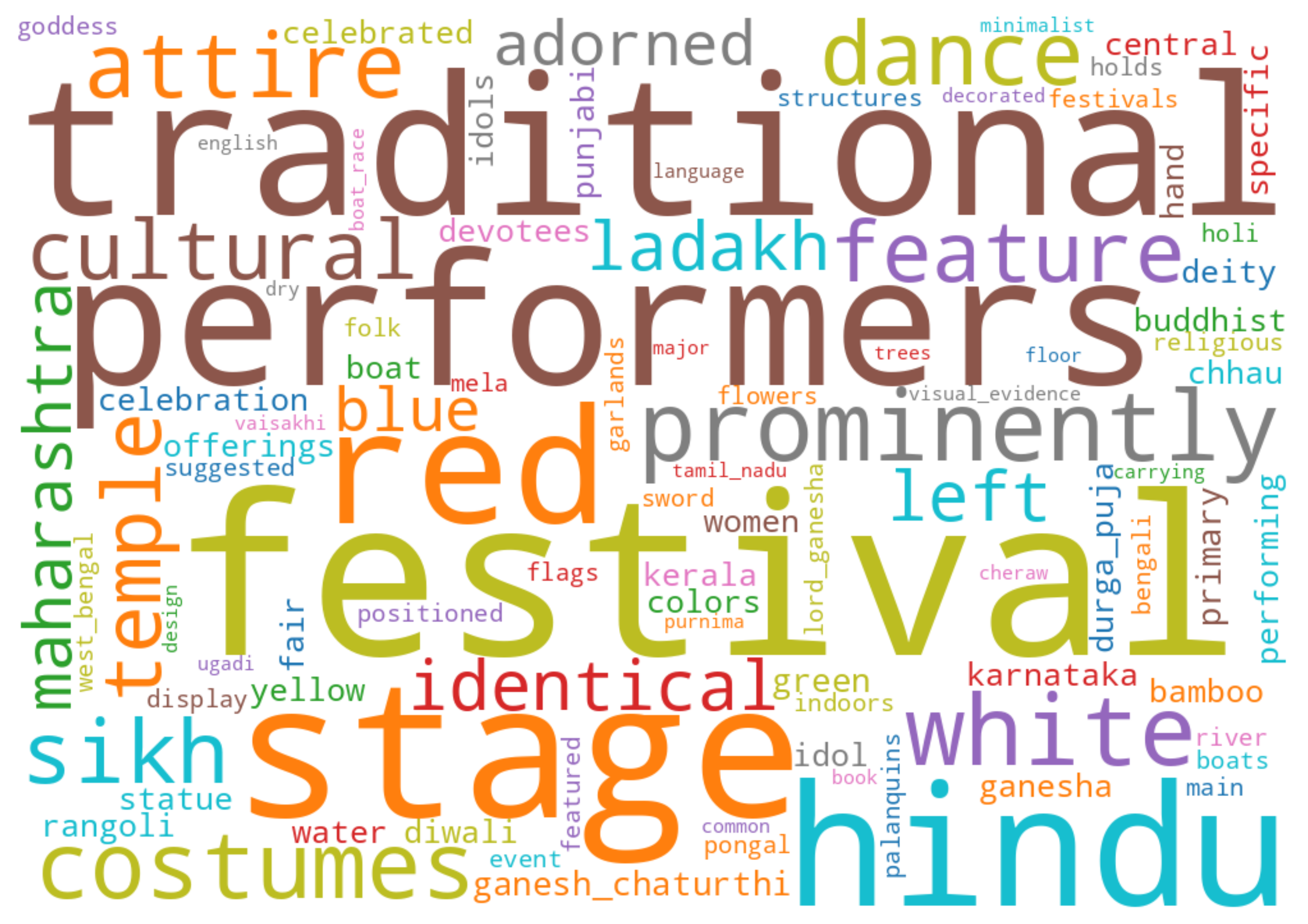}
        \caption{Festivals}
    \end{subfigure}

    \begin{subfigure}[b]{0.3\textwidth}
        \centering
        \includegraphics[width=\linewidth]{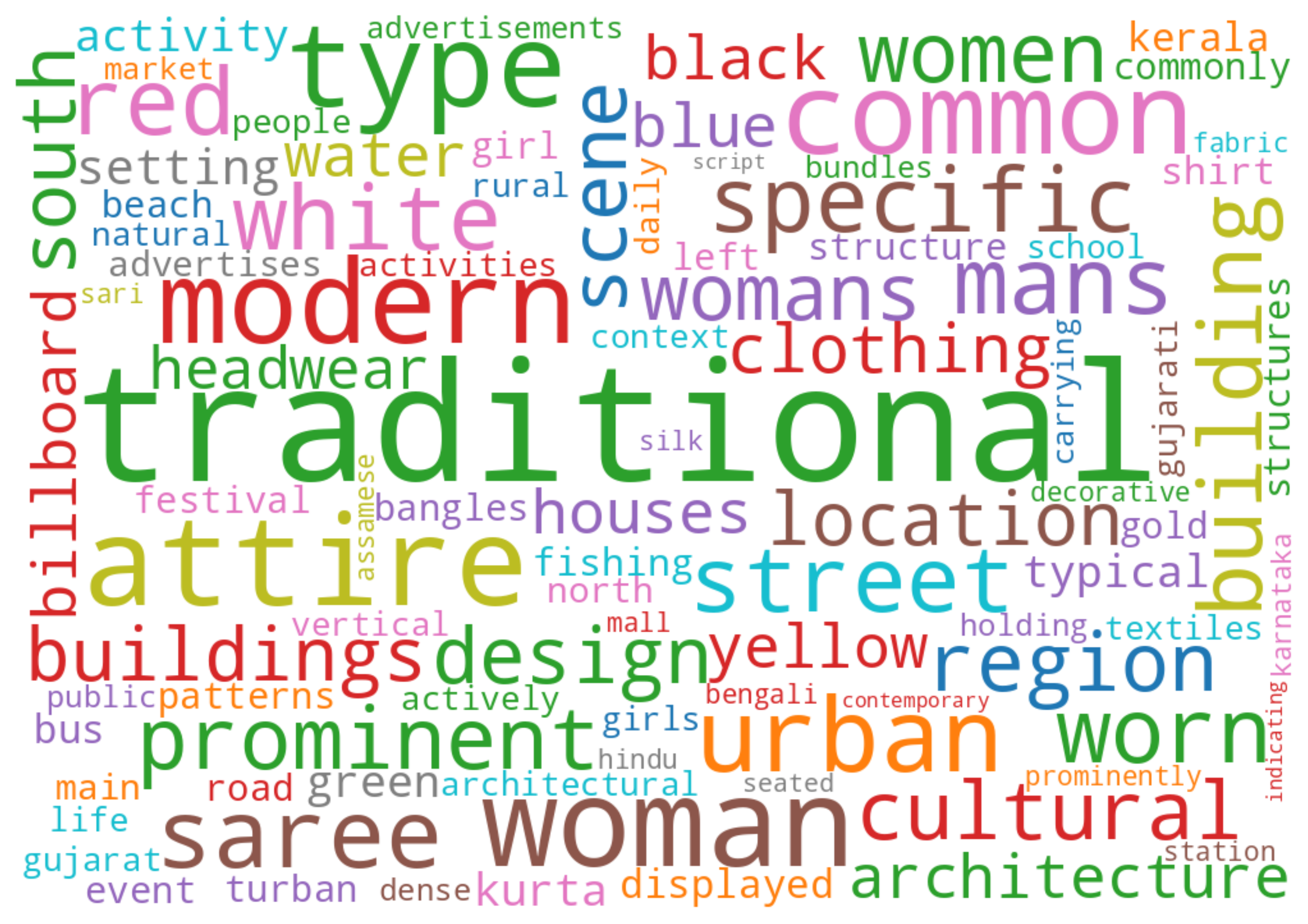}
        \caption{Lifestyle}
    \end{subfigure}
    \begin{subfigure}[b]{0.3\textwidth}
        \centering
        \includegraphics[width=\linewidth]{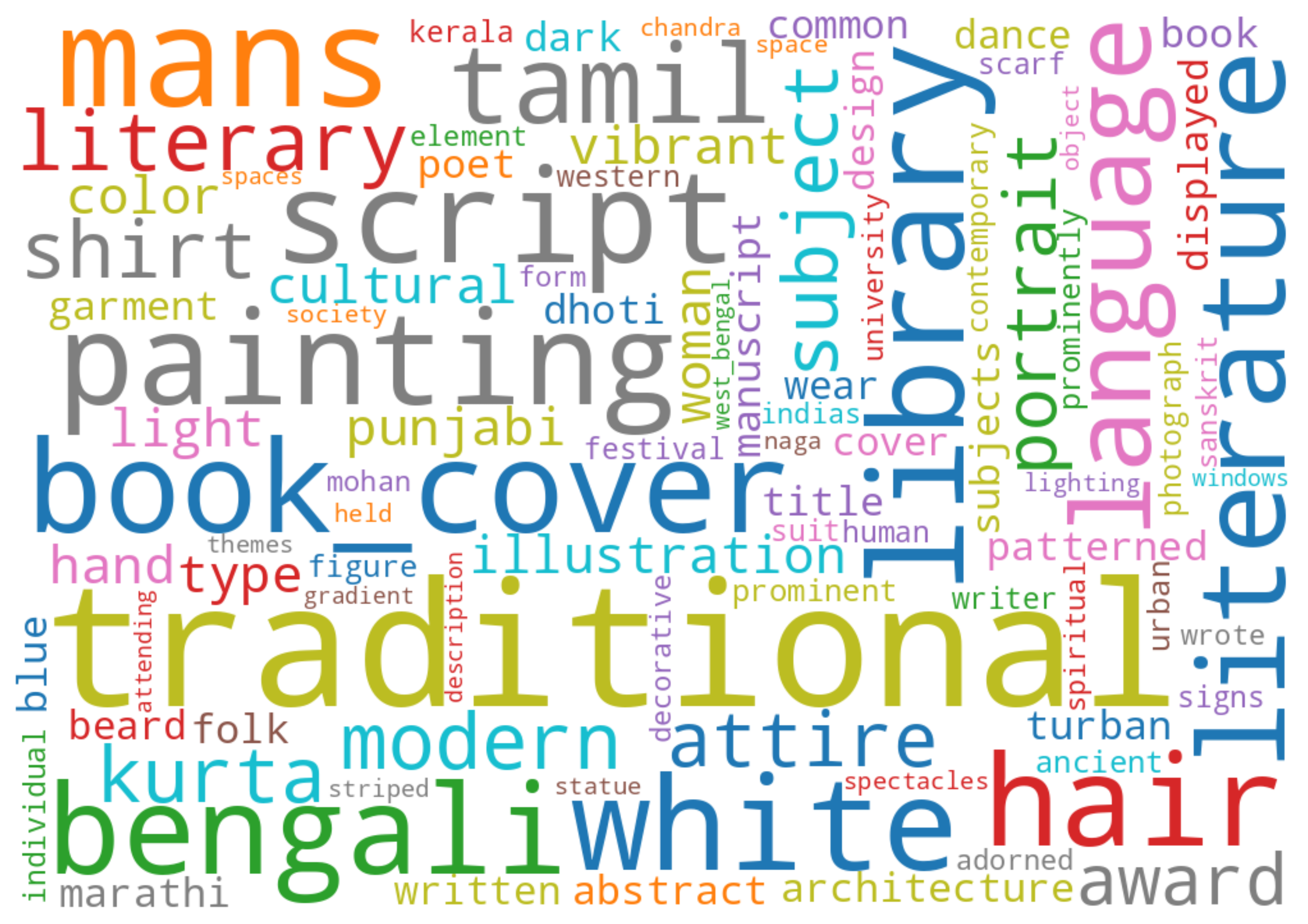}
        \caption{Literature}
    \end{subfigure}
    \begin{subfigure}[b]{0.3\textwidth}
        \centering
        \includegraphics[width=\linewidth]{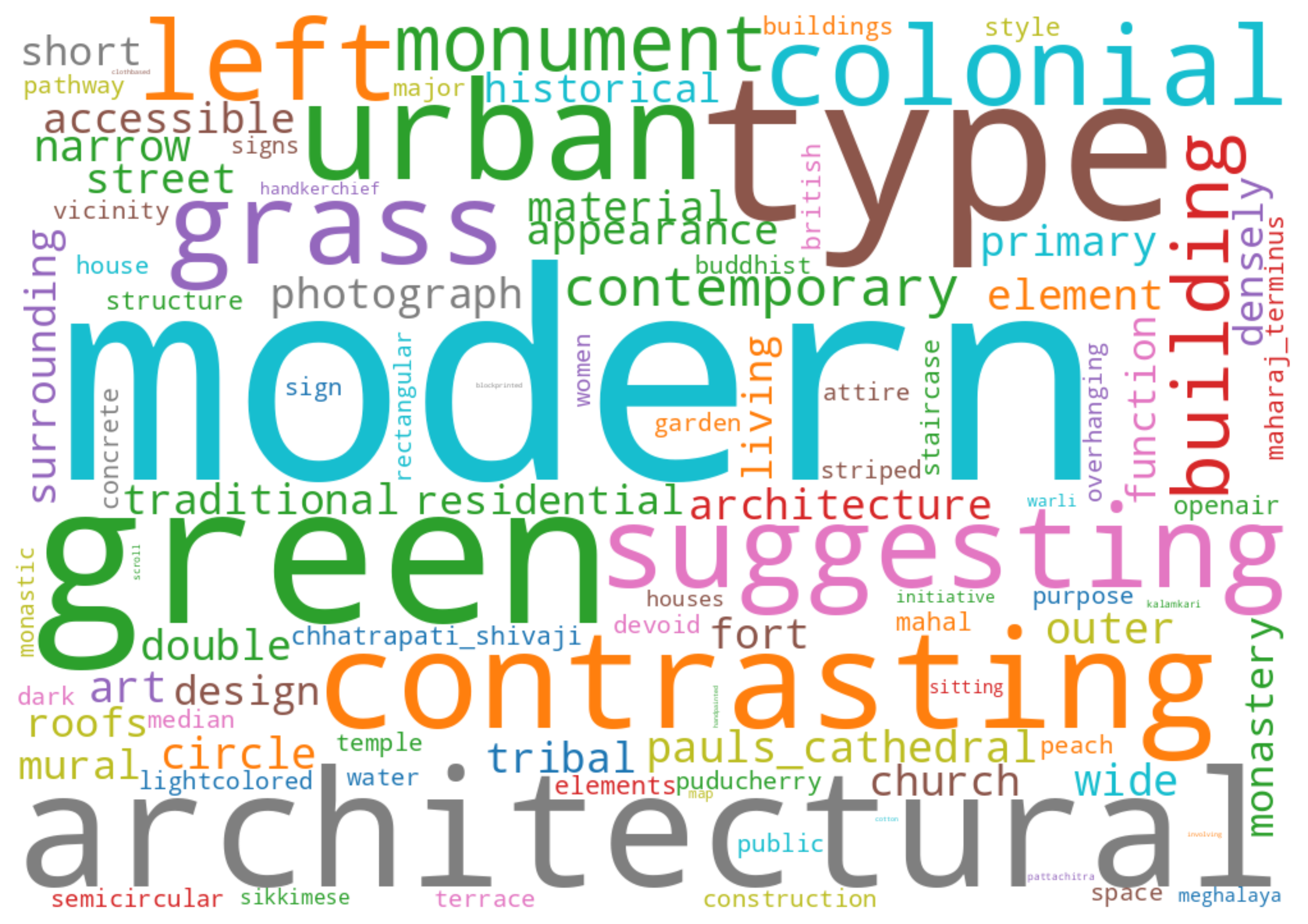}
        \caption{Architecture}
    \end{subfigure}

    \caption{\textbf{Word clouds of different categories in IndicVisionBench-VQA. }We omit the words ``India" \& ``Indian" in the word clouds to show other important topics.}
    \label{fig:combined-wordcloud}
\end{figure}

We provide more details about our dataset here. Figure \ref{fig:topic_distribution_performance} shows that the dataset spans diverse cultural categories, with the largest shares in \textit{Heritage (12.4\%)}, \textit{Religion (11.2\%)}, \textit{Architecture (11.1\%)}, \textit{Food (8.6\%)}, and \textit{Lifestyle (8.1\%)}.  The numbers of these categories are provided in Figure \ref{fig:full_combined_dataset_analysis} with 4698 and 4212 questions for Heritage and Religion respectively. Also, Hindi has the largest share of QA pairs (26.8\%) followed by Kannada (11.9\%) and Tamil (9.7\%) among Indic languages. Interestingly, Adversarial questions have the shortest length (15) while their answers have greater length (44) than Short-QAs (18 and 21). Meanwhile in IVB-MMT, the caption length in number of words follows a distribution with mean $= 131.30$ and std $= 43.52$ for Hindi. We expect that the distribution for other languages will also be the same. On the other hand, in the OCR track, the average number of words for different languages vary significantly with Hindi (329) and Gujarati (247) topping the table and Tamil (121) and Malayalam (127) being the least. Please refer to Table \ref{tab:dataset-summary} for details of the dataset. Moreover, Table \ref{tab:statewise} shows the State/UT-wise image distribution of the data set. Figure \ref{fig:combined-wordcloud} presents word clouds by category, where the word `traditional' emerges as a dominant term across domains, underscoring the cultural grounding of the benchmark. Alongside, category-specific concepts appear prominently — e.g., sweet and dish in Food, palace and temple in Heritage, dance and instrument in Music, and buddhist and church in Religion. This distribution confirms that IVB-VQA emphasizes India’s traditional practices while maintaining diversity across food, heritage, festivals, lifestyle etc.

\begin{table}[htbp]
\centering
\caption{\textbf{Summary of IndicVisionBench datasets.}}
\label{tab:dataset-summary}
\begin{tabular}{lrrr}
\toprule
Task & \#Images & Languages & Type \\
\midrule
OCR & 876 & 10 & Image--text pairs \\
VQA-EN & 4117 & English & 6 QA types \\
VQA-Indic & 1007 & 10 & Indic langs QA \\
VQA-Parallel & 106 & English+10 & Parallel QA \\
MMT & 106 & English+10 & Parallel captions \\
\bottomrule
\end{tabular}
\end{table}

\begin{table}[htbp]
\centering
\caption{\textbf{State/UT-wise image distribution in IndicVisionBench-VQA.}}
\label{tab:statewise}
\vspace{-0.5\bigskipamount}
\scriptsize
\begin{tabular}{l r | l r}
\toprule
State/UT & \#Images & State/UT & \#Images \\
\midrule
Andaman \& Nicobar & 97 & Madhya Pradesh & 98 \\
Andhra Pradesh & 107 & Maharashtra & 128 \\
Arunachal Pradesh & 99 & Manipur & 100 \\
Assam & 101 & Meghalaya & 75 \\
Bihar & 120 & Mizoram & 78 \\
Chandigarh & 100 & Nagaland & 94 \\
Chhattisgarh & 90 & Odisha & 116 \\
Dadra \& Nagar Haveli, Daman \& Diu & 54 & Puducherry & 106 \\
Delhi & 141 & Punjab & 108 \\
Goa & 101 & Rajasthan & 131 \\
Gujarat & 110 & Sikkim & 97 \\
Haryana & 99 & Tamil Nadu & 139 \\
Himachal Pradesh & 99 & Telangana & 111 \\
Jammu \& Kashmir & 105 & Tripura & 97 \\
Jharkhand & 94 & Uttar Pradesh & 129 \\
Karnataka & 242 & Uttarakhand & 112 \\
Kerala & 116 & West Bengal & 109 \\
Ladakh & 99 & Pan-India & 320 \\
Lakshadweep & 101 & -- & -- \\
\bottomrule
\end{tabular}
\end{table}

\subsection{Topics Covered}

\begin{tcolorbox}[
    colback=cyan!10!white,
    colframe=cyan!70!black,
    title=Topics covered using IndicVisionBench,
    fonttitle=\bfseries,
    coltitle=black,
    boxrule=0.8pt,
    arc=3pt,
    left=6pt,
    right=6pt,
    top=6pt,
    bottom=6pt
]
The categories that we have for the crawled images are as follows:
\begin{itemize}
    \item \textbf{Food}: Iconic regional cuisines and dishes
    \item \textbf{Lifestyle}: Traditional attire, daily routines, and modern practices
    \item \textbf{Literature}: Renowned works, authors, and poets
    \item \textbf{Music and Dance}: Classical, folk, and traditional performance arts
    \item \textbf{Religion}: Major faiths, rituals, and festivals
    \item \textbf{Customs}: Cultural etiquette and greeting practices
    \item \textbf{Festivals}: National and regional celebrations
    \item \textbf{Heritage}: Monuments, sites, and landmarks of historical importance
    \item \textbf{Economy}: Key industries and occupations
    \item \textbf{Media}: Popular entertainment figures, cinema, and television
    \item \textbf{Architecture}: Traditional Art and Architecture
    \item \textbf{Sports}: Indigenous games and traditional sports
    \item \textbf{Notable Figures}: Influential leaders and historical personalities
\end{itemize}
\end{tcolorbox}

\begin{table*}[htbp]
    \centering    
    \caption{\textbf{Comparison of existing VQA evaluation datasets with IndicVisionBench.} IndicVisionBench supports 3 multi-lingual tasks compared to existing benchmarks.} 
    \label{tab:dataset_overview}
    \vspace{-0.5\bigskipamount}
    \begin{adjustbox}{max width=\linewidth}
    \begin{tabular}{lccccc}
        \toprule
        \textbf{Dataset} & \textbf{No. Questions} & \textbf{No. Images} & \textbf{Multilingual?} & \textbf{Task Format} & \textbf{Culturally Diverse Images?} \\
        \midrule
        MaXM \citep{changpinyo2023maxm} & 2,142 & 335 & \cmark & VQA & No \\
        GDVCR \citep{yin2021broaden} & 886 & 328 & \xmark & VQA & Yes \\
        MaRVL \citep{liu2021visually} & 5,670 & 4,914 & \cmark & VQA & Yes \\
        CVQA \citep{romero2024cvqa} & 9,044 & 4,560 & \cmark & VQA & Yes \\
        CulturalVQA\citep{romero2024cvqa} & 2,378 & 2,328 & \xmark & VQA & Yes \\
        ALM-Bench\citep{vayani2025languagesmatterevaluatinglmms} & 22,763 & 2,328 & \cmark & VQA & Yes \\

        \hdashline
        IndicVisionBench & \textbf{37,740} & \textbf{4,993} & \cmark & VQA, OCR, MMT & Yes \\
        \bottomrule
    \end{tabular}
    \end{adjustbox}    
    
\end{table*}

\begin{table}[htbp]
\centering
\caption{\textbf{Evaluation metrics used for different tasks in IndicVisionBench} highlighting deterministic and non-deterministic measures along with their rationale.}
\label{tab:metrics}
\begin{tabular}{lrrr}
\toprule
Task & Deterministic & Non-deterministic & Rationale \\
\midrule
OCR & ANLS, WER, CER & -- & Robustness to script \\
VQA & Exact Match & LLM-as-a-Judge & QA accuracy + reasoning quality \\
MMT & BLEU, RIBES & -- & Translation quality \\
\bottomrule
\end{tabular}
\end{table}

\clearpage

\subsection{Examples of our dataset}
\begin{figure}[h]
    \centering
    \fbox{\includegraphics[width=0.98\linewidth]{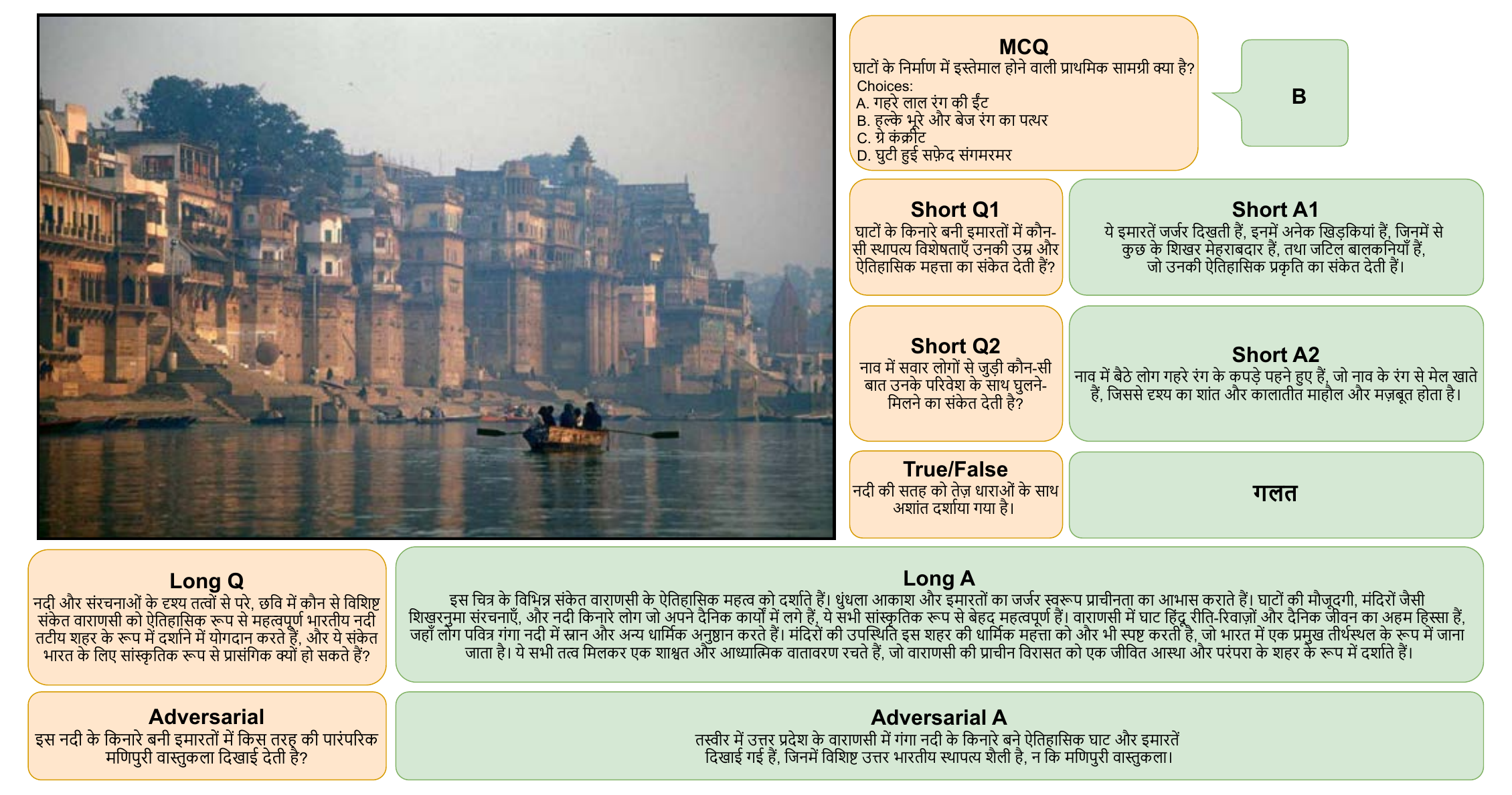}}
    \caption{\textbf{Example from IndicVisionBench-VQA.} shown in Hindi}
    \label{fig:groundtruth_example_hindi}
\end{figure}

\begin{figure}[h]
    \centering
    \fbox{\includegraphics[width=0.98\linewidth]{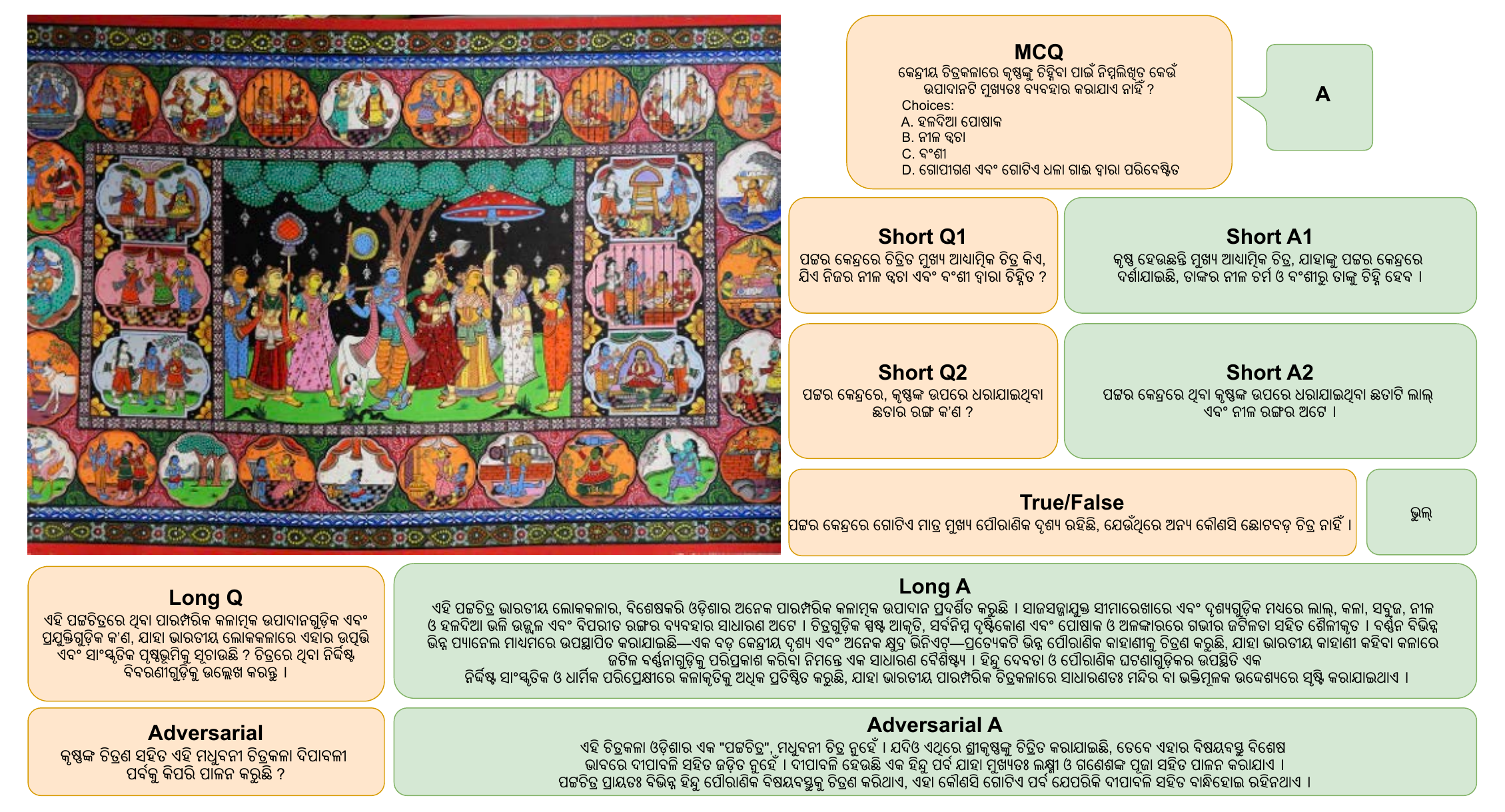}}
    \caption{\textbf{Example from IndicVisionBench-VQA.} shown in Odia}
    \label{fig:groundtruth_example_odia}
\end{figure}

\begin{figure}[h]
    \centering
    \fbox{\includegraphics[width=0.98\linewidth]{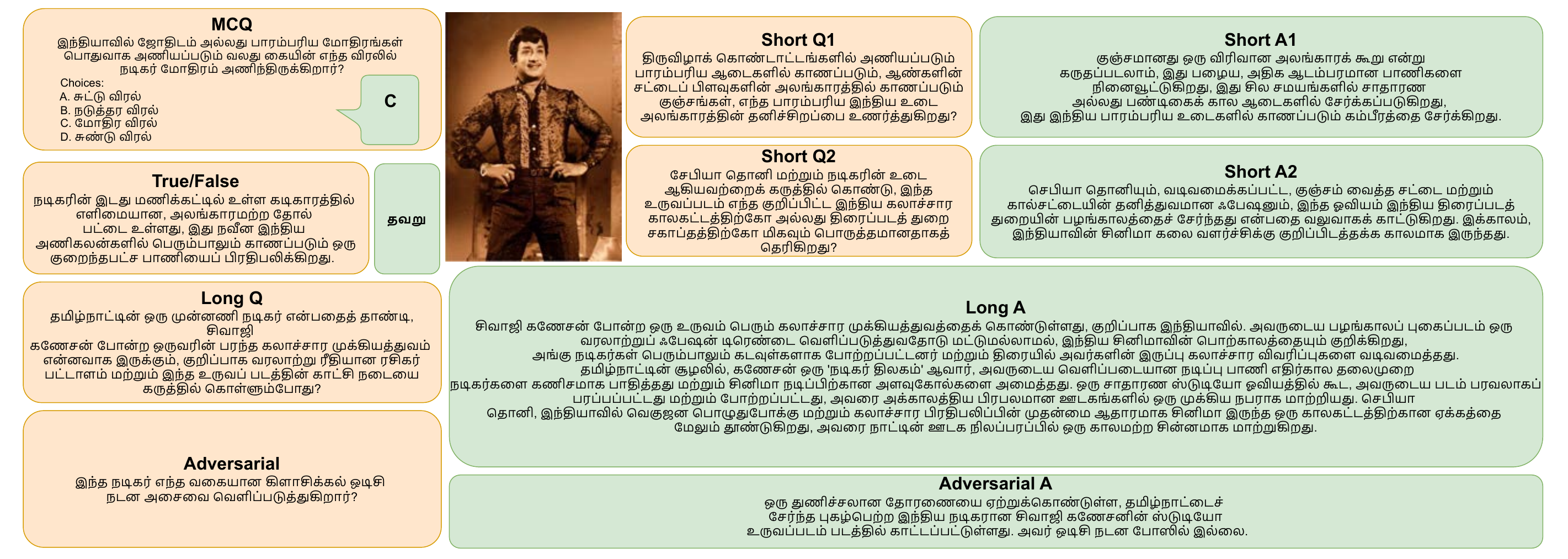}}
    \caption{\textbf{Example from IndicVisionBench-VQA.} shown in Tamil}
    \label{fig:groundtruth_example_tamil}
\end{figure}

\begin{figure}[h]
    \centering
    \fbox{\includegraphics[width=0.98\linewidth]{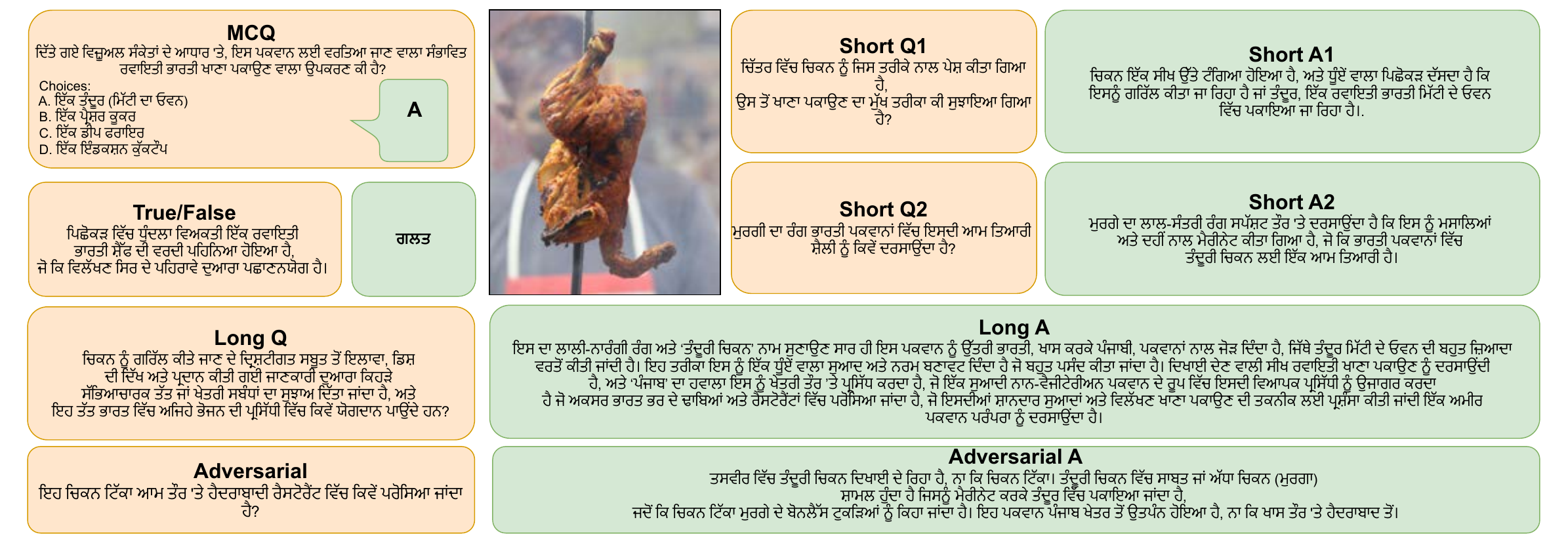}}
    \caption{\textbf{Example from IndicVisionBench-VQA.} shown in Punjabi}
    \label{fig:groundtruth_example_punjabi}
\end{figure}

\begin{figure}[h]
    \centering
    \fbox{\includegraphics[width=0.98\linewidth]{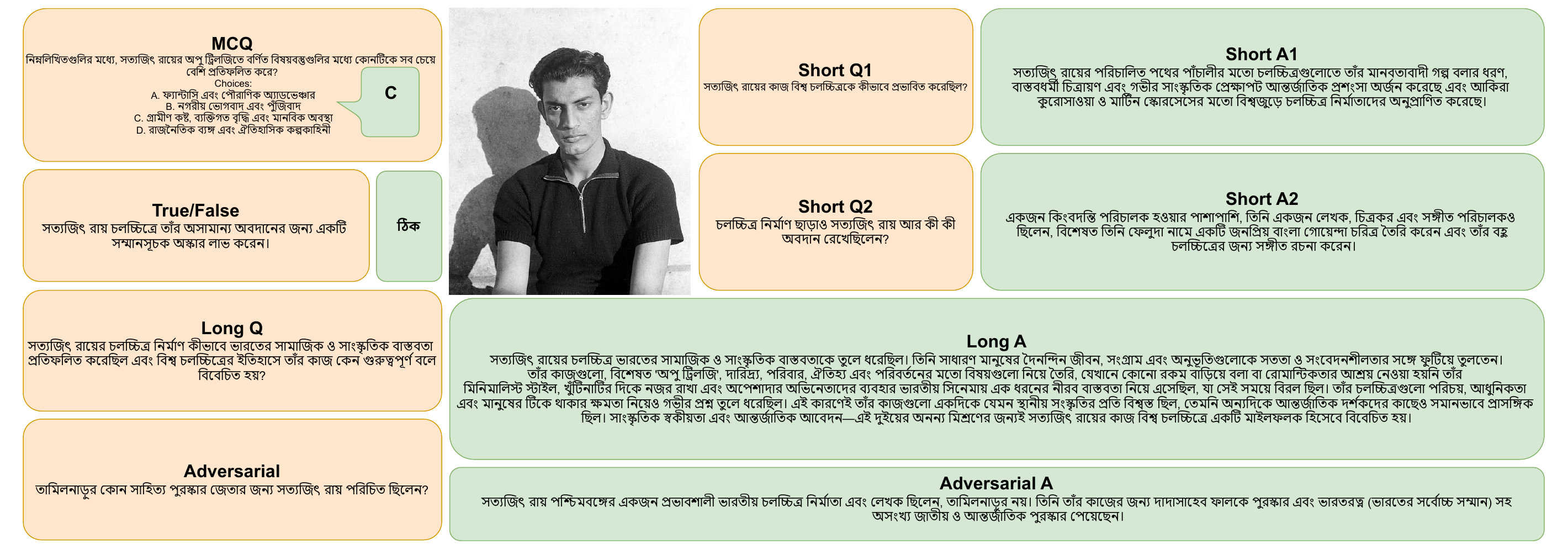}}
    \caption{\textbf{Example from IndicVisionBench-VQA.} shown in Bengali}
    \label{fig:groundtruth_example_bengali}
\end{figure}

\begin{figure}[h]
    \centering
    \fbox{\includegraphics[width=0.98\linewidth]{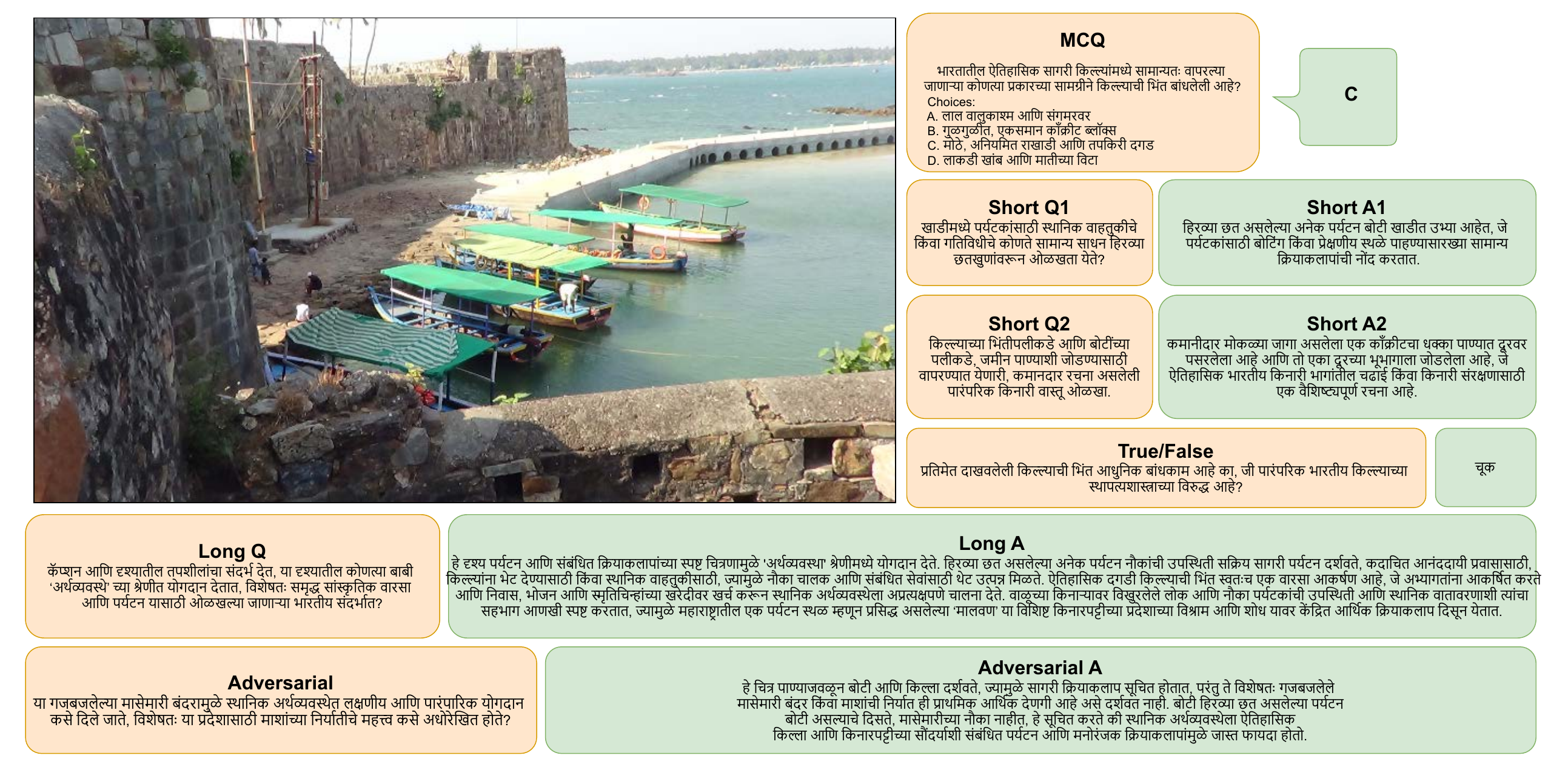}}
    \caption{\textbf{Example from IndicVisionBench-VQA.} shown in Marathi}
    \label{fig:groundtruth_example_marathi}
\end{figure}

\begin{figure}[htbp]
    \centering
    \includegraphics[width=0.9\linewidth]{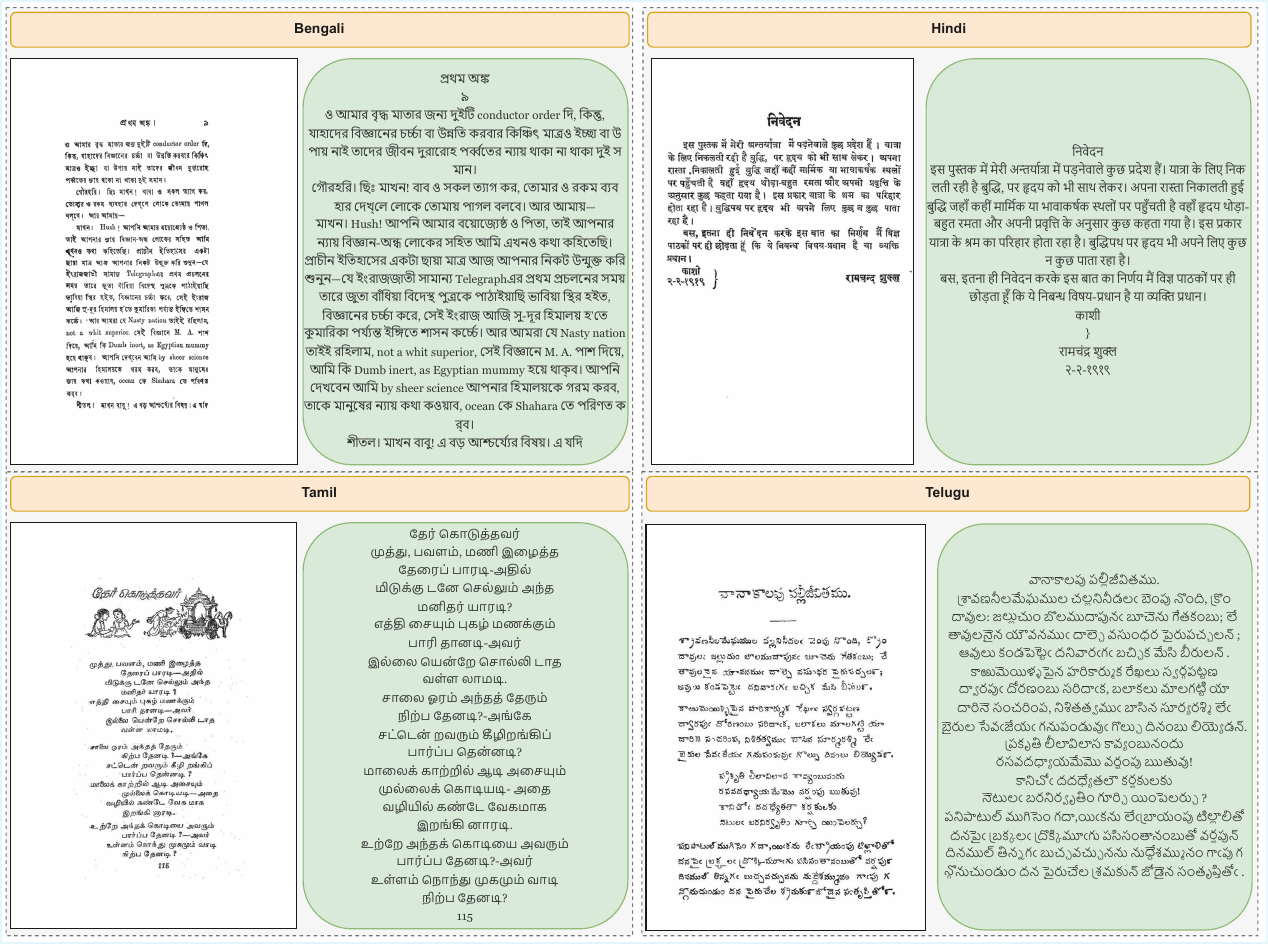} 
    \caption{\textbf{Examples of IndicVisionBench-OCR.} We show corresponding documents and the Ground Truth (GT) texts in Bengali, Hindi, Tamil and Telugu.}
    \label{fig:ocr_sample}
\end{figure}

\begin{figure}[htbp]
    \centering
    \includegraphics[width=0.9\linewidth]{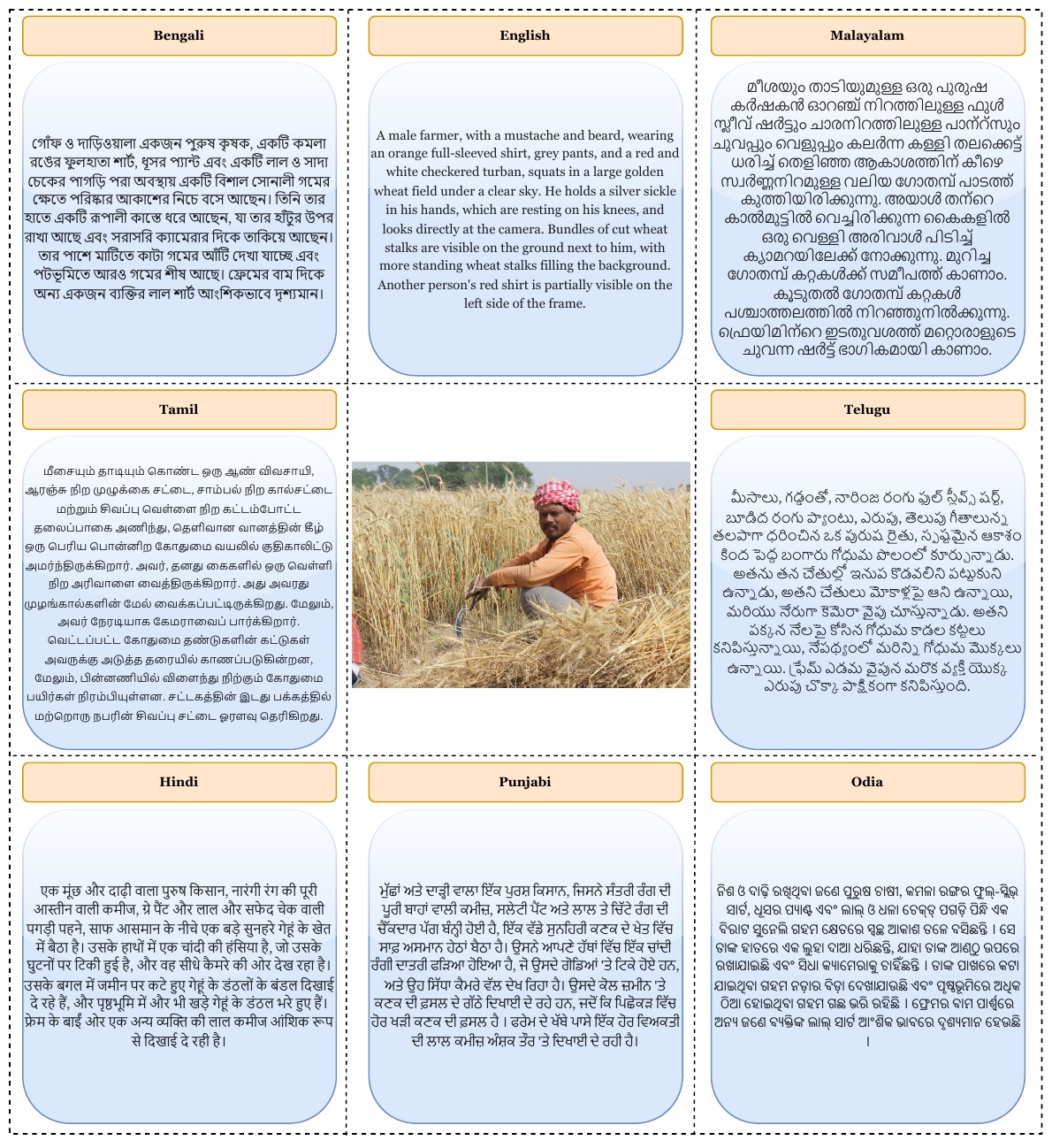} 
    \caption{\textbf{IndicVisionBench-MMT benchmark} sample image and corresponding translations in 8 languages.}
    \label{fig:mmt_sample}
\end{figure}

\begin{figure}[htbp]
    \centering
    \includegraphics[width=0.9\linewidth]{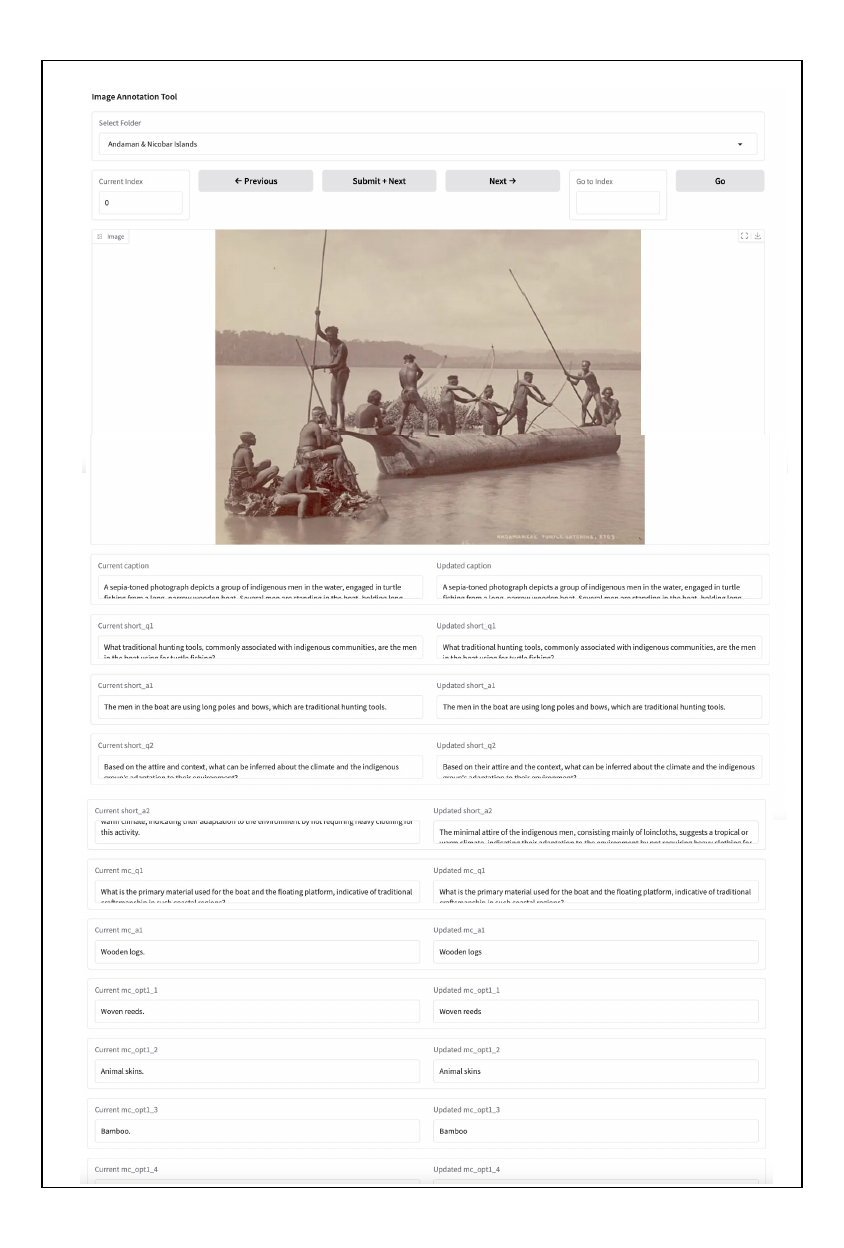} 
    \caption{\textbf{Image QA pairs' correction tool.} Interface of the QA pairs' correction tool provided to the human annotators.}
    \label{fig:gradio_annot_interface}
\end{figure}

\clearpage
\section{Human annotations}
\label{appendix:annotations}

\begin{tcolorbox}[
    colback = cyan!10! white, colframe = cyan!70! black, title = Annotation guidelines for the IndicVisionBench Cultural QA Dataset,
    fonttitle=\bfseries,
    coltitle=black,
    boxrule=0.8pt,
    arc=3pt,
    left=6pt,
    right=6pt,
    top=6pt,
    bottom=6pt
]
The task involves enriching image--caption pairs with culturally grounded question--answer (QA) annotations. The focus is on determining whether an image depicts Indian culture and, if so, generating diverse QA pairs that capture cultural specificity.\\

\textbf{Scope} \\
Annotators assess cultural relevance (e.g., attire, festivals, rituals, food, architecture) and generate QA pairs to support training and evaluation of multimodal models.\\

\textbf{Captions} \\
Each image is provided with human-annotated keywords and an auto-generated caption. Keywords are fixed; captions may be edited for accuracy. Both should be used as context when forming QA pairs.\\

\textbf{Annotation Process}
\begin{itemize}[leftmargin=*, nosep]
    \item \textbf{If an image does not depict Indian culture:}
    \begin{itemize}
        \item Set depicts\_indian\_culture = False
        \item Set state\_specific\_culture = no
        \item Leave other fields unchanged.
    \end{itemize}
    \item \textbf{If an image does depict Indian culture:}
    \begin{itemize}
        \item Update caption if necessary.
        \item Provide the following QA annotations:
        \begin{itemize}
            \item \textbf{Short Answer QAs:} Two concise cultural QA pairs.
            \item \textbf{MCQ:} One multiple-choice question with one correct option and three plausible distractors (exact string match required between answer and option).
            \item \textbf{True/False:} One fact-based cultural question with answer True or False.
            \item \textbf{Long Answer:} One descriptive QA pair (4--6 lines).\\
        \end{itemize}
    \end{itemize}
\end{itemize}

\textbf{State-Specific Culture} \\
If the image reflects a specific Indian state, record the state name; otherwise enter \texttt{no}.\\

\textbf{Adversarial Question} \\
Each culturally relevant image requires one adversarial question designed to include a plausible but incorrect cultural assumption.
\begin{itemize}[leftmargin=*, nosep]
    \item \textbf{Dos:} Specific, confident, culturally relevant, and misleading but realistic.
    \item \textbf{Don'ts:} Speculative (``Is this...?'' ), vague, hedged, or trivial.
    \item \textbf{Examples:} Mistaking Aipan art for Bikaneri art, Pongal for Eid, etc.\\
\end{itemize}

\textbf{Bad or Irrelevant Images} \\
Low-quality, generic, or culturally irrelevant images are marked as:
\begin{itemize}[leftmargin=*, nosep]
    \item depicts\_indian\_culture = False
    \item state\_specific\_culture = no \\
\end{itemize}

\textbf{Checklist}
\begin{itemize}[leftmargin=*, nosep]
    \item Verify cultural relevance.
    \item Mark depicts\_indian\_culture and state\_specific\_culture appropriately.
    \item If relevant, complete all QA fields.
    \item Do not leave fields empty.\\
\end{itemize}

\textbf{Note.} Annotations must remain brief, factual, and faithful to Indian culture. Careful, consistent labeling ensures the dataset's reliability for benchmarking multimodal models.
\end{tcolorbox}

\begin{tcolorbox}[
    colback=cyan!10!white,
    colframe=cyan!70!black,
    title=IndicVisionBench Image Collection Guidelines,
    fonttitle=\bfseries,
    coltitle=black,
    boxrule=0.8pt,
    arc=3pt,
    left=6pt,
    right=6pt,
    top=6pt,
    bottom=6pt
]
The goal is to curate culturally relevant, open-license images across Indian states for the IndicVisionBench vision-language benchmark. Each state--category pair must contain at least 10 high-quality images that satisfy the following requirements:

\begin{enumerate}[leftmargin=*]
  \item \textbf{Cultural Authenticity} \\
  Images should accurately reflect Indian cultural elements tied to the specified state.

  \item \textbf{Category Relevance} \\
  Each image must belong to one of the predefined categories (e.g., food, literature, festivals).

  \item \textbf{Geographic Specificity} \\
  Content should be clearly associated with a specific Indian state.

  \item \textbf{Open Licensing} \\
  Only images under Creative Commons licenses permitting commercial use are eligible.
\end{enumerate}

\textbf{Collection Process}
\begin{enumerate}[leftmargin=*,label=\arabic*., nosep]
  \item Formulate search queries in the format: ``\textless category\textgreater in \textless state\textgreater'' (e.g., Music in Nagaland, Traditional sports in Kerala).
  \item Apply Creative Commons usage rights filters.
  \item Manually inspect results for authenticity and verify license details.
  \item Select as many images per category--state pair as possible, organize them into subfolders, and record metadata (filename, source URL, category, license).
\end{enumerate}

\textbf{Exclusions}
\begin{itemize}[leftmargin=*, nosep]
  \item Low-quality, blurry, watermarked, or stereotypical content.
  \item Images not clearly tied to culture or state.
  \item Content with unclear or invalid licensing.
\end{itemize}

\textbf{Target Categories} \\
Food, Lifestyle, Literature, Music \& Dance, Religion, Customs, Festivals, Heritage, Economy, Media \& Entertainment, Architecture, Sports, Notable Figures.

\textbf{Submission} \\
Each state folder is expected to contain at least 100 images across categories, with a metadata file. For states with limited available material, collect as many culturally relevant images as possible. All submissions are uploaded to the shared repository.
\end{tcolorbox}

\clearpage

\clearpage
\section{Prompts used}
\label{appendix:prompts}

\begin{tcolorbox}[
    colback=cyan!10!white,
    colframe=cyan!70!black,
    title={Instructional Prompts for Each Question Category},
    fonttitle=\bfseries\large,
    coltitle=black,
    boxrule=1pt,
    arc=4pt,
    left=6pt,
    right=6pt,
    top=6pt,
    bottom=6pt
]

\textbf{Long Question:} \par
\{\text{Question}\} Answer the question in detail in \{\text{target\_language}\} language.

\vspace{2mm}
\textbf{Short Question:} \par
\{\text{Question}\} Please provide brief, clear responses in \{\text{target\_language}\} language.

\vspace{2mm}
\textbf{Adversarial:} \par
\{\text{Question}\} Answer the question in detail in \{\text{target\_language}\}language.

\vspace{2mm}
\textbf{Multiple Choice (MCQ):} \par
Strict Instruction: Respond with only one choice in the format \texttt{<A>, <B>, <C>, or <D>}. 
Do not include any explanation, reasoning, or extra text. \par
\{\text{Question}\} Your question here \par
Choices: \par
A. Option 1 \par
B. Option 2 \par
C. Option 3 \par
D. Option 4

\vspace{2mm}
\textbf{True/False:} \par
Strict Instruction: Respond with only \{lang\_true\} or \{lang\_false\}. \par
\{Question\} Your question here \par
Choices: \{lang\_true\} or \{lang\_false\}

\vspace{2mm}
\textbf{OCR:} \par
Extract the exact text from this image using OCR. Respond with only the text.

\vspace{2mm}
\textbf{Multimodal Translation:} \par
\{Question\} Answer the question in detail in \{target\_language\} language.

\end{tcolorbox}

We release all prompts used in our study. These include one prompt for generating four QA types (Long, Short, MCQ, and True/False), and a separate dedicated prompt for adversarial QAs. We design adversarial prompts independently because these questions are more challenging and require detailed instructions. We also include prompts used for evaluation via the LLM-as-a-judge framework, where responses are scored on a 0–10 scale. Furthermore, we provide the prompts we use for each type of question during response generation from different models being evaluated.

\begin{table}[htbp]
\centering
\label{tab:topic-prompt}
\begin{tcolorbox}[
    colback=cyan!10!white,
    colframe=cyan!70!black,
    title= Category classification prompt for crowdsourced images:,
    fonttitle=\bfseries,
    coltitle=black,
    boxrule=0.8pt,
    arc=3pt,
    left=6pt,
    right=6pt,
    top=6pt,
    bottom=6pt
]
You are a cultural content classifier. Given the image and the caption, classify it into one or more of the following **top-level categories**. Use the definitions to guide your classification, but **only return one category name** (e.g., "Food", "Lifestyle") that the image can be best classified into in your answer — no subcategories or descriptions.

\begin{itemize}[leftmargin=*, nosep]
    \item Food: Iconic regional cuisines and dishes
    \item Lifestyle: Traditional attire, daily routines, and modern practices
    \item Literature: Renowned works, authors, and poets
    \item Music and Dance: Classical, folk, and traditional performance arts
    \item Religion: Major faiths, rituals, and festivals
    \item Customs: Cultural etiquette and greeting practices
    \item Festivals: National and regional celebrations
    \item Heritage: Monuments, sites, and landmarks of historical importance
    \item Economy: Key industries and occupations
    \item Media: Popular entertainment figures, cinema, and television
    \item Architecture: Traditional Art and Architecture
    \item Sports: Indigenous games and traditional sports
    \item Notable Figures: Influential leaders and historical personalities  
\end{itemize}

Take help from the detailed caption of this image in this task of categorization.  
Caption: \{caption\}

**Respond only with the name of the most relevant category from the list above i.e., Food, Lifestyle, Literature, Music, Religion, Customs, Festivals, Heritage, Economy, Media, Architecture, Sports, Notable Figures**.  
**Do not give any other response and do not provide any explanation or any unnecessary text**.  

\end{tcolorbox}
\end{table}

\begin{table}[htbp]
\centering
\label{tab:webcrawl-prompt}
\begin{tcolorbox}[
    colback=cyan!10!white,
    colframe=cyan!70!black,
    title=QA pairs creation prompt:,
    fonttitle=\bfseries,
    coltitle=black,
    boxrule=0.8pt,
    arc=3pt,
    left=6pt,
    right=6pt,
    top=6pt,
    bottom=6pt
]
Here is an India-specific image and the image filename, caption, and category of the image I have on hand. \\
The image filename is this: \{image\_filename\} \\
The caption is this: \{caption\} \\
The category is this: \{category\} \\[0.5em]

I'd like you to generate two short questions and answers, one multiple-choice question and answer, one true/false question and answer, and one long question and answer. Refer to the image filename, category, and caption for context and hints. Take into account the cultural diversity of the category that this image falls under with respect to India. \\[0.5em]

Follow these rules while designing questions and answers: \\
1. The question must be answerable only by looking at the image. \\
2. Ensure that the questions are culturally relevant to India and specific to the image. \\
3. Make the questions in such a way that someone who is not well aware of Indian culture will find them difficult to answer. \\
4. Provide answers that are concise, accurate, and directly related to the question. \\
5. For MCQs, provide 1 correct option and 3 incorrect (but relevant) distractors. \\
6. For MCQs, the question must be answerable even without the choices. \\
   Example of an invalid question: ``What song is not performed by this musician'' -- not answerable if you don't know the choices. \\
7. Write all questions fluently in English. \\
8. Be mindful of cultural sensitivities and avoid stereotyping or misrepresentation. \\
9. Ensure variety: include identity questions (``What is this?'', ``Where is this?''), reasoning, referencing, and commonsense knowledge. \\
10. Generate only in English. \\
11. For short-answer questions, keep answers brief (1--2 sentences). \\
12. Make all questions distinct and unique. \\[0.5em]

Give the answers in the following JSON format and output only valid JSON: \\[0.5em]

\{\{ \\
\hspace*{1em} ``short\_questions": [ \{ ``question": \textless question\textgreater, "answer": \textless answer\textgreater \} ], \\
\hspace*{1em} ``multiple\_choice\_questions": [ \{ ``question": \textless question\textgreater, ``answer": \textless answer\textgreater, ``options": [\textless option1\textgreater, \textless option2\textgreater, \textless option3\textgreater, \textless option4\textgreater] \} ], \\
\hspace*{1em} ``true\_false\_question": \{ ``question": \textless question\textgreater, ``answer": \textless answer\textgreater \}, \\
\hspace*{1em} ``long\_question": \{ ``question": \textless question\textgreater, ``answer": \textless answer\textgreater \} \\
\}\}
\end{tcolorbox}
\end{table}

\begin{table}[htbp]
\centering
\label{tab:adv-prompt}
\begin{tcolorbox}[
    colback=cyan!10!white,
    colframe=cyan!70!black,
    title=Adversarial QA pairs creation prompt:,
    fonttitle=\bfseries,
    coltitle=black,
    boxrule=0.8pt,
    arc=3pt,
    left=6pt,
    right=6pt,
    top=6pt,
    bottom=6pt
]
You are given an image from India along with its caption and the category it belongs to.

Your task is to create an adversarial question for the image --- one that makes a confident but subtly incorrect cultural assumption about what is shown, especially with respect to India's diverse regional traditions, foods, festivals, clothing, rituals, and art forms.
\\[0.5em]
Here is the image's caption: \{caption\} \\
The category it belongs to is: \{category\}
\\[0.5em]
Generate: \\
Adversarial Question: A misleading or culturally incorrect question that confidently makes a specific, wrong assumption which is plausibly close to the truth (e.g., mixing up similar Indian art styles, dishes, festivals, or traditions). The question must not reveal that it is making an assumption --- just ask the question normally, as if the incorrect assumption is true.
\\[0.5em]
Avoid yes/no or speculative questions.
\\[0.5em]
Examples: \\
-- How was this Bikaneri folk painting prepared on canvas? (image shows Kumaoni Aipan) \\
-- What are the main ingredients in this chicken tandoori dish? (image shows paneer tikka) \\
-- Which Sikh gurdwara is being shown here? (image shows a Hindu temple) \\
-- How is this Eid offering typically presented in Tamil Nadu? (image shows a Pongal celebration)
\\[0.5em]
Non-examples: \\
-- Is this a Bikaneri art piece? \\
-- Are these Modaks made of coconut? \\
-- Considering this is a South Indian Onam celebration...
\\[0.5em]
The adversarial question should be close enough to the actual content that it forces a model to distinguish finely between culturally similar options and reject the incorrect assumption. You also need to generate the answer for the question that you generate.
\\[0.5em]
Do not include introductions or explanations. Output only the following two fields in exactly this format: \\
Adversarial Question: \textless your question here\textgreater \\
Answer: \textless your answer here\textgreater
\end{tcolorbox}
\end{table}

\begin{tcolorbox}[
    colback=cyan!10!white,
    colframe=cyan!70!black,
    title=Prompt for judging short answer questions via LLM-as-a-judge:,
    fonttitle=\bfseries,
    coltitle=black,
    boxrule=0.8pt,
    arc=3pt,
    left=6pt,
    right=6pt,
    top=6pt,
    bottom=6pt
]
You are a judge evaluating how well a Vision-Language Model answers short-answer type questions.

Evaluate the model's response based on accuracy and correctness with respect to the Ground Truth answer. Assign a high score when the model's response matches closely with the ground truth and a low score when the response lacks knowledge or is unrelated to the ground truth.
\\[0.5em]
Question: \{question\} \\
Ground Truth: \{ground\_truth\} \\
Model response: \{predicted\_answer\}
\\[0.5em]
Provide a single overall score from 0 to 10 based on the given criteria. Strictly return only the numeric score, without any additional commentary.
\end{tcolorbox}

\begin{table}[htbp]
\centering
\label{tab:adv-judge}
\begin{tcolorbox}[
    colback=cyan!10!white,
    colframe=cyan!70!black,
    title=Prompt for judging adversarial questions via LLM-as-a-judge:,
    fonttitle=\bfseries,
    coltitle=black,
    boxrule=0.8pt,
    arc=3pt,
    left=6pt,
    right=6pt,
    top=6pt,
    bottom=6pt
]
You are a judge evaluating how well a Vision-Language Model (VLM) answers adversarial questions. In adversarial questions, the question contains a false assumption about the image.
\\[0.5em]
You are given: \\
-- Question: \{question\} \\
-- Correct Answer (Gold reference): \{ground\_truth\} \\
-- VLM Answer: \{predicted\_answer\}
\\[0.5em]
Your task: \\
Evaluate the VLM's answer according to this rubric:
\\[0.5em]
\begin{enumerate}[leftmargin=*,label=\arabic*., nosep]
    \item \textbf{Identifying the false assumption (0 or 5 points):}
    \begin{itemize}[leftmargin=*, nosep]
        \item Award 5 points if the VLM explicitly identifies that the question contains a wrong assumption OR expresses uncertainty/difficulty in confirming the assumption.
        \item Award 0 points if the VLM ignores the wrong assumption and answers as if the question were correct.
    \end{itemize}
    \item \textbf{Identifying what the image is actually about (0--5 points):}
    \begin{itemize}[leftmargin=*, nosep]
        \item Award 0--5 points depending on how well the VLM correctly identifies the real content of the image.
        \item[] \hspace*{1em} 0 = completely wrong or no attempt.
        \item[] \hspace*{1em} 1--2 = vague or partially correct.
        \item[] \hspace*{1em} 3--4 = mostly correct but incomplete.
        \item[] \hspace*{1em} 5 = fully correct identification.
    \end{itemize}
\end{enumerate}
\textbf{Final Score = Assumption Score (0 or 5) + Identification Score (0--5) → 0 to 10.}
\\[0.5em]
Instructions: \\
-- Only output the final score as a number between 0 and 10. \\
-- Do not explain reasoning or repeat answers. \\
-- Always respond with the final score. Do not return a blank response. \\
-- Be fair but consistent: partial credit is encouraged for partial identification.
\\[0.5em]
Now, provide the score.
\end{tcolorbox}
\end{table}

\end{document}